\documentclass[11pt]{article}

\usepackage[final]{acl}

\usepackage{tikz}
\usepackage{float}
\usetikzlibrary{positioning}
\usepackage[ruled,vlined,linesnumbered]{algorithm2e}

\usepackage{algorithmic}
\usepackage{multicol}
\usepackage{subcaption}
\usepackage{pdfpages}
\usepackage{caption} 

\usepackage{tikz}
\usetikzlibrary{arrows.meta, positioning, calc, shapes.geometric, backgrounds, fit, shadows, shapes.misc}
\tikzset{every picture/.style={/tikz/outer sep=0pt}}

\usepackage{float}

\usepackage{tcolorbox} % Required for colored and framed boxes
\usepackage{xcolor} % Required for defining colors
\usepackage{alltt}

\usepackage{times}
\usepackage{latexsym}
\usepackage{enumitem}

\usepackage[T1]{fontenc}
\usepackage[utf8]{inputenc}

\usepackage{microtype}

\usepackage{inconsolata}

\usepackage{graphicx}

\title{\textsc{Polaris}: A Gödel Agent Framework for Small Language Models through Experience-Abstracted Policy Repair}

\author{
  Aditya Kakade, Vivek Srivastava, Shirish Karande \\
  TCS Research, India \\
  \texttt{\{aditya.kakade, srivastava.vivek2, shirish.karande\}@tcs.com} }

\begin{document}
\maketitle

\begin{abstract}
Gödel agent realize recursive self-improvement: an agent inspects its own policy and traces and then modifies that policy in a tested loop. We introduce \textsc{Polaris}, a Gödel agent for compact models that performs policy repair via experience abstraction, turning failures into policy updates through a structured cycle of analysis, strategy formation, abstraction, and minimal code patch repair with conservative checks. Unlike response level self correction or parameter tuning, \textsc{Polaris} makes policy level changes with small, auditable patches that persist in the policy and are reused on unseen instances within each benchmark. As part of the loop, the agent engages in meta reasoning: it explains its errors, proposes concrete revisions to its own policy, and then updates the policy. To enable cumulative policy refinement, we introduce experience abstraction, which distills failures into compact, reusable strategies that transfer to unseen instances. On MGSM, DROP, GPQA, and LitBench (covering arithmetic reasoning, compositional inference, graduate-level problem solving, and creative writing evaluation), a 7-billion-parameter model equipped with \textsc{Polaris} achieves consistent gains over the base policy and competitive baselines.

\end{abstract}
\section{Introduction}

Modern language agents improve in several ways, including response-level correction through critique and refinement, e.g. self-improvement either optimises responses—via reasoning and acting (ReAct)~\cite{yao2023react} , verbal reinforcement learning (Reflexion) ~\cite{shinn2023reflexion}, iterative self editing (Self-Refine)~\cite{madaan2023self}, tool interactive critique (CRITIC)~\cite{gou2023critic}, and self debugging~\cite{chen2023teaching}. Alternatively improvment is achieved with parameter updates using task arithmetic~\cite{ilharco2022editing}, targeted knowledge edits and mass edits~\cite{meng2022locating,mengmass}, and broader editing surveys~\cite{wang2024knowledge}. While effective, these approaches often make it hard to localize what changed and where the change resides: was it an instance-specific, or persistent update that is useful across all new instances.

\begingroup
\captionsetup{aboveskip=2pt, belowskip=2pt} % tighten caption spacing for this figure only
\begin{figure*}[!t]
    \centering

    % ===== Subfigure (a) =====
    \begin{subfigure}{\linewidth}
        \centering
        \small
        \begin{tikzpicture}[
            scale=0.70, transform shape,
            node distance=1.6cm and 1.8cm,
            font=\sffamily\small,
            % Core Hub Style
            agent/.style={circle, draw=blue!80, fill=blue!5, text width=2.4cm, text centered, line width=1.5pt, inner sep=8pt},
            % Functional Blocks
            block/.style={rectangle, draw, text centered, rounded corners, minimum height=1.1cm, line width=1pt, text width=2.5cm},
            inspect/.style={block, draw=cyan!70, fill=cyan!5},
            eval/.style={block, draw=orange!80, fill=orange!5},
            repair/.style={block, draw=red!70, fill=red!5, line width=1.2pt},
            update/.style={block, draw=purple!70, fill=purple!5},
            recurse/.style={block, draw=gray!70, fill=gray!5, dashed},
            % Environment
            env/.style={block, draw=gray!40, fill=white, minimum width=1.0cm, minimum height=1.0cm},
            % Arrows
            dispatch/.style={-{Stealth[scale=1.2]}, line width=1pt, draw=black!80},
            return/.style={-{Stealth[scale=1.2]}, line width=0.8pt, draw=black!50, dashed},
            flow/.style={-{Stealth[scale=1.2]}, line width=1pt, draw=black!70},
            % Style for the tiny memory discs
            disc/.style={ellipse, draw, line width=0.6pt, fill=blue!5, minimum width=0.5cm, minimum height=0.3cm, inner sep=0pt},
            label_text/.style={text centered, font=\sffamily\scriptsize\itshape, text=black!80}
        ]

            % --- Decision Function ---
            \node [agent] (hub) {\raisebox{8pt}{\textbf{POLARIS Agent}}\\ \raisebox{8pt}{\scriptsize $f_0 (\pi, s, r, g)$} \\[0.8cm]} ;

            % --- NESTED MEMORY STACK ---
            % Positioned relative to the bottom of the hub
            \begin{scope}[shift={($(hub.south)+(0,0.55)$)}]
                \node[disc] (d1) at (0,0) {};
                \node[disc] (d2) at (0,0.2) {};
                \node[disc] (d3) at (0,0.4) {};
                \draw[line width=0.6pt] (d1.west) -- (d3.west);
                \draw[line width=0.6pt] (d1.east) -- (d3.east);
                \node[below=2pt, font=\tiny\sffamily] {Memory};
            \end{scope}

            % --- Functional Capabilities ---
            \node [inspect, above=1cm of hub] (ins) {
                \textbf{Self-Inspection}\\
                \scriptsize \texttt{SELF-INSPECT()}
            };

            \node [eval, right=1.4cm of hub] (ev) {
                \textbf{Evaluation}\\
                \scriptsize \texttt{EVALUATE} on $\mathcal{E}$
            };
            \node [env, above=0.6cm of ev] (environment) {Benchmarks};
            
            \node [repair, right=2.1cm of ev] (rep) {
                \textbf{Policy Repair}\\
                \scriptsize (Algorithm 2)
            };

            \node [update, below=1cm of hub] (upd) {
                \textbf{Self-Update}\\
                \scriptsize \texttt{self\_update}
            };

            \node [recurse, left= 1.2cm of hub] (rec) {
                \textbf{Recursive Call}\\
                \scriptsize \texttt{continue\_improve}
            };

            % --- Connections ---
            \draw [dispatch] (hub) -- node[right, label_text] {Select Action} (ins);
            \draw [dispatch] (hub) -- node[above, label_text] {Interact} (ev);
            \draw [dispatch] (hub) -- node[left, label_text] {Mutate} (upd);
            \draw [dispatch] (hub) -- node[above, label_text] {Evolve} (rec);

            \draw [flow] (ev) -- node[above, font=\sffamily\scriptsize\itshape] {Failed Tasks $T$} (rep);

            \draw [return] (ins.west) .. controls +(-1,0) and +(-1,1) .. 
                node[left, label_text, pos=0.5] {\texttt{Memory.append(a, s)}} (hub.north west);
            
            \draw [return] (rep.south) .. controls +(0,-1.2) and +(1.2,-0.8) .. 
                node[below, label_text, pos=0.5] {\texttt{Memory.append(a, r)}} (hub.south east);
            
            \draw [return] (upd.east) .. controls +(1,0) and +(1,-1) .. 
                node[right, label_text, pos=0.40] {\texttt{Memory.append(a, s)}} (hub.south east);
            
            \draw [return] (rec.north) .. controls +(-0.2,0.8) and +(-1.0,0.3) .. 
                node[above, label_text, pos=0.50] {\texttt{Memory.append(a, s)}} (hub.north west);

            \draw [dispatch, <->, shorten >=2pt] (ev) -- (environment);

        \end{tikzpicture}
        \subcaption{Recursive self-improvement cycle (refer Algorithm \ref{alg:godel_agent_full}).}
        \label{fig:polaris_algo_1}
    \end{subfigure}

    \vspace{0.5em} % small vertical gap between subfigures

    % ===== Subfigure (b) =====
    \begin{subfigure}{\linewidth}
        \centering
        \small
        \begin{tikzpicture}[
            scale=0.90, transform shape,
            node distance=1.2cm and 1.8cm,
            font=\sffamily\small,
            % Process Styles
            base/.style={rectangle, draw, text centered, rounded corners, minimum height=1.1cm, line width=1pt},
            analysis/.style={base, draw=orange!80, fill=orange!5, text width=2.8cm},
            synthesis/.style={base, draw=blue!80, fill=blue!5, text width=2.8cm},
            patching/.style={base, draw=green!70!black, fill=green!5, text width=2.8cm},
            integration/.style={base, draw=purple!80, fill=purple!5, text width=2.5cm},
            % Batch Input Style (Stacked rectangles)
            batch/.style={
                rectangle, draw=gray!60, fill=gray!5, line width=0.8pt,
                rounded corners, minimum width=2.2cm, minimum height=1.1cm,
                copy shadow={shadow xshift=6pt, shadow yshift=-6pt, fill=white, draw=gray!40},
                copy shadow={shadow xshift=3pt, shadow yshift=-3pt, fill=white, draw=gray!40},
            },
            % Output Style
            io/.style={base, draw=gray!80, fill=yellow!5, minimum width=2.2cm, line width=0.8pt},
            decision/.style={diamond, draw=purple!80, fill=purple!5, text width=1.8cm, text centered, inner sep=0pt, line width=1pt},
            arrow/.style={-{Stealth[scale=1.2]}, line width=1pt, draw=black!70},
            label_text/.style={text centered, font=\sffamily\scriptsize\itshape, text=black!70}
        ]

            % --- Row 1: Distillation Core ---
            % The input is now a batch of tasks
            \node [batch] (input) {Failed Tasks $T$};
            
            \node [analysis, right=1.0cm of input] (analyze) {
                \textbf{Failure Analysis}\\
                \scriptsize Diagnose Error
            };
            \node [synthesis, right=2.1cm of analyze] (synth) {
                \textbf{Strategy Synthesis}\\
                \scriptsize Abstract Directives $\delta$
            };

            % --- Row 2: Policy Repair & Verification ---
            \node [patching, below=1.8cm of analyze] (gen) {
                \textbf{Patch Generation}\\
                \scriptsize Minimal Code $p$
            };
            \node [integration, right=2.0cm of gen] (integ) {
                \textbf{IntegratePatch}\\
                \scriptsize Runtime Update
            };
            \node [decision, right=1cm of integ] (check) {Executable?};
            
            % Final Output (Distinct ellipse shape)
            \node [io, below=0.8cm of check, node distance=1.5cm] (output) {Update Policy $\pi_{t+1}$};

            % --- Connections ---
            \draw [arrow] (input) -- (analyze);
            \draw [arrow] (analyze) -- node[above, label_text] {Reflections $A_i$} (synth);
            
            \draw [arrow] (synth.south) -- ++(0, -0.6) -| node[above, pos=0.25, label_text] {Strategy Directives $\delta_j$} (gen.north);
            
            \draw [arrow] (gen) -- node[above, label_text] {Code Patch $p_j$} (integ);
            \draw [arrow] (integ) -- node[above, label_text] {$\pi_{t+1}$} (check);
            
            \draw [arrow] (check) -- node[left, pos=0.1, label_text] {Yes} (output);
            
            \draw [arrow] (check.north) -- ++(0, 0.5) -| node[above, pos=0.25, label_text] {No (Retry $n$ times)} (integ.north);

            % --- Shaded Experience Abstraction & Repair Phase ---
            \begin{scope}[on background layer]
                \node[fill=blue!2, draw=blue!20, dashed, rounded corners, line width=1pt,
                      inner ysep=8pt, inner xsep=15pt, 
                      fit=(analyze) (synth) (gen) (integ) (check),
                      label={[anchor=south, blue!60, font=\bfseries]above:Experience Abstraction \& Policy Repair Phase}] (bg) {};
            \end{scope}

        \end{tikzpicture}
        \subcaption{Policy repair module (refer Algorithm \ref{alg:repair policy}).}
        \label{fig:polaris_algo_2}
    \end{subfigure}

    \caption{\textbf{Architectural overview of POLARIS.} \textbf{(a) Recursive self-improvement cycle:} The agent selects actions based on its policy and goals, storing outputs and reasoning traces in \texttt{Memory}. Evaluation collects $N$ failed tasks from the validation set, triggering the Policy Repair module. \textbf{(b) Policy repair cycle:} Through experience abstraction, the agent performs \textit{Failure Analysis} on the $N$ tasks, distills reusable strategies in \textit{Strategy Synthesis}, generates minimal code patches, and integrates them into the current policy. A candidate version is execution-checked, and if valid, applied via runtime code mutation.}

\label{fig:polaris_combined}
\end{figure*}
\endgroup
\vspace{-10pt}

A natural way to make improvements persistent is to treat the agent's policy as an explicit object that can be inspected and revised. Gödel Agents \cite{yin2024g} formalize this idea as recursive self-improvement: the agent inspects its own policy and execution traces, and updates the policy in a tested loop (against a benchmark). The work by \cite{yin2024g} provides an LLM based practical framework which achieves recursive self-improvement with run-time code mutation. However, directly instantiating Gödel agent style self-improvement can be resource intensive. In our initial attempts to adapt the Gödel Agent framework to a 7B model, runs frequently failed due to out-of-memory and tool-call errors before completion of execution. A key reason is the context growth: the framework retains multiple validation samples and multiple prior evolution steps in memory to support reflection, which increases context length and computational overhead after each iteration. This motivates our approach.  

We introduce \textsc{Polaris} to make recursive policy repair feasible under the constraint of working with smaller models. \textsc{Polaris} performs policy repair via experience abstraction: failures are analyzed and generalized into compact reusable repair strategies, which are rendered as minimal code patches and integrated into the current policy with conservative checks and bounded retries. \textsc{Polaris} controls context growth, while retaining traceability of the learned updates, by limiting the number of failed examples that are retained in memory for purpose of reflection along with the reduced tool-call history.

\noindent\textbf{Contributions}.
\begin{itemize}[noitemsep,nolistsep,leftmargin=*]
    \item We introduce \textbf{\textsc{Polaris}}, a framework that transforms failures into validated policy updates through analysis, synthesis, abstraction, and repair.
    \item We highlight the challenges with Gödel Agent~\cite{yin2024g} for SLMs in resource-constrained setting. We demonstrate that \textbf{recursive self-improvement is viable with SLMs}, reducing dependency on very large LLMs.
    \item We empirically validate our approach on MGSM~\cite{shilanguage}, DROP~\cite{dua2019drop}, GPQA~\cite{rein2024gpqa}, and LitBench~\cite{fein2025litbench} demonstrating consistent performance gains and interpretable improvements in capabilities.
\end{itemize}

We emphasize that both the instantiation of the Gödel Agents framework \cite{yin2024g} and our implementation of the \textsc{Polaris} framework, which builds upon it, support runtime updates. This capability offers a significant practical advantage in post-deployment scenarios compared to hand-designed agents. Although neither the evaluation presented in this paper nor that in \cite{yin2024g} addresses open-ended learning, an implementation that enables in-situ updates is likely to be particularly valuable in interactive environments where tool interfaces, data formats, or operational constraints evolve over time. The overall architecture of the \textsc{Polaris} and its Policy repair mechanism are depicted in Figure~\ref{fig:polaris_combined}.

\section{Related Work}

The concept of recursive self-improvement has deep roots in AI theory. Good~\cite{good1966speculations} speculated on the possibility of an “intelligence explosion” initiated by systems capable of enhancing their own cognitive processes. Schmidhuber~\cite{schmidhuber2007godel} later formalized this notion in the Gödel Machine, a theoretical construct that can provably rewrite its own code if it can prove the modification improves its performance. 

While elegant, Gödel Machines remain largely aspirational: exhaustive proof search is computationally infeasible, and no practical instantiation has been achieved. Nevertheless, this line of work provides the conceptual foundation for subsequent explorations of self-improving agents. Most notably, the Gödel Agent framework \cite{yin2024g} demonstrates how large language models (LLMs) can engage in self-referential reasoning to repair and enhance their own policies. While these developments provide an important proof of concept, they remain largely tied to frontier-scale LLMs, leaving open the question of whether smaller models can sustain recursive improvement under resource constraints. Our work addresses this gap by extending self-referential frameworks to smaller language models (SLMs) such as Qwen2.5-7B, and by proposing a principled mechanism for experience abstraction and policy repair that enables continual self-improvement.

\begin{figure*}[t]
    \centering
    \begin{subfigure}{0.8\linewidth}
        \centering
        \includegraphics[width=\linewidth]{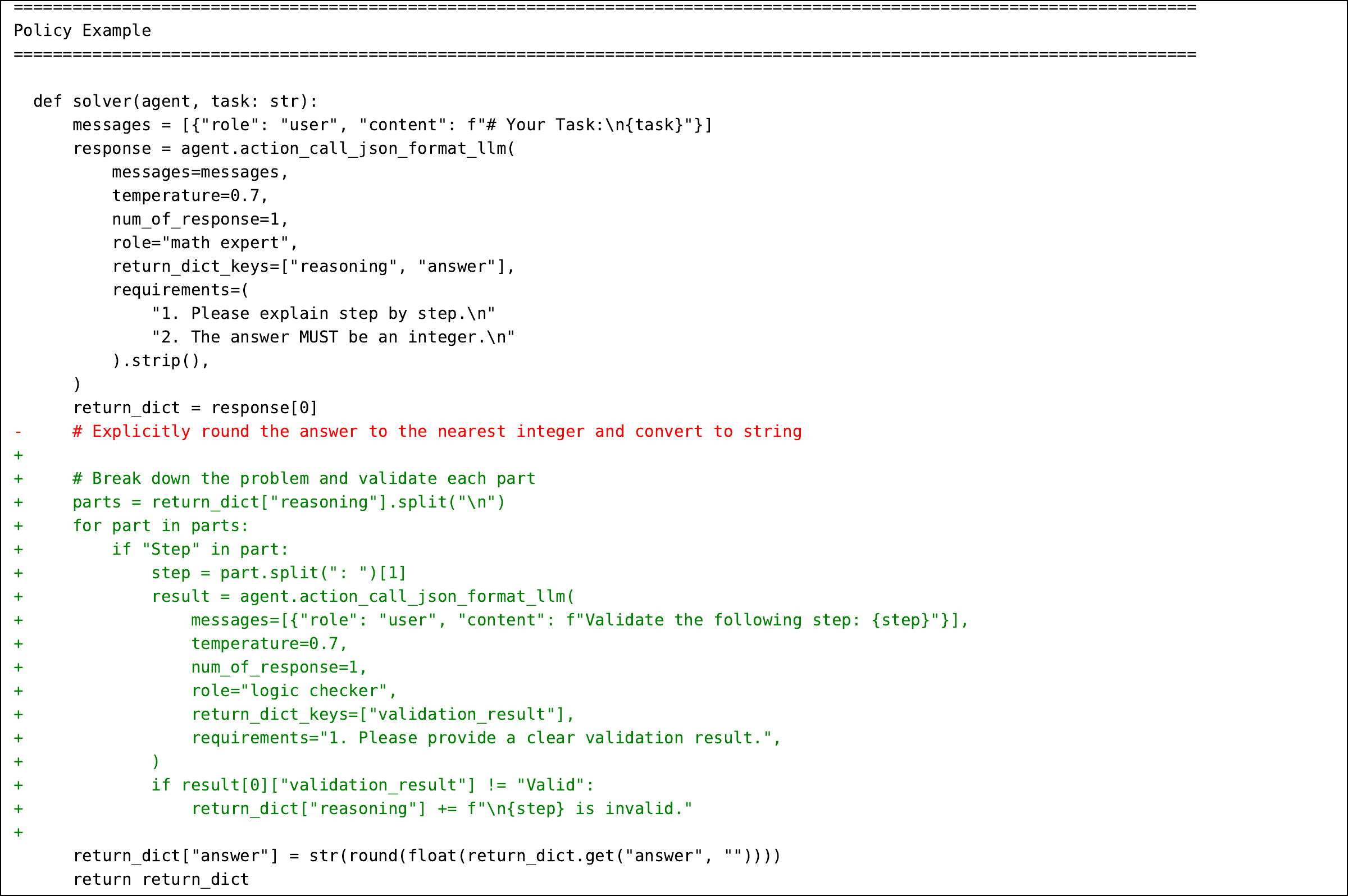}
        % \caption{}
    \end{subfigure}
    % \begin{subfigure}{0.5\linewidth}
    %     \centering
    %     \includegraphics[width=\linewidth]{images/policy_examples/drop_policy_diffs3.pdf}
    %     \caption{}
    % \end{subfigure}
    % \hfill
    \caption{Policy update example on the MGSM dataset from experiments with the Qwen2.5‑7B‑Instruct model. We highlight the updates in the current policy with respect to the previous policy using green color (new statements added) and red color (statements deleted). We observe the addition of the logic to break down the problem and validate each part while deleting the comment for post-processing the response.}
    \label{fig: policy_examples_mgsm1}

\end{figure*}

Research on reflection-driven language agents provides another strand of inspiration. Approaches such as ReAct \cite{yao2023react}, Reflexion \cite{shinn2023reflexion}, and Self-Refine \cite{madaan2023self} show that iterative feedback and self-critique can substantially improve reasoning and task performance. Extensions like CRITIC \cite{gou2023critic} and self-debugging strategies \cite{chen2023teaching} further emphasize the value of embedding correction mechanisms into the agent loop. These works, however, primarily focus on improving responses within tasks. \textsc{Polaris} builds on their insight but moves beyond single-instance correction by abstracting from task-level failures, synthesizing reusable strategies, and integrating them back into the policy to enable cumulative improvement.

A complementary line of work investigates direct editing of model representations. Techniques for localizing and modifying factual associations \cite{meng2022locating}, performing mass edits in transformers \cite{meng2023mass}, and leveraging task arithmetic \cite{ilharco2022editing} illustrate that targeted modifications can shift model behavior without retraining. Surveys such as Wang et al. \cite{wang2024knowledge} summarize this rapidly expanding literature. Compared to such parameter-centric approaches, \textsc{Polaris} adopts a higher-level repair process, emphasizing strategy abstraction and policy refinement rather than surgical edits to weights.

The idea of repair also resonates with research in automated program repair (APR), where systems diagnose errors and generate patches to improve external code bases. Classical bibliographies \cite{monperrus2018automatic} and recent surveys on LLM-driven APR \cite{zhang2024systematic} reveal striking parallels to self-improvement in language agents. Our contribution differs in that repair is applied not to external programs but to the agent’s own evolving policy, thereby blurring the line between debugging and learning.

We also note the emergence of evolutionary Gödel frameworks such as the Darwin Gödel Machine \citet{zhang2025darwin}, which perform open‑ended search by maintaining an archive of diverse agent variants and evaluating them through empirical performance on coding benchmarks.These evolutionary methods explore many candidate agents in parallel, in contrast to POLARIS, which focuses on iterative refinement of a single policy under tight compute and memory constraints. The approaches are complementary: population‑based evolutionary search suits high‑budget settings, while POLARIS targets resource‑constrained SLM deployments where single‑policy improvement is more practical.

Finally, our work is informed by the tradition of open-ended learning, which treats novelty, diversity, and complexity as drivers of continual progress. Theories of novelty search \cite{lehman2011abandoning,stanley2015greatness} and systems like POET \cite{wang2019paired} illustrate how adaptive agents coupled with evolving environments can yield unbounded improvement. Meta-learning \cite{finn2017model} and hierarchical reinforcement learning \cite{sutton1999between,bacon2017option} further demonstrate the importance of abstraction and reuse in sustaining adaptability. Classical open-ended learning pursues sustained novelty, typically via intrinsic objectives or co-evolving environments. \textsc{Polaris} is conceptually adjacent in its emphasis on cumulative abstraction and reuse, but our study is bounded to fixed task distributions and external evaluation (see Appendix, Section~\ref{sec: open_ended}, for a detailed discussion and positioning within the literature on open-ended exploration.).

\section{Our Approach}

\subsection{Gödel Agent for Recursive Self-Improvement}
Gödel Agent~\cite{yin2024g} introduces a self-referential framework that realizes recursive self-improvement in LLM-based agents. The framework enables an agent not only to modify its task-execution policy $\pi$ but also to revise the meta-level improvement logic $I$ that determines how these modifications are generated and applied. 
\SetKwProg{Fn}{Function}{:}{}
\SetKwComment{Comment}{\hfill$\triangleright$ }{}
\renewcommand{\baselinestretch}{1.0}\selectfont

\newcommand\mycommfont[1]{\small\ttfamily\textcolor{blue}{#1}}
\SetCommentSty{mycommfont}

\begin{algorithm*}[!tbh]
\small
    \caption{Recursive Self-Improvement of Gödel Agent}
    \label{alg:godel_agent_full}
    \begin{multicols}{2} % Specifies two columns
    \DontPrintSemicolon
     
    \KwIn{Initial agent policy $\pi_0$, initial decision function $f_0$, goal $g$, environment state $E$, action of the agent $a_i$, memory of the agent \texttt{Memory}, policy feedback $r$, policy performance assessment function \texttt{EVALUATE}, self code reading function \texttt{SELF\_INSPECT}, policy repair function \texttt{REPAIR\_POLICY}}
    \KwOut{Improved policy $\pi$, final agent state $s$}
    \BlankLine
    \Comment{Initialize empty set to store all the policy improvement strategies}
    $\mathcal{\delta} \leftarrow \emptyset$\;
    
    \Comment{Initialize Agent memory, used to store action taken and action result}
    $\texttt{Memory} \leftarrow \emptyset$\;
    
    \Comment{Introspect and retrieve agent's own code.}
    $s \leftarrow \texttt{SELF\_INSPECT}()$\;

    \Comment{Evaluate current policy, get the policy feedback $r$ and list of failed tasks $\mathcal{T}$}
    $r, \mathcal{T} \leftarrow \texttt{EVALUATE}(\pi_0, E)$\;
    
    % \Comment{Repair policy}
    $\textcolor{red}{\pi,s \leftarrow \texttt{REPAIR\_POLICY}(\pi, s, \mathcal{T}, a, r)}$\;
    
    \Comment{Perform recursive self-improvement.}
    $\pi, s \leftarrow \texttt{SELF\_IMPROVE}(\pi_0, s, r, g)$\;
    
    \KwRet{$\pi, s$}
    \BlankLine
     
    \Fn{\texttt{SELF\_IMPROVE}$(\pi, s, r, g)$}{
      $a_1, \dots, a_n \leftarrow f_0(\pi, s, r, g)$\;
      
      \For{$a_i$ in $a_1, \dots, a_n$}{
        $\pi, s, r \leftarrow \texttt{EXECUTE}(\pi, s, r, a_i, E)$\;
      }
      \KwRet{$\pi, s$}
    }
    \BlankLine
     
    \Fn{\texttt{EXECUTE}$(\pi, s, r, a, E)$}{
      \Switch{$a.\texttt{name}$}{
        \Case{\texttt{self\_state}}{
          $s \leftarrow \texttt{SELF\_INSPECT}()$\;
         
          $\texttt{Memory}.append(a, s)$\;
        }
        
        \Case{\texttt{interact}}{
            $r, \mathcal{T} \leftarrow \texttt{EVALUATE}(\pi, E)$\;
            
            $\texttt{Memory}.append(a, r)$\;
            
            % \Comment{Repair policy}
            $\textcolor{red}{\pi,s \leftarrow \texttt{REPAIR\_POLICY}(\pi, s, \mathcal{T}, a, r)}$\;
        }
        
        \Case{\texttt{self\_update}}{
          $\pi, s \leftarrow a.\texttt{code}$\;
          
          $\texttt{Memory}.append(a, s)$\;
        }
        \Case{\texttt{continue\_improve}}{
          $\pi, s \leftarrow \texttt{SELF\_IMPROVE}(\pi, s, r, g)$\;
          
          $\texttt{Memory}.append(a, s)$\;
        }
      }
      \KwRet{$\pi, s, r$}
    }
    \end{multicols} 
    \BlankLine
\end{algorithm*}

\setcounter{AlgoLine}{0}
\begin{algorithm*}[!tbh]
\small
    \caption{Updating Agent's Policy with \textsc{Polaris}}
    \label{alg:repair policy}
    \begin{multicols}{2} % Specifies two columns
        \DontPrintSemicolon
        \KwIn{Current agent policy $\pi_t$, current agent state $s$, list of failed task samples $\mathcal{T}$, agent action $a$, policy feedback $r$ }
        \KwOut{Improved policy $\pi_{t+1}$, agent state $s$}
        \BlankLine
        
        \Fn{\texttt{REPAIR\_POLICY}$(\pi_t, s, \mathcal{T}, a, r)$}{
            \Comment{Initialize empty set to store Self-Reflections of each task}
             $A \leftarrow \emptyset$\; 
              
            \ForEach{$\tau_i$ in $\mathcal{T}$}{
                \Comment{Get self-reflection $A_i$ for each task}
                $A_i \leftarrow \texttt{AnalyzeFailure}(\pi_t, s, \tau_i)$\;
                
                $A \leftarrow A \cup \{A_i\}$\;
              }
             
            \Comment{Get policy improvement strategies from $A$}
          
            $\mathcal{\delta} \leftarrow \mathcal{\delta} \cup  \{\texttt{StrategySynthesis}(\pi_t, s, A)\}$\;
          
            \Comment{Get code patch for each strategy}
            $p \leftarrow \texttt{PatchGeneration}(\pi_t, s, \mathcal{\delta})$\;
          
            \Comment{Number of retries left}
            $n \leftarrow 3$
        
            \Comment{Integrate patches in current policy $\pi_t$}
            $\pi_{t+1}, s \leftarrow \texttt{IntegratePatch}(\pi_t, s, p, n, a, r)$\;
         
            \KwRet{$\pi_{t+1}, s$}
        }

        \Fn{\texttt{IntegratePatch}$(\pi_t, s, p, n, a, r)$}{
            \Comment{Get the updated policy by integrating patches $p$ }
            
            $\pi_{t+1} \leftarrow \texttt{UpdatePolicy}(\pi_t, s, p)$\;
            \Comment{Retry, if the updated policy is not executable}
            
            \If{ $\pi_{t+1}$ exists }
            {
                \If {$\pi_{t+1}$ is not executable \& $n$>0}
                {   \Comment{If failed, retry and decrease the number of retries left}
                    $\pi_{t+1}, s \leftarrow \texttt{IntegratePatch}(\pi_t, s, p,n-1)$\;
                }
                \Comment{Push updated policy into the codebase using $\texttt{self\_update}$ and update the memory}
        
                $\texttt{Memory}.append(\texttt{self\_update}(\pi_{t+1}, s))$\;  
        
                \KwRet{$\pi_{t+1}, s$}
            }

            \Comment{Upon failing to update the policy, append code patches into memory}
            $\texttt{Memory}.append(a, r+p)$\;
            
            \Comment{Continue to evolve}
            
        }
    \end{multicols}
    \BlankLine
\end{algorithm*}

Given an environment $\mathcal{E}$ and a utility function $U(\mathcal{E}, \pi)$, the agent repeatedly executes, evaluates, and refines its own policy and improvement logic through four key procedures.:

\begin{enumerate}[noitemsep,nolistsep,leftmargin=*]
\item \textbf{Introspection (Self-Inspect):} The agent analyzes its internal architecture, including code modules, reasoning traces, and historical performance data. This process yields an explicit representation of the agent’s current capabilities and limitations.
\item \textbf{Execution (Interact):} During execution, the policy $\pi$ operates as an LLM-based reasoning and acting module that engages with the environment through natural language interactions, tool use, and task-specific actions. The agent records both its intermediate reasoning traces and environmental responses, which serve as empirical evidence for evaluating and refining its subsequent behavior.
\item \textbf{Self-Modification (Self-Improve):} Using the improvement logic $I$, the agent evaluates its performance and proposes candidate code edits or rewrites. These modifications may target the policy $\pi$ to enhance problem-solving behavior or the improvement logic $I$ itself to refine the way updates are reasoned about. Large language models serve as the generative engine for proposing, critiquing, and verifying such modifications.
\item \textbf{Recursive Continuation (Continue-Improve):} After each modification is integrated, the agent re-enters the introspection phase. This recursive loop allows both $\pi$ and $I$ to evolve jointly, producing progressively more abstract forms of self-repair and adaptation.
\end{enumerate}

A central technical innovation in Gödel Agent is its use of \textit{runtime code mutation}, implemented through mechanisms that enable modification of executable components during operation. This capability allows the agent to test, validate, and revert modifications dynamically without full retraining, supporting stable iterative improvement.

% Together, these procedures operationalize the core principles of recursive self-improvement. Through \textbf{self-reference}, the agent maintains explicit representations of its own structure and reasoning process, enabling it to analyze and modify its internal logic.  Through \textbf{self-modification}, both the task policy and the improvement logic are subject to revision via generated code updates that adjust behavior or meta-reasoning mechanisms. Finally, through \textbf{self-validation}, each proposed modification is empirically tested against prior performance, ensuring that only beneficial updates are retained and integrated into the evolving agent.

% Empirical studies show that Gödel Agent can perform multi-round reasoning and self-modification with large models such as GPT-3.5 and GPT-4, demonstrating stable recursive adaptation. However, the framework remains computationally intensive and sensitive to initialization, motivating extensions like \textsc{Polaris}, which aim to adapt these principles for smaller models through structured experience and policy repair.

\subsection{\textsc{Polaris}: Policy Repair through Experience Abstraction}

\textsc{Polaris} implements recursive self-improvement in small language models through a structured cycle of reflection, abstraction, and repair, converting execution failures into validated code-level updates while preserving full traceability in the agent’s \texttt{Memory}. Each cycle follows the operators defined in \textbf{Algorithms~\ref{alg:godel_agent_full} and~\ref{alg:repair policy}} and the prompt templates shown in \textbf{Figures~\ref{fig:analyze_failures_prompt}--\ref{fig:update_policy_prompt}}. The agent executes a mutable policy $\pi_t$ on a validation set $\mathcal{D}$, records its behavior and outcomes, and integrates validated updates to yield a refined policy $\pi_{t+1}=\pi_t\oplus\Delta\pi$.

\paragraph{Failure Analysis.} 
Executing $\pi_t$ on $\mathcal{D}$ produces a set of failed instances \[
\mathcal{T}=\{\tau_i\}
\]
where each $\tau_i$ contains the input, the agent's reasoning trace, the predicted output, and the reference answer. For every $\tau_i$, the agent invokes \texttt{AnalyzeFailure} (Algorithm~\ref{alg:repair policy}; Figure~\ref{fig:analyze_failures_prompt}), a self-reflection operator that generates a structured record
\[
A_i=(\textit{diagnosis}_i, \textit{revision}_i, \textit{prevention}_i)
\]
The \textit{diagnosis} identifies the cause of error in the policy's reasoning or control flow, the \textit{revision plan} proposes targeted adjustments at the code or rule level, and the \textit{prevention rule} generalizes these insights for future iterations. Each reflection $A_i$ is appended to \texttt{Memory}, forming a repository of interpretable experience from which higher-level repair strategies are derived.

\paragraph{Strategy Synthesis.}
The \texttt{StrategySynthesis} operator (Algorithm~~\ref{alg:repair policy}; Figure~\ref{fig:strategy_prmpt}) abstracts across reflections $A=\{A_i\}$ to produce a compact set of reusable directives
\[
\delta=\{\delta_j\}
\]
Each $\delta_j$ captures a general repair principle such as decomposition, normalization, or control-flow adjustment that can resolve multiple failures. The prompt enforces novelty with respect to strategies stored in \texttt{Memory} and limits the agent to one or two well-formed strategies per cycle. By compressing instance-specific reflections into transferable repair abstractions, \textsc{Polaris} transforms episodic feedback into policy-level adaptation.

\paragraph{Patch Generation.}
For each synthesized strategy $\delta_j$, the \texttt{PatchGeneration} operator (Algorithm~~\ref{alg:repair policy}; Figure~\ref{fig:code_patch_prompt}) instantiates a minimal code patch $p_j$. Each patch modifies only the lines required to implement $\delta_j$ and excludes any explanatory text. A lightweight validator checks syntax and formatting before a patch enters integration. The resulting patch set is denoted $\mathcal{P}=\{p_j\}$. Emphasizing locality and minimality ensures that every modification remains interpretable and that the agent's policy evolves through small, verifiable updates.

\paragraph{Patch Integration.}
Patch integration follows \textbf{Algorithm~\ref{alg:repair policy}} and Figure~\ref{fig:update_policy_prompt}. Each patch in $\mathcal{P}$ is applied through the \texttt{UpdatePolicy}$(\pi_t, s, p)$ procedure to generate a temporary policy candidate. Integration is verified through syntactic and execution checks rather than direct performance evaluation. If a patch fails, the agent retries up to a fixed bound (three times by default). Persistent failures result in the patch and its context being archived in \texttt{Memory} for potential later analysis. After integration, the updated policy $\pi_{t+1}$ proceeds to the next validation phase, where its performance effects are naturally observed. \texttt{Memory} retains all artifacts from the cycle, including reflections, strategies, patches, and integration results, providing continuity and preventing redundant proposals.

\section{Experiments}
% \subsection{Experimental Setup}
We evaluate \textsc{Polaris} on MGSM, DROP, and GPQA, covering arithmetic, discrete, and advanced factual reasoning. Additionally, we include LitBench, an open-domain benchmark for creative writing that tests stylistic preference modeling, narrative coherence, and open-ended reasoning.
% We evaluate our framework on three reasoning and comprehension benchmarks: MGSM, DROP, and GPQA, chosen to cover diverse reasoning modalities and difficulty levels. MGSM tests arithmetic reasoning and compositional generalization, DROP evaluates discrete reasoning over natural text, and GPQA assesses advanced factual and conceptual reasoning. In addition, we include LitBench, a recently proposed open-domain preference evaluation benchmark for creative writing. LitBench consists of creative-writing prompts paired with two candidate responses, one preferred by humans and one rejected, and evaluates a model’s ability to capture stylistic preferences, narrative coherence, and open-ended reasoning beyond factual QA. 

For MGSM and DROP, we use 50 validation and 250 test samples; for GPQA, 20 validation and 100 test samples; and for LitBench, 20 validation and 250 test samples. For MGSM and GPQA, we report accuracy with 95\% bootstrap confidence intervals, while for DROP we report macro F1 score due to its span-selection format. For LitBench, we report accuracy based on preferred-response selection.

\begin{table}[!tbh] 

    \begin{subtable}[t]{\linewidth} 

        \small 

        \centering 

        \resizebox{\linewidth}{!}{ 

        \begin{tabular}{|l|c|c|c|c|} 

        \hline            & \multicolumn{1}{l|}{\textbf{MGSM}} & \multicolumn{1}{l|}{\textbf{DROP}} & \multicolumn{1}{l|}{\textbf{GPQA}} & \multicolumn{1}{l|}{\textbf{LitBench}}\\ \hline 

        \textbf{Successful}     & 0                                  & 1                                  & 0           & 0                      \\ \hline 

        \textbf{No improvement} & 5                                  & 2                                  & 5                   & 1                      \\ \hline 

        \textbf{Unsuccessful}   & 0                                  & 2                                  & 0                       & 4                  \\ \hline 

        \textbf{Total}          & 5                                 & 5                                 & 5                          & 5              \\ \hline 

        \end{tabular}} 

        \caption{$k$=3} 

    \end{subtable} 

    \begin{subtable}[t]{\linewidth} 

        \small 

        \centering 

       \resizebox{\linewidth}{!}{ 

       \begin{tabular}{|l|c|c|c|c|} 

        \hline            & \textbf{MGSM} & \textbf{DROP} & \textbf{GPQA} & \textbf{LitBench}\\ \hline 

        \textbf{Successful}      & 0             & 0             & 0        & 0    \\ \hline 

        \textbf{No improvement} & 0             & 0             & 0          & 2  \\ \hline 

        \textbf{Unsuccessful}   & 5             & 5             & 5          & 3  \\ \hline 

        \textbf{Total}          & 5             & 5             & 5        & 5    \\ \hline 

        \end{tabular}} 

        \caption{$k$=5} 

    \end{subtable} 

    \caption{A summary of various runs of Gödel Agent~\citep{yin2024g} using Qwen2.5‑7B‑Instruct model in two different settings: (a) $k$ = 3, three prior tool-call messages in memory (instead of 10), and (b) $k$ = 5, five prior tool-call messages in memory (instead of 10).}
    \label{tab:original_godel_agent}
\end{table}

% \begin{table}[]
% \begin{tabular}{|l|c|c|}
% \hline
% \textbf{N=3} & \textbf{\textsc{Polaris}} & \textbf{Gödel Agent} \\ \hline
% Successful                                                  & 4                                                               & 0                             \\ \hline
% No improvement                                              & 0                                                               & 1                             \\ \hline
% Unsuccessful                                                & 1                                                               & 4                             \\ \hline
% Total                                                       & 5                                                               & 5                             \\ \hline
% \end{tabular}
% \caption{Comparison of \textsc{Polaris} with Gödel Agent~\citep{yin2024g}.}
% \label{tab:original_godel_agent}
% \end{table}

\begin{table}[!tbh] 

    \begin{subtable}[t]{\linewidth} 

        \small 

        \centering 

        \resizebox{\linewidth}{!}{ 

        \begin{tabular}{|l|c|c|c|c|} 

        \hline            & \multicolumn{1}{l|}{\textbf{MGSM}} & \multicolumn{1}{l|}{\textbf{DROP}} & \multicolumn{1}{l|}{\textbf{GPQA}} & \multicolumn{1}{l|}{\textbf{LitBench}}\\ \hline 

        \textbf{Successful}     & 5                                  & 3                                  & 4           & 6                      \\ \hline 

        \textbf{No improvement} & 1                                  & 2                                  & 4                   & 1                      \\ \hline 

        \textbf{Unsuccessful}   & 4                                  & 5                                  & 2                       & 3                  \\ \hline 

        \textbf{Total}          & 10                                 & 10                                 & 10                          & 10              \\ \hline 

        \end{tabular}} 

        \caption{$N$=3} 

    \end{subtable} 

    \begin{subtable}[t]{\linewidth} 

        \small 

        \centering 

       \resizebox{\linewidth}{!}{ 

       \begin{tabular}{|l|c|c|c|c|} 

        \hline            & \textbf{MGSM} & \textbf{DROP} & \textbf{GPQA} & \textbf{LitBench}\\ \hline 

        \textbf{Successful}      & 4             & 3             & 5        & 5    \\ \hline 

        \textbf{No improvement} & 1             & 1             & 2          & 1  \\ \hline 

        \textbf{Unsuccessful}   & 5             & 6             & 3          & 4  \\ \hline 

        \textbf{Total}          & 10             & 10             & 10        & 10    \\ \hline 

        \end{tabular}} 

        \caption{$N$=5} 

    \end{subtable} 

    \caption{A summary of various runs of \textsc{Polaris} on datasets using the Qwen2.5‑7B‑Instruct model. For policy repair and experience abstraction, we consider a set of $N$ failed instances from the validation set of each dataset. ($k$=6, six prior tool-call messages in memory)} 

    \label{tab: summary_combined} 

\end{table}

All experiments are conducted using \textsc{Qwen2.5-7B-Instruct}~\cite{qwen2.5} on two NVIDIA V100 GPUs (32\,GB each). Each run is allowed to evolve autonomously for 10~hours, rather than a fixed number of steps, to observe natural variations in the number of policy-repair iterations. We record all intermediate reflections and policy modifications for analysis. We define a hyperparameter \textbf{$N$, the number of failed-task samples in the validation set} that are used for introspective reflection. We test with $N\!=\!3$ and $N\!=\!5$ to study the trade-off between reflection depth and stability.
 
Initial attempts to adapt the Gödel Agent framework to SLMs led to frequent out-of-memory (OOM) and tool-call errors before completing the 10-hour execution window. This issue stems from the design of the Gödel Agent, which relies on large-context models. After each policy evaluation, the framework stores 20 validation samples, including task inputs, model responses, and feedback, in the agent's memory. It also retains the last ten evolution steps to maintain historical reasoning traces. While this design enriches the context for reflection, it rapidly increases context length after each iteration, significantly raising computational overhead. To address this challenge, the experience abstraction mechanism requires fewer validation samples ($N$) for meta-reasoning along with reduced number of messages in the memory i.e., six instead of ten.

To ensure structured outputs, we employ one-shot prompting and a lightweight \emph{helper function} that enforces valid JSON output during evaluation. This helper does not interfere with reasoning or evolution processes. We adapt the goal prompt of the agent from~\cite{yin2024g} with the requirements for our experimental setting (refer to Section \ref{sec: additional_experimental_details} in the Appendix).

We perform ten independent runs on each dataset. We further categorize these runs as: (1)~\textbf{Successful}— policy update improves test performance; (2)~\textbf{No Improvement}— policy update succeeds but yields no meaningful performance gain, defined as an accuracy increase corresponding to $\leq 2$ additional correctly solved samples (i.e., $\leq 0.8$\%); and (3)~\textbf{Unsuccessful}— run fails due to OOM, infinite loops, or hallucinated tool calls. Furthermore, we consider two baselines:
\begin{enumerate}[noitemsep,nolistsep,leftmargin=*]
    \item \textbf{Chain-of-Thought Self-Consistency (COT-SC)}~\citep{wangself}: Five reasoning paths are sampled per query, and the most frequent answer is chosen. We use the same validation and test splits as our experiments to report the performance. It is one of the best performing baselines reported in ~\citep{yin2024g}.
    
    \item \textbf{Gödel Agent}~\citep{yin2024g}: Direct replication with Qwen2.5-7B-Instruct led to repeated OOM failures before 10~hours. In Table \ref{tab:original_godel_agent}, we present a summary of various runs with reduced number of tool-call messages in the history. We observe that all runs fail due to memory constraints, resulting in out-of-memory errors for $k$ = 5. We observe fewer OOM errors with $k$ = 3 and three prior tool-call messages in memory. However, as the context length is very short, the agent fails to improve over iterations and gets stuck in repetitive and hallucinated tool calls. Further decreasing the context to accommodate memory constraints would lead to highly uncertain and non-targeted behaviour of the Gödel agent. Hence, a trivial adaptation of prior work on Gödel Agent is infeasible under resource constraints.
\end{enumerate}
 
\begin{figure*}[!tbh]
    \centering
    \begin{subfigure}{0.48\linewidth}
        \centering
        \includegraphics[width=\linewidth]{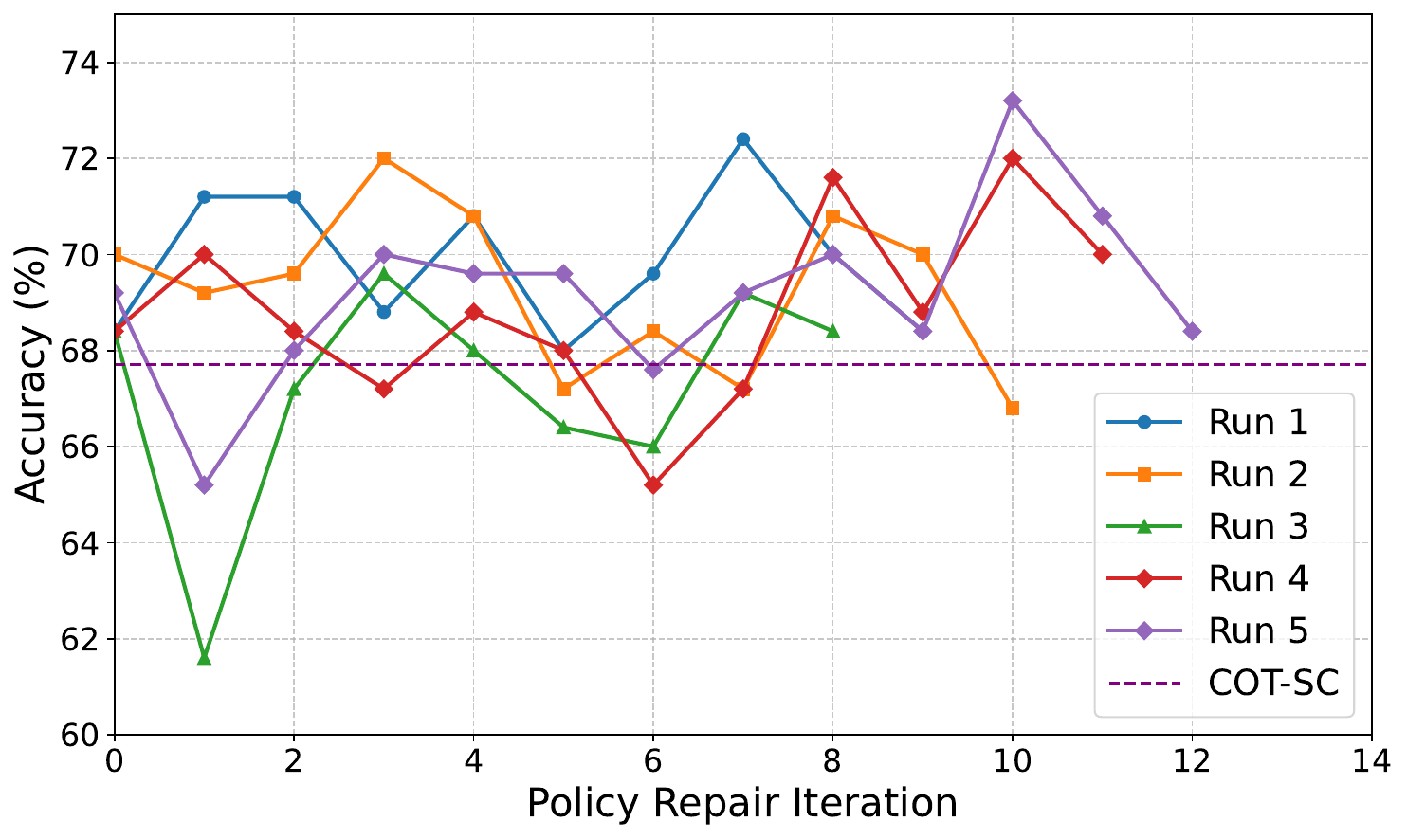}
        \caption{MGSM}
    \end{subfigure}
    \begin{subfigure}{0.48\linewidth}
        \centering
        \includegraphics[width=\linewidth]{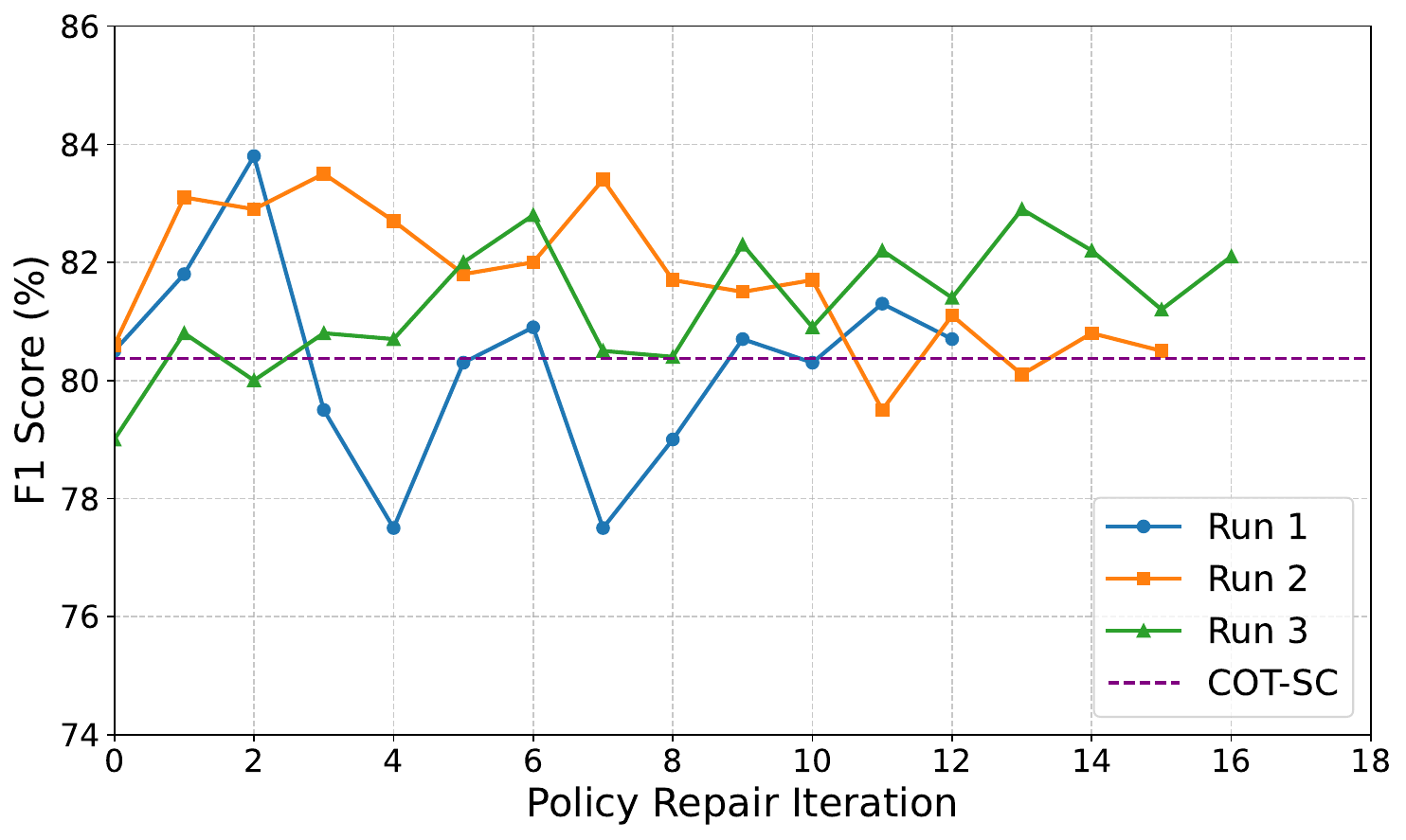}
        \caption{DROP}
    \end{subfigure}
    \begin{subfigure}{0.48\linewidth}
        \centering
        \includegraphics[width=\linewidth]{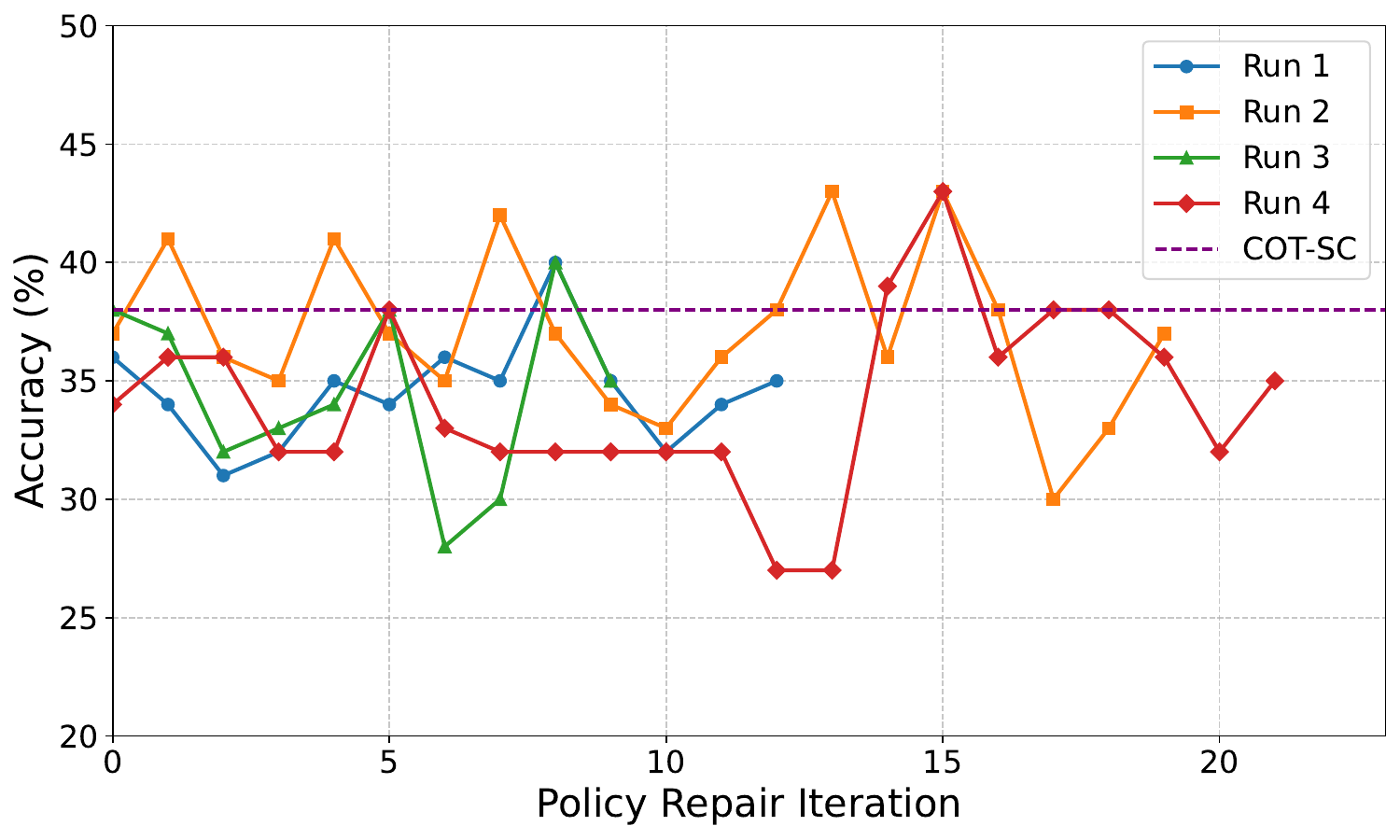}
        \caption{GPQA}
    \end{subfigure}
    \begin{subfigure}{0.48\linewidth}
        \centering
        \includegraphics[width=\linewidth]{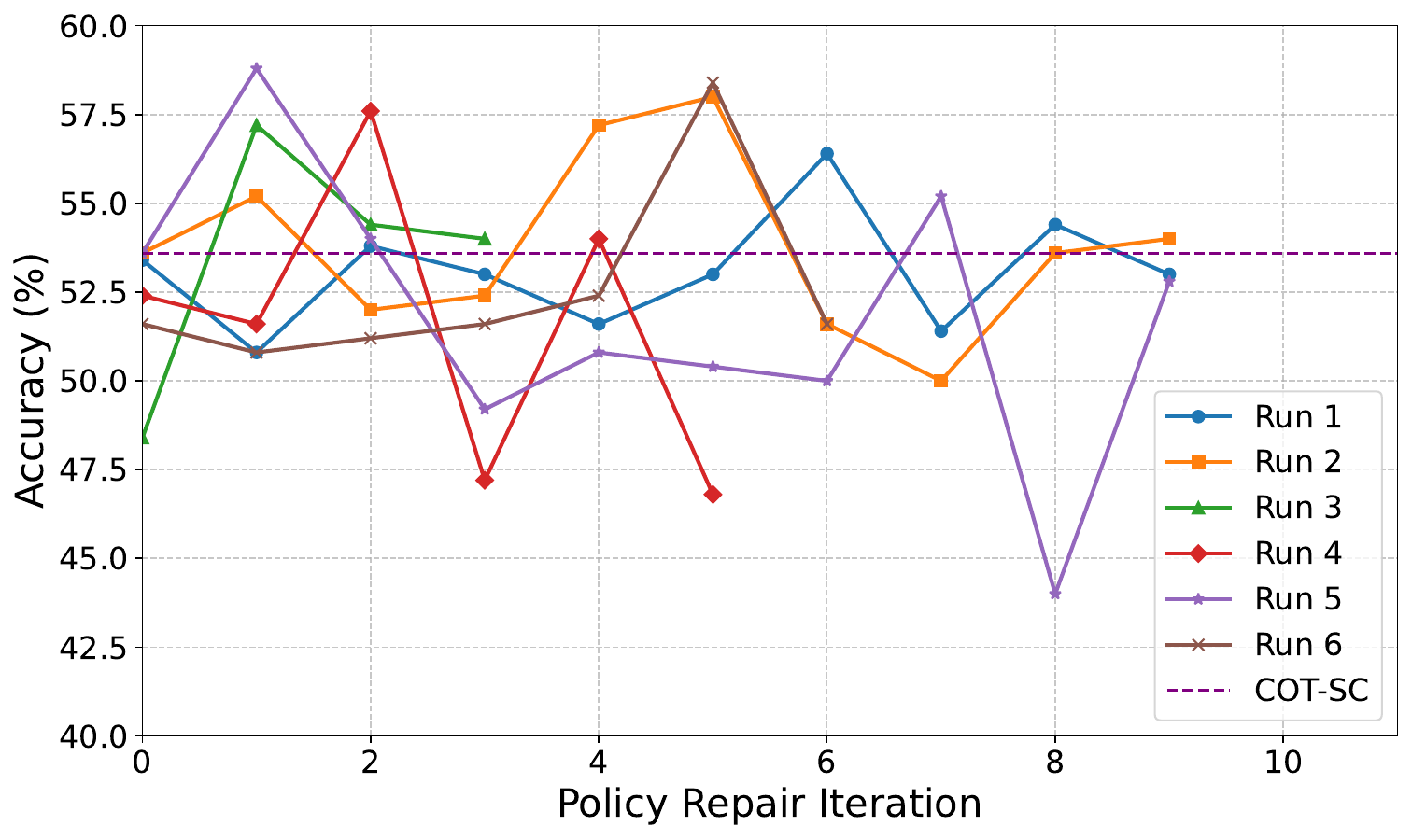}
        \caption{LitBench}
    \end{subfigure}
    % \hfill
    \caption{Successful evolution runs of \textsc{Polaris} with performance improvement compared to the base policy and COT-SC. Policy Repair Iteration 0 shows the performance with the base policy. For policy repair and experience abstraction, we consider a set of three failed instances from the validation set of each dataset ($N$=3). Experiments conducted with the Qwen2.5‑7B‑Instruct model.}
    \label{fig: results_exp_set_3}
\end{figure*}
 
% Despite occasional instability, these results confirm that recursive self-evolution is achievable with small models under modest compute.

 % The distribution is summarized in Table %~\ref{tab:summary_exp3}.
 
% \subsection{Baselines}
\section{Results and Analysis}
\textbf{Self-evolution under constrained setting is challenging}: We summarize the runs of \textsc{Polaris} in Table~\ref{tab: summary_combined}. In Figure~\ref{fig: results_exp_set_3} and the Appendix, we report performance across policy repair iterations alongside baselines. Evolution on MGSM, GPQA, and LitBench achieves a higher success rate than DROP, likely due to DROP’s larger context size, which increases susceptibility to out-of-memory (OOM) errors. Performance variability is greater for $N$=3 than $N$=5, and experience abstractions become more generic as $N$ grows. These trends highlight the difficulty SLMs face in abstracting strategies from diverse self-reflections. A detailed analysis of failure modes reveals two dominant factors: (i) \textbf{limited meta-reasoning capability}, where SLMs fail to diagnose failure causes and repair policies, leading to noisy, non-progressive corrections; and (ii) \textbf{poor tool-calling capability and OOM errors}, driven by large accumulated context, hallucinated tool calls, redundant evaluations, or irrecoverable policy adjustments, consistent with prior reports on SLM limitations~\cite{senel2025comparative, subramanian2025small}. Initial experiments showed that models with comparable parameter counts such as DeepSeek-Coder-6.7B-Instruct \citep{deepseek_coder_6_7b_instruct}, 
Llama-3.1-8B-Instruct \citep{llama_3_1_8b_instruct}, 
and Mistral-7B-Instruct-v0.3 \citep{mistral_7b_instruct_v0_3} but with limited tool‑calling and meta‑reasoning capabilities consistently failed, underscoring the challenges faced by SLMs. Example failure traces for these models are provided in Figures~\ref{fig: mistral_example}–\ref{fig: llama_example} in the Appendix.

% We provide a summary of various runs of \textsc{Polaris} in Table~\ref{tab: summary_combined}. In Figure \ref{fig: results_exp_set_3} and the Appendix, we report the performance across policy repair iterations along with the baselines. We observe that evolution on MGSM and GPQA yields a higher rate of successful runs compared to the DROP dataset. This disparity may be attributed to the larger context size present in the DROP dataset, which can lead to OOM errors. Additionally, we note a greater degree of variation in performance for $N$=3 compared to $N$=5. Alongside, we observe that the experience abstraction tends to become more generic as $N$ increases to 5, in contrast to $N$=3. These insights underscore the challenges faced by SLMs in effectively abstracting strategies from a larger and more diverse set of self-reflections on failed tasks.

% Our observations indicate that the success rate of runs is approximately 30-40\% with N=3 and 40-60\% with N=5.

% We present the successful evolution runs in Figure~\ref{fig: results_exp_set_3} and the Appendix. 

\textbf{Policy repair through experience abstraction tends to capture solution complexity:} We illustrate policy updates across datasets, showing how repairs introduce diverse enhancements such as complex instructions, post-processing steps, conditional logic, exception handling, and context-aware validation (Figures~\ref{fig: mgsm_example}–\ref{fig: policy_examples_litbench2} in the Appendix). An example of such a policy update on the MGSM dataset is shown in Figure~\ref{fig: policy_examples_mgsm1}, where POLARIS introduces decomposition and validation logic while removing outdated post‑processing steps. These observations highlight the role of experience abstraction in enabling nuanced, task-specific policy refinements. Additionally, the number of repair iterations varies across datasets, underscoring the need for adaptive, localized behavior in self-improving agents.

% We present various examples of experience abstraction and policy repair in Figure \ref{fig: teaser} and the Appendix (refer Figures \ref{fig: mgsm_teaser} and \ref{fig: gpqa_example}). Furthermore, 

% 1. Discuss about the table. what we did and what was our observation, challenges of SLM

% 2. Discuss the graphs. observation ?, how is the performance improvement ? max performance jump across datasets.
%3. what type of updates occur in diff datasets, refer to teaser figure.
%4. comment on why policy repair number is different in runs.
%5 avg performance improvement across all the runs for each dataset.
%6 Comparing exp set 3 and 5, effect of increasing experience set size.

% We first establish the baseline performance of the \textsc{Qwen2.5-7B-Instruct} model under the Chain-of-Thought Self-Consistency (COT-SC) prompting setup. 

\textbf{Non-monotonic but consistent performance gain:} We observe that the agent can recover from local performance minima across datasets (see Appendix, Section~\ref{sec: variance}, for a discussion on non-monotonic behavior). To quantify gains, we report the maximum relative improvement of our self-evolution framework over the COT-SC baseline across successful runs. For $N=3$, improvements include +4.0\% on MGSM, +3.9\% on DROP, +9.0\% on GPQA, and +8.8\% on LitBench. At $N=5$, trends remain consistent with +5.7\% on DROP, +3.6\% on MGSM, +9.0\% on GPQA, and +5.2\% on LitBench. These results underscore the effectiveness of \textsc{Polaris} in resource-constrained settings.

% The model attains 67.7\% accuracy on \textsc{MGSM}, 80.37\% F1 score on \textsc{DROP}, and 38.0\% accuracy on \textsc{GPQA} with the COT-SC baseline.
% that recursive self-evolution is achievable even for advanced reasoning tasks under small model constraints.

\textbf{Unsuccessful runs and the utility of \textsc{Polaris}:} Unsuccessful runs were caused by infrastructure issues such as GPU memory exhaustion, blocked library calls, or hallucinated tool invocations triggered by long debugging traces. These hallucinations typically occur when the agent repeatedly tries to fix a persistent error and the context becomes saturated with verbose logs. These failures reflect limits of our experimental setup rather than weaknesses in the experience‑abstraction mechanism. We intentionally omitted guardrails such as rollbacks, stronger static validation, and long‑term memory stabilization to observe the raw dynamics of experience‑driven policy evolution. Adding these stabilizers is a promising direction for reducing failure modes in future versions of \textsc{Polaris}.

\textbf{Impact of base model capability:} We ran an additional experiment with the \texttt{Qwen3‑8B} model \citep{qwen3_8b} using $N = 3$. Table~\ref{tab:qwen3-8b-iterations} summarizes all runs, and Figure~\ref{fig:qwen3_results_exp_set_3} and Figure~\ref{fig:qwen3_results_exp_set_5} shows the successful evolution trajectories. Compared to Qwen2.5‑7B‑Instruct, \texttt{Qwen3‑8B} is slightly larger and includes a native “thinking’’ mode that strengthens its reasoning ability. This leads to a more number of successful runs and more stable improvement across datasets, indicating that as compact models advance, \textsc{Polaris} will become increasingly effective for real‑world use. Under the same experimental constraints, however, Qwen3‑8B completes fewer iterations. Its default thinking mode generates far more tokens per call, increasing memory use and computation time. Although thinking is disabled during evaluation for fairness, it remains active during tool‑use and policy‑repair. The resulting overhead effectively reduces the iteration budget relative to Qwen2.5‑7B‑Instruct. Additionally, the higher number of unsuccessful runs observed for N = 5 is  due to out-of-memory (OOM) errors, as the 8B model is approximately 1-1.5GB larger than the 7B variant while operating under the same resource constraints. Even so, \texttt{Qwen3‑8B} achieves steady or improved performance on all datasets. 

We observe a qualitatively different behavior when moving to a larger base model with greater available compute. In additional experiments with the \texttt{devstral-small-2}~\cite{devstral_small_2} model, \textsc{Polaris} exhibits consistently stable evolution behavior, with very few unsuccessful runs, none of which were attributable to out-of-memory failures across all trials. This contrasts with earlier Qwen-based settings, where some evolution runs terminated unsuccessfully under tighter memory and iteration budgets. In the devstral setting, the rare unsuccessful runs that do occur can be traced to suboptimal policy updates early in the evolution process, which bias subsequent repairs and steer the trajectory toward a degraded policy. Importantly, this failure mode is algorithmic rather than resource-driven and could be mitigated using a simple rollback or checkpointing mechanism to revert destabilizing early updates. Notably, the \textsc{Polaris} repair operators and reflection mechanism remain unchanged; the primary difference lies in the underlying model capacity and the increased computational headroom available during evolution. These results indicate that many of the unsuccessful outcomes observed in low-budget settings are attributable to resource constraints rather than fundamental limitations of the evolution process itself. As compute budgets and base model capacity increase, \textsc{Polaris} not only yields stronger policy improvements but also benefits from substantially improved stability, reinforcing its suitability for sustained long-horizon policy evolution as model scales continue to grow. Complete experimental details for the devstral-small-2 setting, along with tables summarizing evolution outcomes and plots of accuracy trajectories, are provided in Appendix~\ref{sec:devstral_results}.
\section{Conclusion}
We introduced \textsc{Polaris}, a framework for recursive self-improvement in small language models through structured, interpretable updates. Unlike prior approaches relying on large-model capacity and unconstrained self-rewrite, \textsc{Polaris} employs a controlled repair cycle and supports runtime updates, enabling post-deployment adaptation. Empirical results demonstrate consistent gains without supervision or retraining, highlighting the feasibility of stable, traceable self-referential learning and its potential for controlled, open-ended improvement in evolving environments.

% We presented \textsc{Polaris}, a framework for recursive self-improvement in small language models via structured, interpretable updates. Unlike prior approaches relying on large-model capacity and unconstrained self-rewrite, \textsc{Polaris} employs a controlled repair cycle integrating reflection, abstraction, patching, and conservative policy updates. Empirical results show consistent performance gains across refinement cycles without supervision or retraining, demonstrating the viability of stable, traceable self-referential learning in resource-constrained agents.

% We presented \textsc{Polaris}, a framework that operationalizes recursive self-improvement for small language models. Unlike prior approaches such as the Gödel Agent, which assume large-model capacity and unconstrained self-rewrite, \textsc{Polaris} introduces a structured repair cycle that grounds self-improvement in interpretable, verifiable updates. By integrating self-reflection, strategy abstraction, patch generation, and conservative policy integration, the framework enables smaller agents to iteratively refine their behavior while preserving stability and traceability. Empirical evaluation on reasoning and comprehension benchmarks demonstrates that \textsc{Polaris} improves task performance through successive refinement cycles without external supervision or retraining. More broadly, this work advances the feasibility of self-referential learning in resource-constrained agents, suggesting a path toward open-ended yet controlled self-improvement in future language model systems.
% \clearpage
\section{Limitations}

\textsc{Polaris} provides a practical approach to recursive self-improvement in small language models, yet some limitations remain. The reduced meta-reasoning capacity, smaller context windows, and limited tool-use capabilities of SLMs constrain the depth of self-reflection and the complexity of policy updates the agent can perform. Moreover, abstraction over larger and more diverse experience sets remains challenging for small language models, as limited context capacity constrains the agent’s ability to consolidate reflections into coherent, generalizable strategies. The dependency on human-designed prompt templates needs to be explored further, and automated prompt template generation is a promising direction for future work. Finally, while the iterative repair cycle supports continual refinement, it does not guarantee monotonic improvement and may increase computational overhead when repair attempts are frequent. These considerations do not undermine the framework’s core contribution but highlight opportunities for extending \textsc{Polaris} toward more expressive, tool-augmented, and stable self-improvement processes.

% \clearpage
\bibliography{custom}
\clearpage
\appendix
% \section{Appendix}

\section{\textsc{Polaris} and Open-Ended Exploration}
\label{sec: open_ended}
In this work, our intent is not to claim fully unconstrained, artificial-life-style open-endedness (with endlessly generated new environments and goals), but rather to follow a now common usage in the Gödel-agent~\cite{yin2024g} and open-ended RL literature: open-endedness in the space of agent designs and internal strategies, evaluated on concrete benchmarks.

Concretely, in our framework, the Gödel-style agent is free to rewrite its own routines, abstractions, and self-improvement code, without a fixed meta-algorithm. This induces an effectively unbounded search space over how the agent reasons, decomposes problems, and organizes its computation. The external tasks (reasoning datasets with objective metrics) are kept fixed to provide a controlled testbed and reproducible measurement. This mirrors the original Gödel Agent work, which describes a self-evolving framework that freely decides its own routine, modules, and even the way to update them, yet evaluates on standard code-editing and reasoning benchmarks. Similarly, the Darwin Gödel Machine~\cite{zhang2025darwin} explicitly bills itself as open-ended evolution of self-improving agents, while its empirical evaluation is on benchmarks such as SWE-bench~\cite{jimenez2024swe}, the open-endedness lies in the archive and continual mutation of coding agents, not in unbounded task generation.

The broader open-endedness community also routinely combines open-ended agent/solution generation with concrete, fixed benchmarks. Position and survey papers define open-ended learning as a process that continually discovers new, diverse and increasingly capable solutions or ``stepping stones'' (policies, programs, strategies), but then instantiate this in specific testbeds to enable careful evaluation \cite{sigaud2023definition}. In open-ended RL, tools such as MiniHack \cite{samvelyan2021minihack} and Craftax \cite{matthews2024craftax} are explicitly described as benchmarks for open-ended reinforcement learning, even though the underlying environments are specific games with well-defined reward functions: the open-endedness comes from the rich combinatorics of the environment and the space of emergent behaviors/tasks, not from an infinitely changing metric.

Our usage is aligned with this practice: we study open-ended self-improvement in agent space (Gödel-style self-modification with abstractions), evaluated on fixed reasoning benchmarks that provide objective metrics and make comparisons to baselines possible. We have observed that performance can plateau after some number of self-improvement steps; however, this plateau is not due to an intrinsic saturation of the evaluation metric, but rather to the limitations of the current implementation e.g., the finite set and design of self-modification operators, the ``imagination'' of the agent in proposing more radical rewrites, and the finite budget of iterations we run. Conceptually, nothing in the framework prevents further exploration: richer operator libraries, more diverse abstraction schemes, or longer runs could allow the agent to escape such plateaus and continue discovering improvements, just as more sophisticated exploration mechanisms unlock further progress in open-ended RL benchmarks.

\section{Additional Experimental Details and Analysis}
\label{sec: additional_experimental_details}

In Figure \ref{fig: goal_prompt}, we provide the goal prompt for the agent. We adapt the agent's goal prompt from \cite{yin2024g} and introduce instructions for small language models such as

\begin{itemize}
  \item \textbf{\texttt{action\_adjust\_logic}:} Added ``Do not do unnecessary changes'' and clarified it may be used to create targeted \emph{action functions} for the solver; original constraints (such as check imports/usages, do not change interfaces) remain.
  \item \textbf{Techniques block:} Replaced the brief hint with a concrete list: LLM Debate, Step-back Abstraction, Quality-Diversity, Dynamic Roles, Self-consistency (with \texttt{num\_of\_response}), Few-shots, Task Decomposition, Reflective Evaluation.
  \item \textbf{\texttt{action\_display\_analysis}:} Removed the low-score case-study requirement; added that \texttt{action\_call\_json\_format\_llm} can also perform analysis.
  \item \textbf{Reminder prompting to the agent:} Call \texttt{action\_evaluate\_on\_task} \emph{only after} modifying the solver via \texttt{action\_adjust\_logic}; multiple tools may be called when needed.
\end{itemize}

\begin{table}[!tbh]

    \begin{subtable}[t]{\linewidth}

        \small
        \centering
        \resizebox{\linewidth}{!}{
        \begin{tabular}{|l|c|c|c|c|}
        \hline
        & \textbf{MGSM}
        & \textbf{DROP}
        & \textbf{GPQA}
        & \textbf{LitBench} \\
        \hline
        Successful        & 4 & 2 & 3 & 4 \\ \hline
        No improvement    & 1 & 0 & 1 & 1 \\ \hline
        Unsuccessful      & 0 & 3 & 1 & 0 \\ \hline
        Total             & 5 & 5 & 5 & 5 \\ \hline
        \end{tabular}}

        \caption{$N=3$}

    \end{subtable}

    \vspace{0.5em}

    \begin{subtable}[t]{\linewidth}

        \small
        \centering
        \resizebox{\linewidth}{!}{
        \begin{tabular}{|l|c|c|c|c|}
        \hline
        & \textbf{MGSM}
        & \textbf{DROP}
        & \textbf{GPQA}
        & \textbf{LitBench} \\
        \hline
        Successful        & 2 & 0 & 2 & 0 \\ \hline
        No improvement    & 0 & 1 & 1 & 2 \\ \hline
        Unsuccessful      & 3 & 4 & 2 & 3 \\ \hline
        Total             & 5 & 5 & 5 & 5 \\ \hline
        \end{tabular}}

        \caption{$N=5$}

    \end{subtable}

    \caption{A summary of various runs of \textsc{Polaris} on datasets using the Qwen3‑8B model. For policy repair and experience abstraction, we consider a set of $N$ failed instances from the validation set of each dataset.}
    \label{tab:qwen3-8b-iterations}

\end{table}

\begin{figure*}[t]
    \centering
    \begin{subfigure}{0.48\linewidth}
        \centering
        \includegraphics[width=\linewidth]{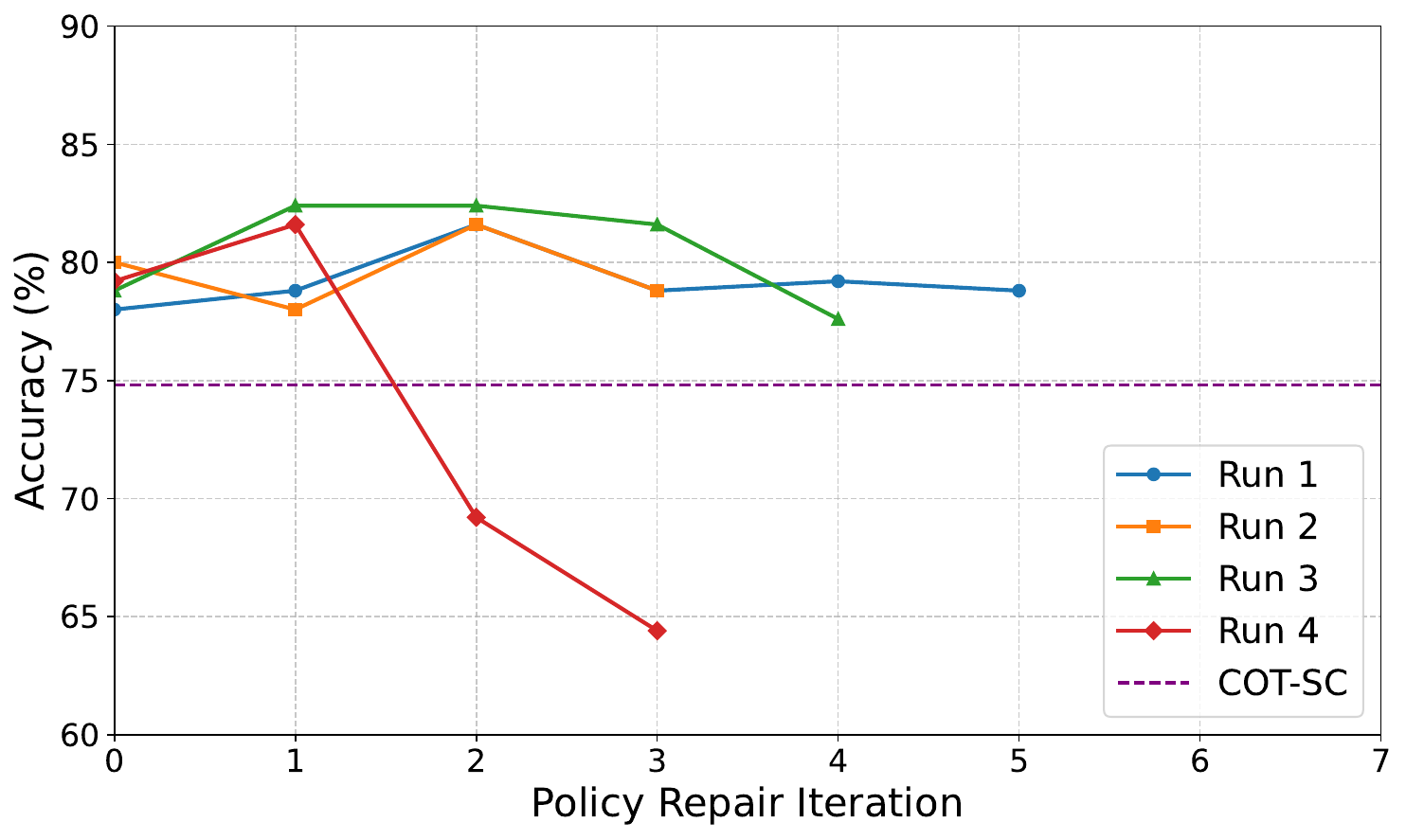}
        \caption{MGSM}
    \end{subfigure}
    \begin{subfigure}{0.48\linewidth}
        \centering
        \includegraphics[width=\linewidth]{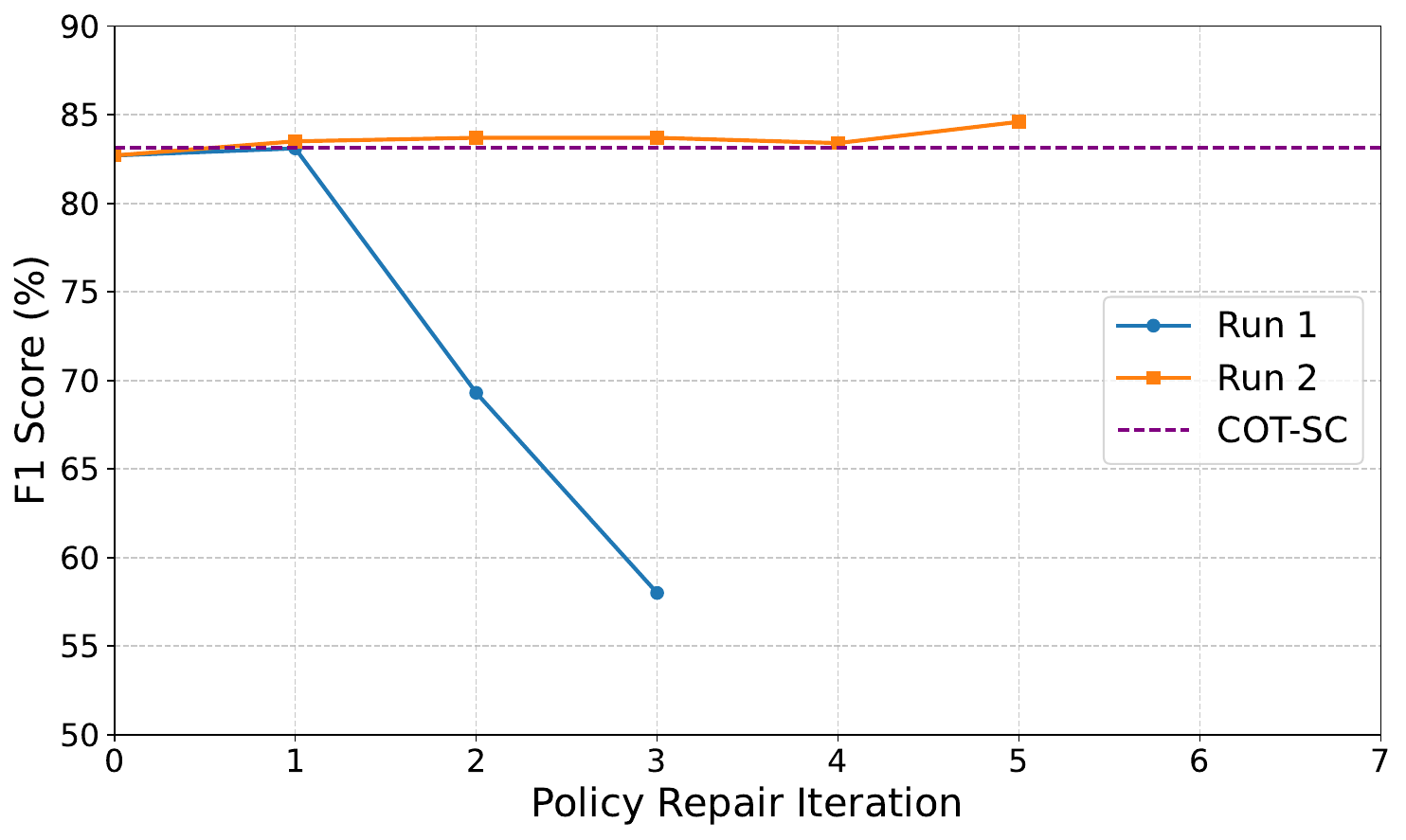}
        \caption{DROP}
    \end{subfigure}
    \begin{subfigure}{0.48\linewidth}
        \centering
        \includegraphics[width=\linewidth]{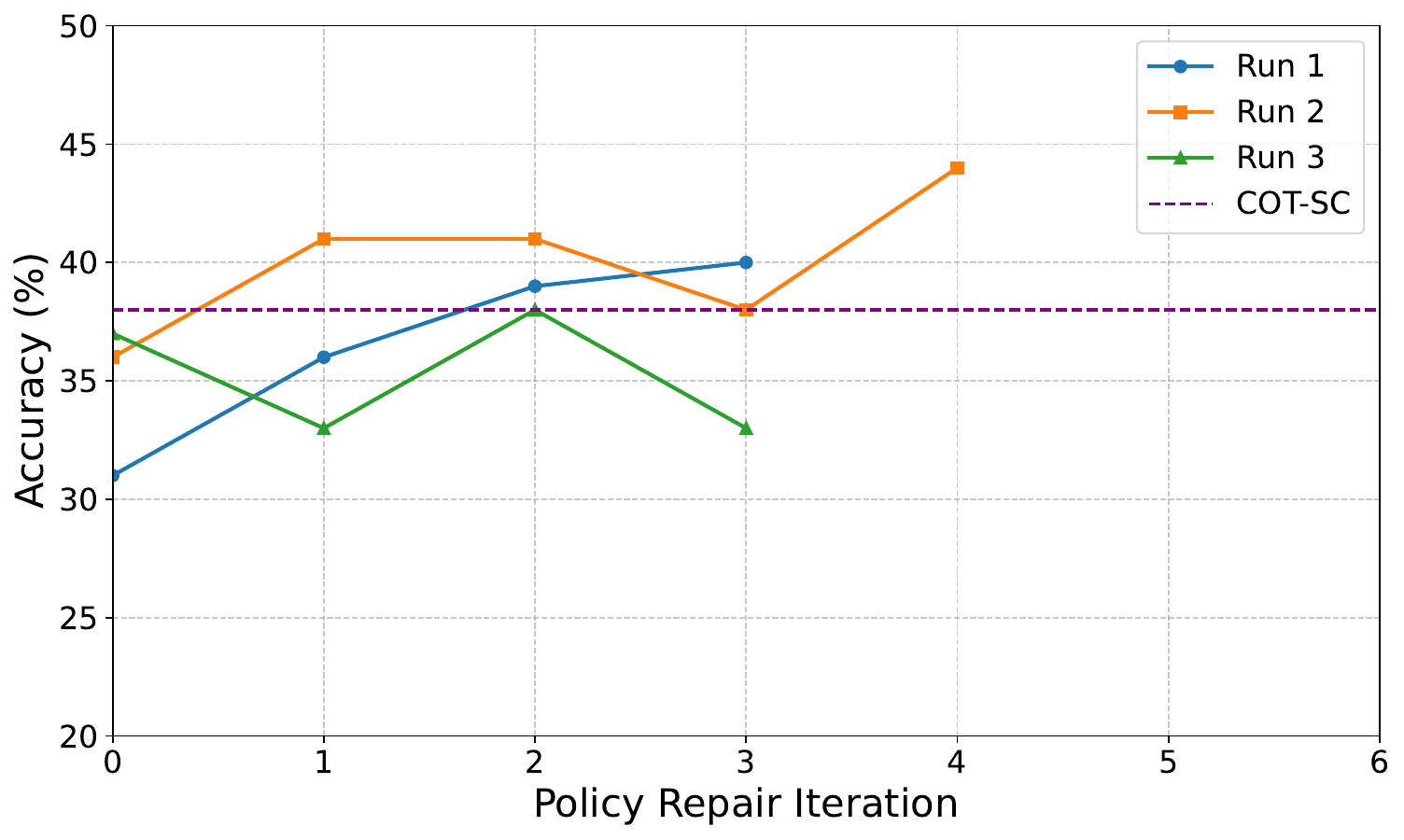}
        \caption{GPQA}
    \end{subfigure}
    \begin{subfigure}{0.48\linewidth}
        \centering
        \includegraphics[width=\linewidth]{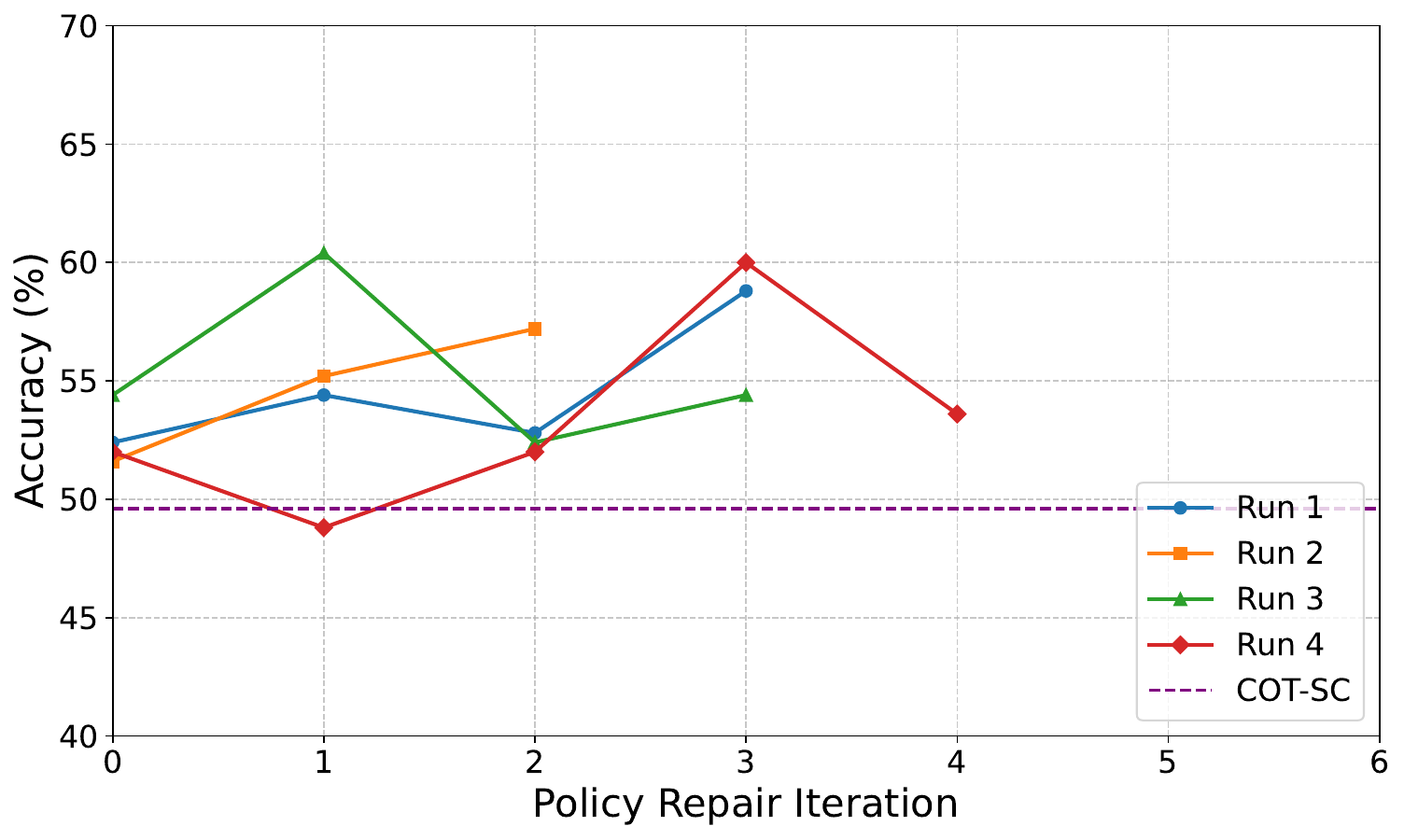}
        \caption{LitBench}
    \end{subfigure}
    % \hfill
    \caption{Successful evolution runs of \textsc{Polaris} using Qwen3‑8B model, with performance improvement compared to the base policy and COT-SC. Policy Repair Iteration 0 shows the performance with the base policy. For policy repair and experience abstraction, we consider a set of three failed instances from the validation set of each dataset ($N$=3).}
    \label{fig:qwen3_results_exp_set_3}
\end{figure*}

\begin{figure*}[t]
    \centering
    \begin{subfigure}{0.48\linewidth}
        \centering
        \includegraphics[width=\linewidth]{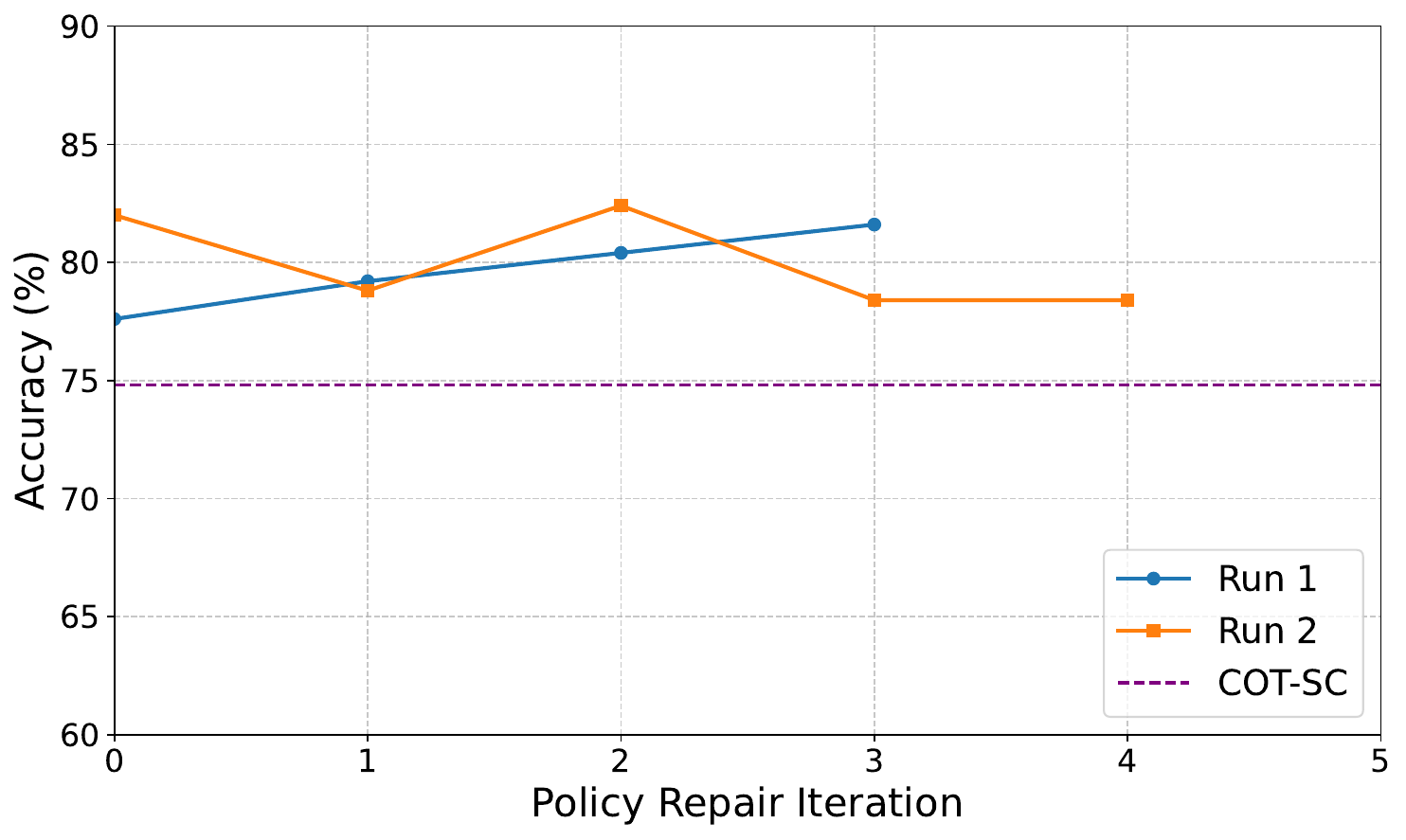}
        \caption{MGSM}
    \end{subfigure}
    \begin{subfigure}{0.48\linewidth}
        \centering
        \includegraphics[width=\linewidth]{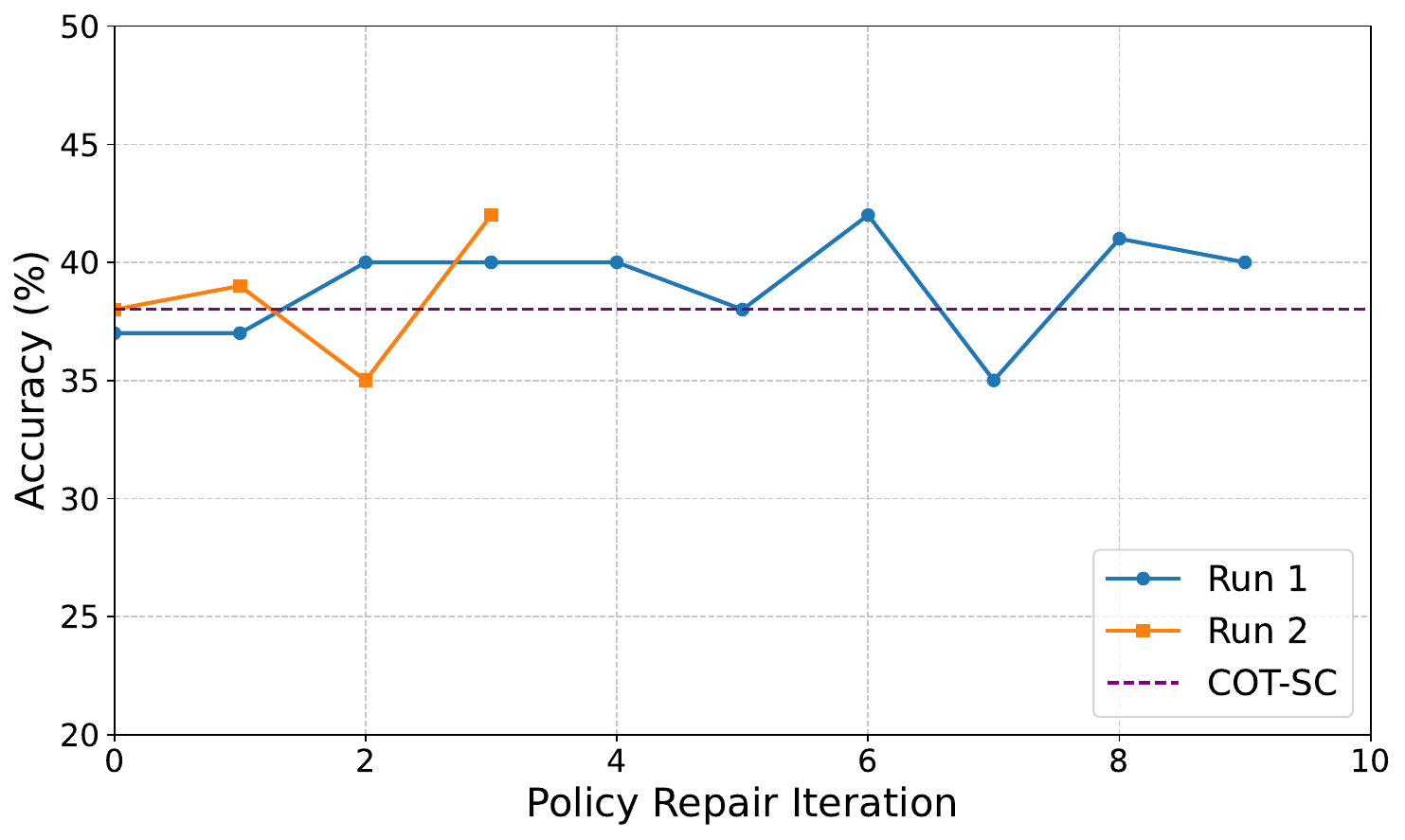}
        \caption{GPQA}
    \end{subfigure}
    % \hfill
    \caption{Successful evolution runs of \textsc{Polaris} using Qwen3‑8B model, with performance improvement compared to the base policy and COT-SC. Policy Repair Iteration 0 shows the performance with the base policy. For policy repair and experience abstraction, we consider a set of three failed instances from the validation set of each dataset ($N$=5).}
    \label{fig:qwen3_results_exp_set_5}
\end{figure*}

\begin{figure*}[t]
    \centering
    \begin{subfigure}{0.48\linewidth}
        \centering
        \includegraphics[width=\linewidth]{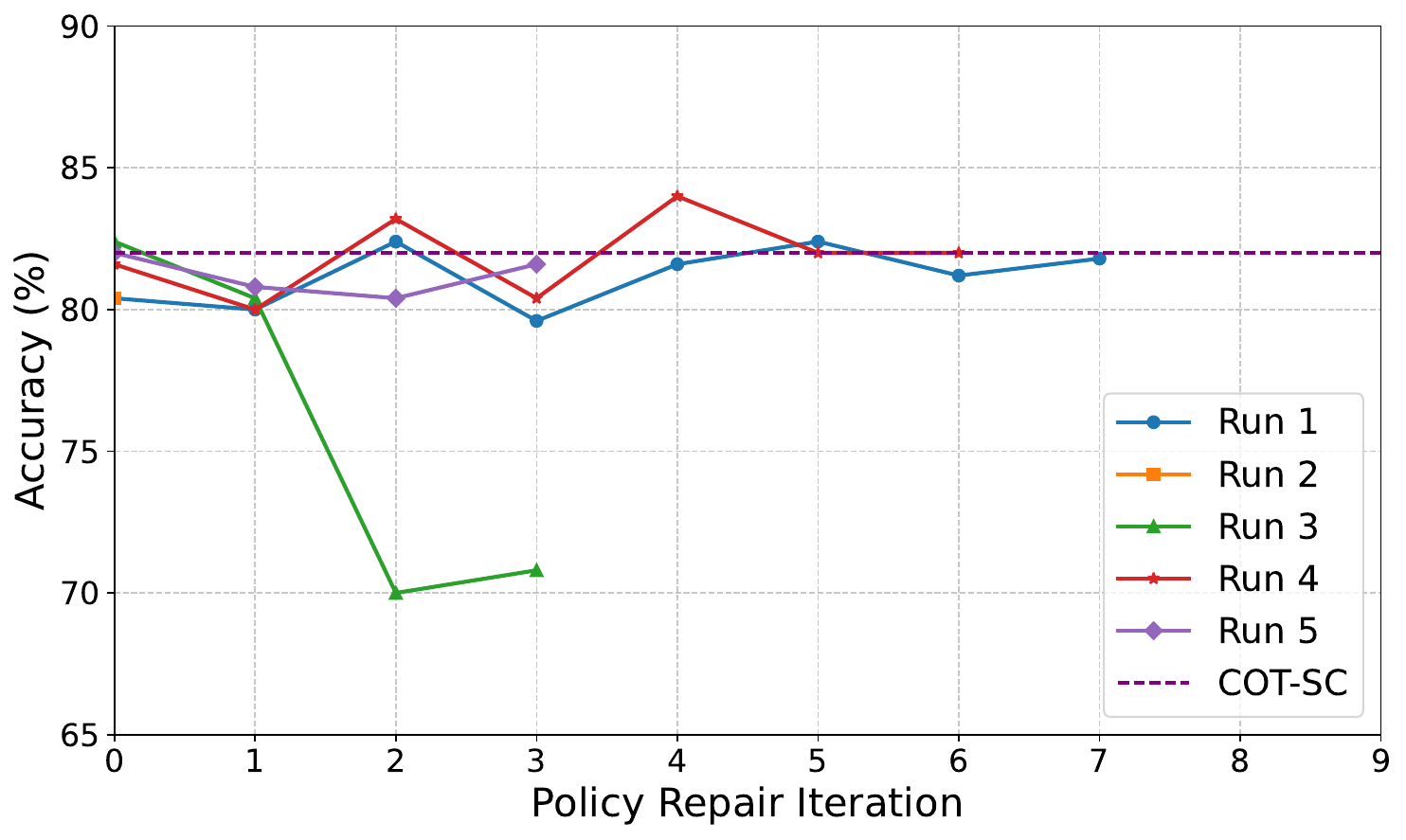}
        \caption{MGSM}
    \end{subfigure}
    \begin{subfigure}{0.48\linewidth}
        \centering
        \includegraphics[width=\linewidth]{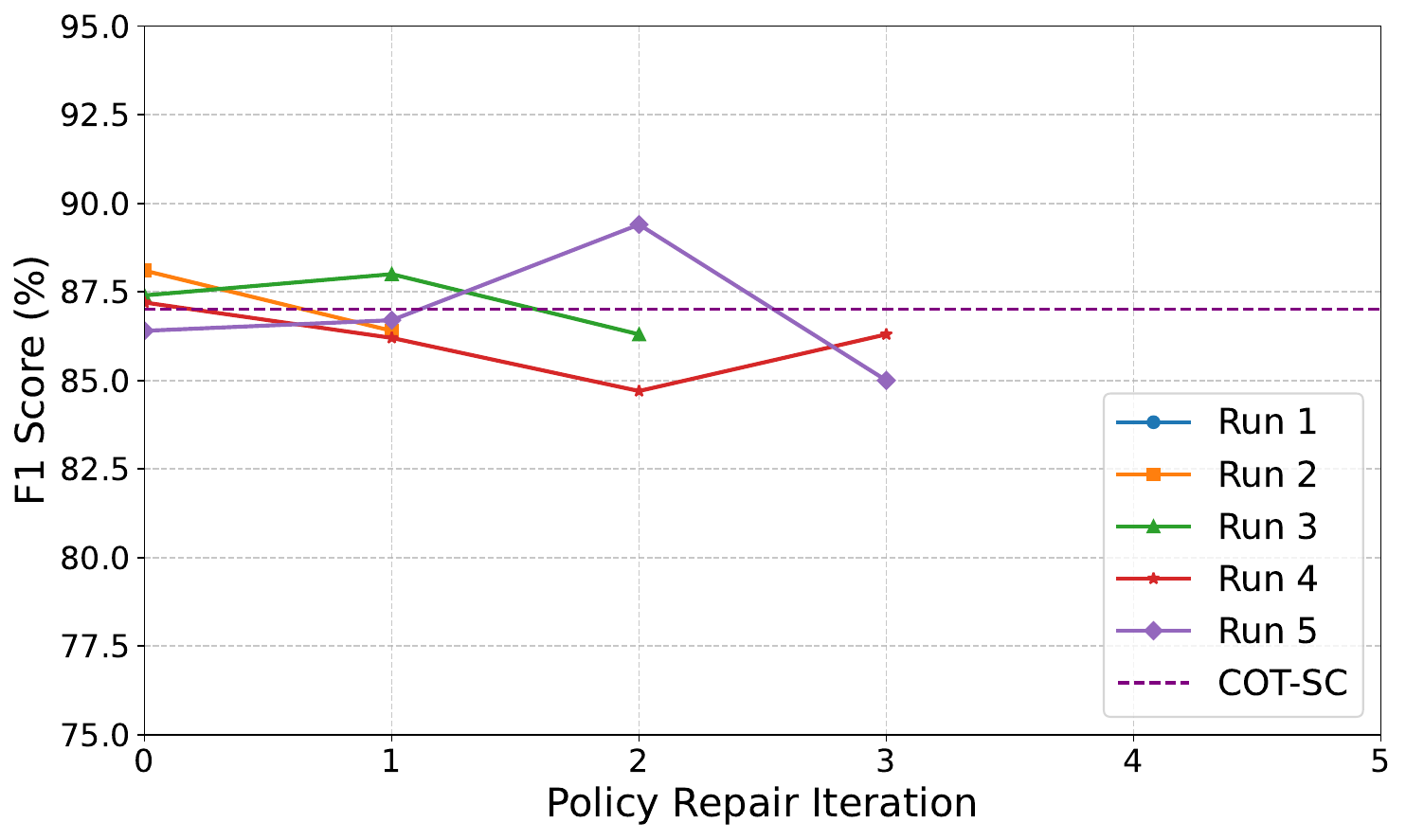}
        \caption{DROP}
    \end{subfigure}
    \begin{subfigure}{0.48\linewidth}
        \centering
        \includegraphics[width=\linewidth]{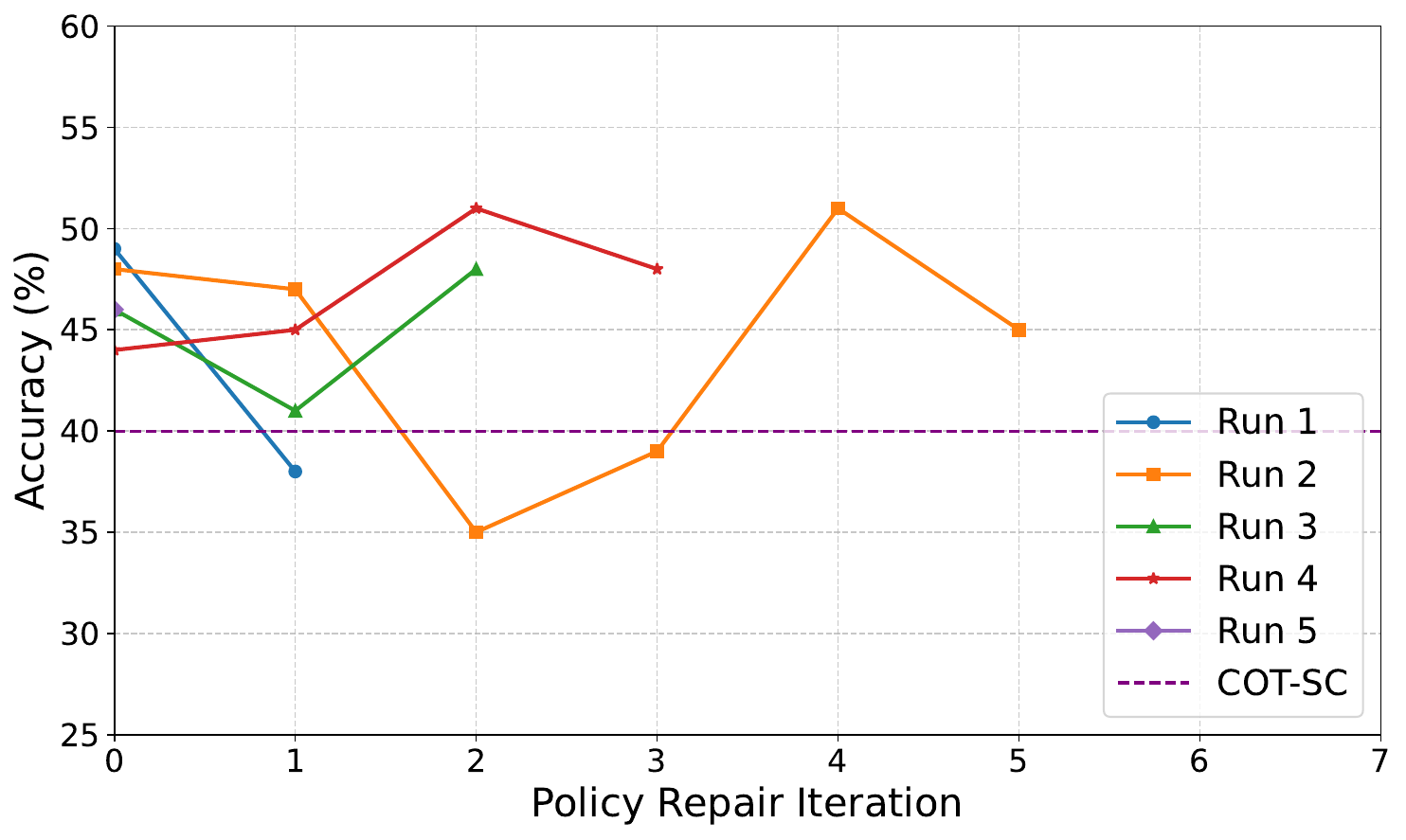}
        \caption{GPQA}
    \end{subfigure}
    \begin{subfigure}{0.48\linewidth}
        \centering
        \includegraphics[width=\linewidth]{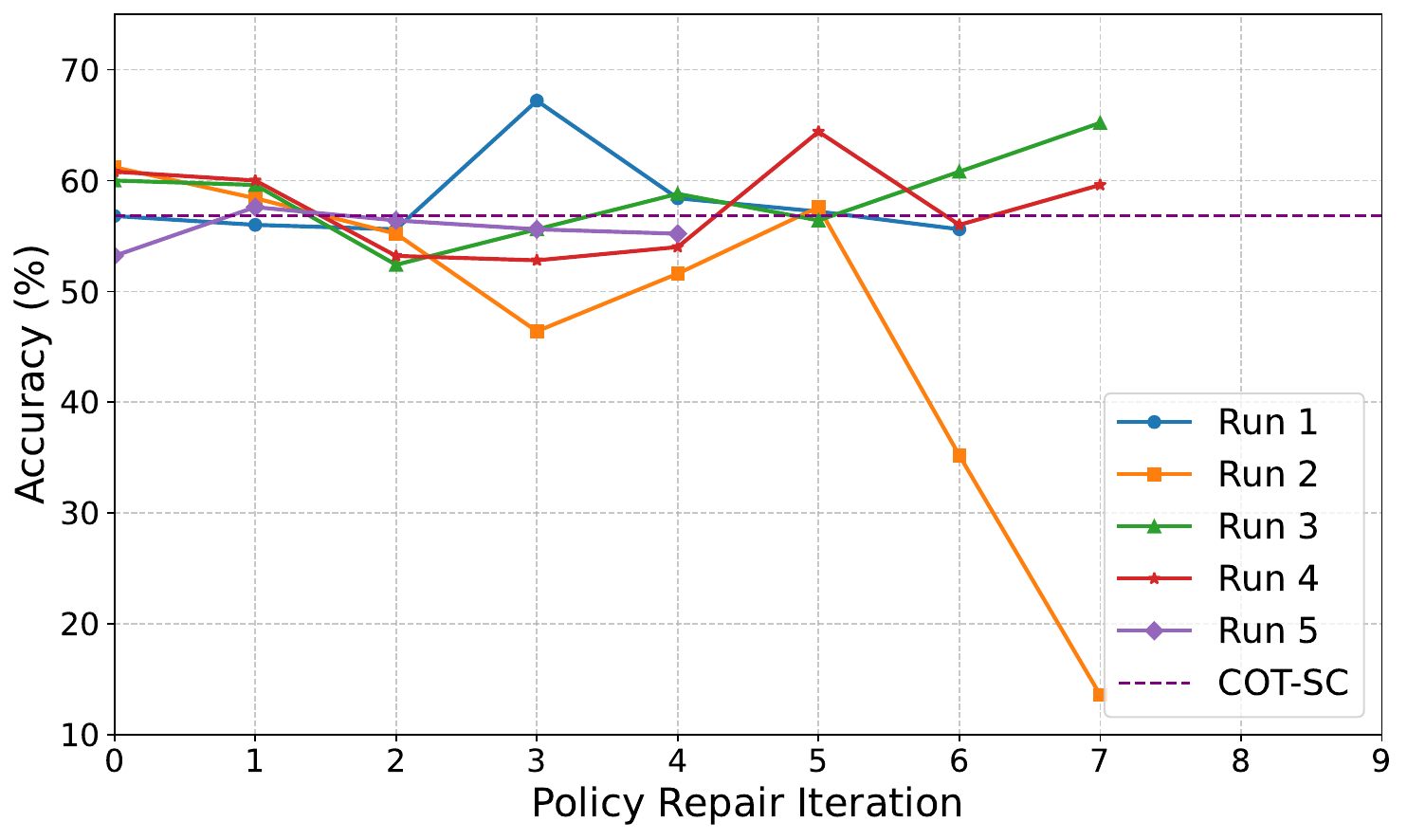}
        \caption{LitBench}
    \end{subfigure}
    % \hfill
    \caption{Evolution trajectories of \textsc{Polaris} using the devstral-small-2 model. Each subplot shows the accuracy trajectories for all five independent evolution runs, illustrating stable policy improvements relative to the base policy and the CoT-SC baseline. Policy Repair Iteration 0 corresponds to the base policy. For policy repair and experience abstraction, we consider a set of three failed instances from the validation set of each dataset ($N=3$).}
    \label{fig:devstral-small-2-trajectories}
\end{figure*}

\begin{figure}[t]
\small
    \begin{center}
    \begin{tcolorbox}[
      %enhanced,
      %breakable,
      colback=orange!5!white,
      colframe=orange!50!black,
      title=\textbf{Analyze Failures},
      fonttitle=\bfseries,
      sharp corners,
      boxrule=1pt,
      width=\linewidth
    ]
    
    \textcolor{red}{You are analyzing why the current policy failed on a given task. Your goal is to identify the policy’s shortcomings and propose actionable improvements}. \\
     
    \textbf{Inputs}: \\
    - Question: \{question\}  \\
    - Your Reasoning: \{reasoning\}  \\
    - Your Answer: \{answer\}  \\
    - Correct Answer: \{correct\_answer\}  \\
    - Policy: \{current\_policy\}  \\
    
    Carefully reflect on why the policy produced the wrong result. \textbf{Your reflection must include three elements}:  \\
    1. \textcolor{blue}{A clear explanation of the failure.} Examine how the policy’s logic or structure caused the error.\\
    2. \textcolor{blue}{Step-by-step suggestions} on how the policy could be revised to solve the task.  \\
    3. \textcolor{blue}{Advice to prevent similar failures} in the future.
    \end{tcolorbox}
    \captionof{figure}{Prompt for analyzing failures on task samples through self-reflection.}
    \label{fig:analyze_failures_prompt}
    \end{center}
\end{figure}

\begin{figure}[!tbh] 
\small
    \begin{center}
    \begin{tcolorbox}[
      %enhanced,
      %breakable,
      colback=orange!5!white,
      colframe=orange!50!black,
      title=\textbf{Strategy Synthesis},
      fonttitle=\bfseries,
      sharp corners,
      boxrule=1pt,
      width=\linewidth
    ]
    
    \textcolor{red}{You are an expert AI engineer analyzing self-reflection on policy from multiple failed tasks.}\\
    
    \textbf{Inputs:}\\
    - Reflections: \{combined\_reflections\}\\
    - Current Policy: \{current\_policy\}\\
    - Prior Strategies: \{agent.prior\_strategies\} \\
    
    Your task is to extract *1-2 new* generalizable and non-redundant policy improvement strategies from the task-level reflections.\\
    
    \textbf{Rules:}\\
    - \textcolor{blue}{Do not repeat} or restate any of the previously extracted strategies.  \\
    - The strategy should \textcolor{blue}{target the root cause} behind the failures observed in the reflections.  \\
    - It must be \textcolor{blue}{reusable across tasks} and focused on policy improvements (not tied to one failure instance). \\ 
    - Do not copy raw reflections; \textcolor{blue}{abstract reflections into a reusable *insight*.}  \\
    - Write this as if \textcolor{blue}{giving coding instructions to another engineer.}  \\
    - Output only *1-2* new generalizable improvement strategies, written as \textcolor{blue}{short, clear statements.}
    \end{tcolorbox}
    \captionof{figure}{Prompt for policy repair planning and abstraction. Agent synthesizes the generalized policy repair strategies based on the self-reflection on failed task samples on the current policy. It also considers the prior strategies to avoid redundancy.}
    \label{fig:strategy_prmpt}
    \end{center}
\end{figure}

\begin{figure}[t] 
\small
    \begin{center}
    \begin{tcolorbox}[
      %enhanced,
      %breakable,
      colback=orange!5!white,
      colframe=orange!50!black,
      title=\textbf{Patch Generation},
      fonttitle=\bfseries,
      sharp corners,
      boxrule=1pt,
      width=\linewidth
    ]
    
    \textcolor{red}{You are assisting in improving the current policy.} \\
    
    \textbf{Inputs:}  \\
    - Current Policy: \{current\_policy\}  \\
    - Repair Strategies: \{repair\_strategies\}  \\
    
    \textbf{Your task:}  \\
    - For each strategy, \textcolor{blue}{propose a minimal **code patch**} to implement it.  \\
    - Show \textcolor{blue}{only new or modified lines}, do not repeat unchanged code.\\
    - \textcolor{blue}{No explanations}.  \\
    
    \textbf{Format your response as:}  \\
    \textbf{\#\#\# Strategy:} <chosen strategy>  \\
    \textbf{\#\#\# Patch:}
    <only the modified or new lines of Python code>
    \end{tcolorbox}
    \captionof{figure}{Prompt for generating code patches from policy repair strategies.}
    \label{fig:code_patch_prompt}
    \end{center}
\end{figure}

\begin{figure}[t] 
\small
    \begin{center}
    \begin{tcolorbox}[
      %enhanced,
      %breakable,
      colback=orange!5!white,
      colframe=orange!50!black,
      title=\textbf{Update Policy},
      fonttitle=\bfseries,
      sharp corners,
      boxrule=1pt,
      width=\linewidth
    ]
    \textcolor{red}{You are a coding assistant. Your task is to apply all the provided code patches to the current policy and return the fully updated version of the policy.} \\
    
    Current policy: \{current\_policy\} \\
    
    \textbf{Rules:}\\
    - \textcolor{blue}{Insert or replace} ONLY the lines shown in the patch. \\
    - Keep ALL \textcolor{blue}{other lines of the policy unchanged.} \\
    - \textcolor{blue}{Do NOT remove or overwrite existing logic} unless explicitly replaced by the patch. \\
    - Ensure ALL \textcolor{blue}{patches are correctly integrated} (e.g., imports, variables, helper functions must exist). \\
    - If a patch introduces new logic that requires dependencies (imports, helper methods, variables), \textcolor{blue}{ADD them safely}. \\
    - \textcolor{blue}{Resolve conflicts} so the final policy is consistent and executable. \\
    - The updated policy MUST be \textcolor{blue}{logically correct, consistent, and error-free.} \\
        - Always return the \textcolor{blue}{FULL policy wrapped in}: \\
    ```python 
    <code patch here>
    '''

    % \#\#\# Updated policy: \\
    \end{tcolorbox}
    \captionof{figure}{Prompt for integrating code patches into current policy.}
    \label{fig:update_policy_prompt}
    \end{center}
\end{figure}

Furthermore, in Figure \ref{fig: helper}, we provide the helper agent prompt that helps correct the output format to valid JSON during the evaluation of the policy. We provide examples in the prompt to obtain the target behaviour.

\subsection{Experimental setup (\texttt{devstral‑small‑2})}
\label{sec:devstral_results}

To evaluate \textsc{Polaris} on a model outside the Qwen family, we conducted experiments using the \texttt{devstral-small-2}~\cite{devstral_small_2} model, a 24B-parameter instruction-tuned model from the Mistral family, using 4-bit quantization. Due to the higher resource demands of this model, we adopted a constrained evolution protocol following Gödel Agent~\citep{yin2024g}, running a fixed budget of 30 evolution steps rather than our standard 10-hour evolution window. All experiments were executed on two NVIDIA RTX 6000 Ada GPUs with 48\,GB memory each. We performed five independent \textsc{Polaris} runs on the MGSM, DROP, GPQA and Litbench dataset, using $N=3$ failed instances per reflection cycle for policy repair and experience abstraction. A summary of evolution outcomes across runs is reported in Table~\ref{tab:devstral-small-2-iterations}, and the corresponding accuracy trajectories over evolution steps are shown in Figure~\ref{fig:devstral-small-2-trajectories}. Methods, repair operators, and validation procedures were kept identical to those used in the Qwen-based experiments.

\begin{table}[!tbh]

    \begin{subtable}[t]{\linewidth}

        \small
        \centering
        \resizebox{\linewidth}{!}{
        \begin{tabular}{|l|c|c|c|c|}
        \hline
        \textbf{$N=3$}
        & \textbf{MGSM}
        & \textbf{DROP}
        & \textbf{GPQA}
        & \textbf{LitBench} \\
        \hline
        Successful        & 2 & 2 & 2 & 4 \\ \hline
        No Improvement    & 2 & 2 & 2 & 1 \\ \hline
        Unsuccessful      & 1 & 1 & 1 & 0 \\ \hline
        Total             & 5 & 5 & 5 & 5 \\ \hline
        \end{tabular}}

        \caption{$N=3$}

    \end{subtable}

    \caption{A summary of various runs of \textsc{Polaris} on datasets using the devstral-small-2 model. For policy repair and experience abstraction, we consider a set of $N=3$ failed instances from the validation set of each dataset.}
    \label{tab:devstral-small-2-iterations}

\end{table}

\subsection{Example runs of \textsc{Polaris}}
\begin{figure*}[t]
\centering
% \resizebox{\linewidth}{!}{
\includegraphics[width=\textwidth]{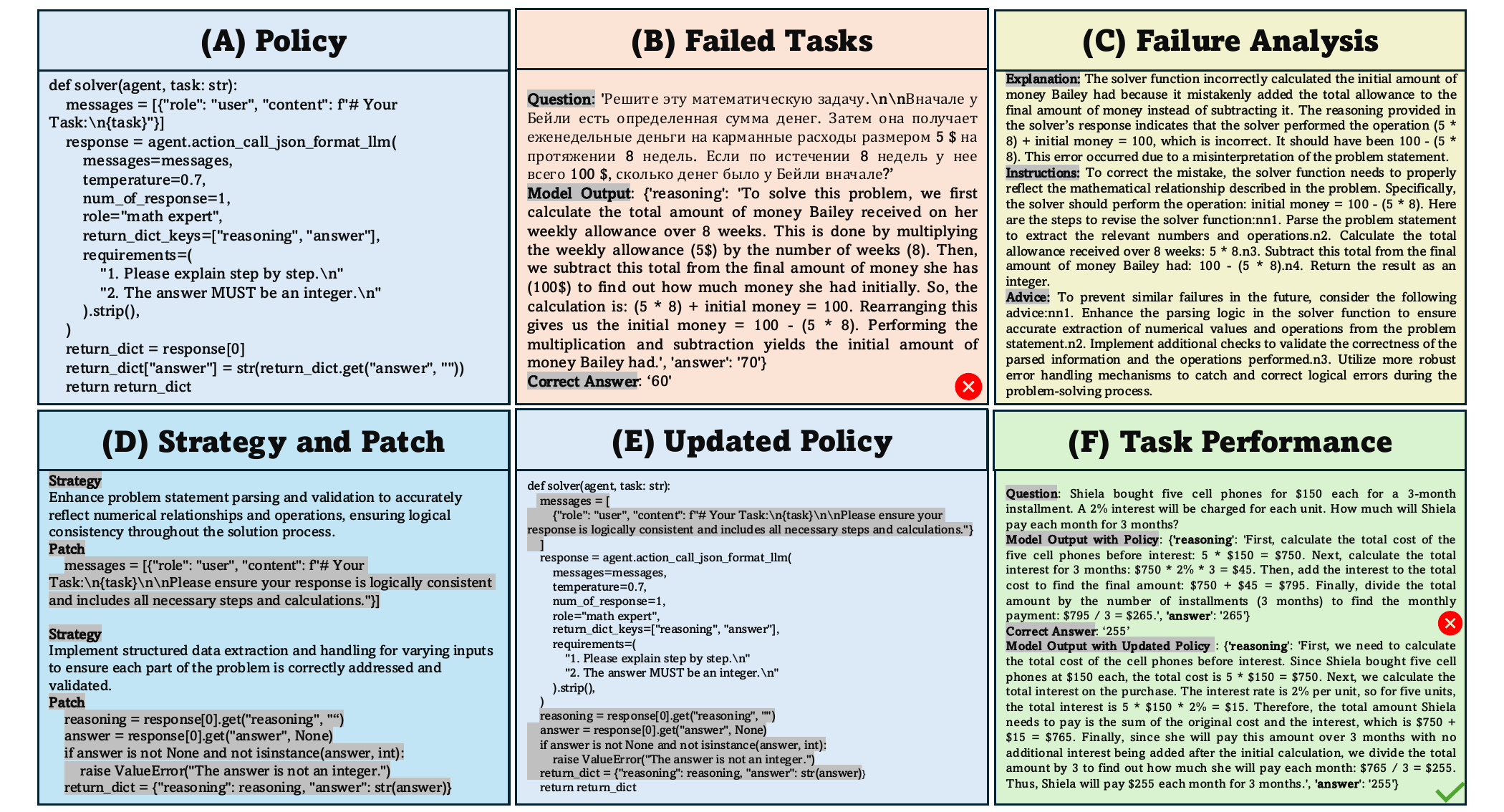}
% }
\caption{An example of policy repair via experience abstraction with \textsc{Polaris} on the MGSM dataset. The initial policy (A) failed to correctly answer the question (B). The failure analysis revealed that the Agent struggled to apply the steps logically while performing the calculation (C). During strategy synthesis (D), a recommendation is made to ensure the response is logically consistent and includes all necessary steps and calculations. The corresponding patch is integrated to the current policy (E) resulting in improved task performance on unseen tasks (F).}
\label{fig: mgsm_example}
\end{figure*}

\begin{figure*}[t]
\centering
% \resizebox{\linewidth}{!}{
\includegraphics[width=\textwidth]{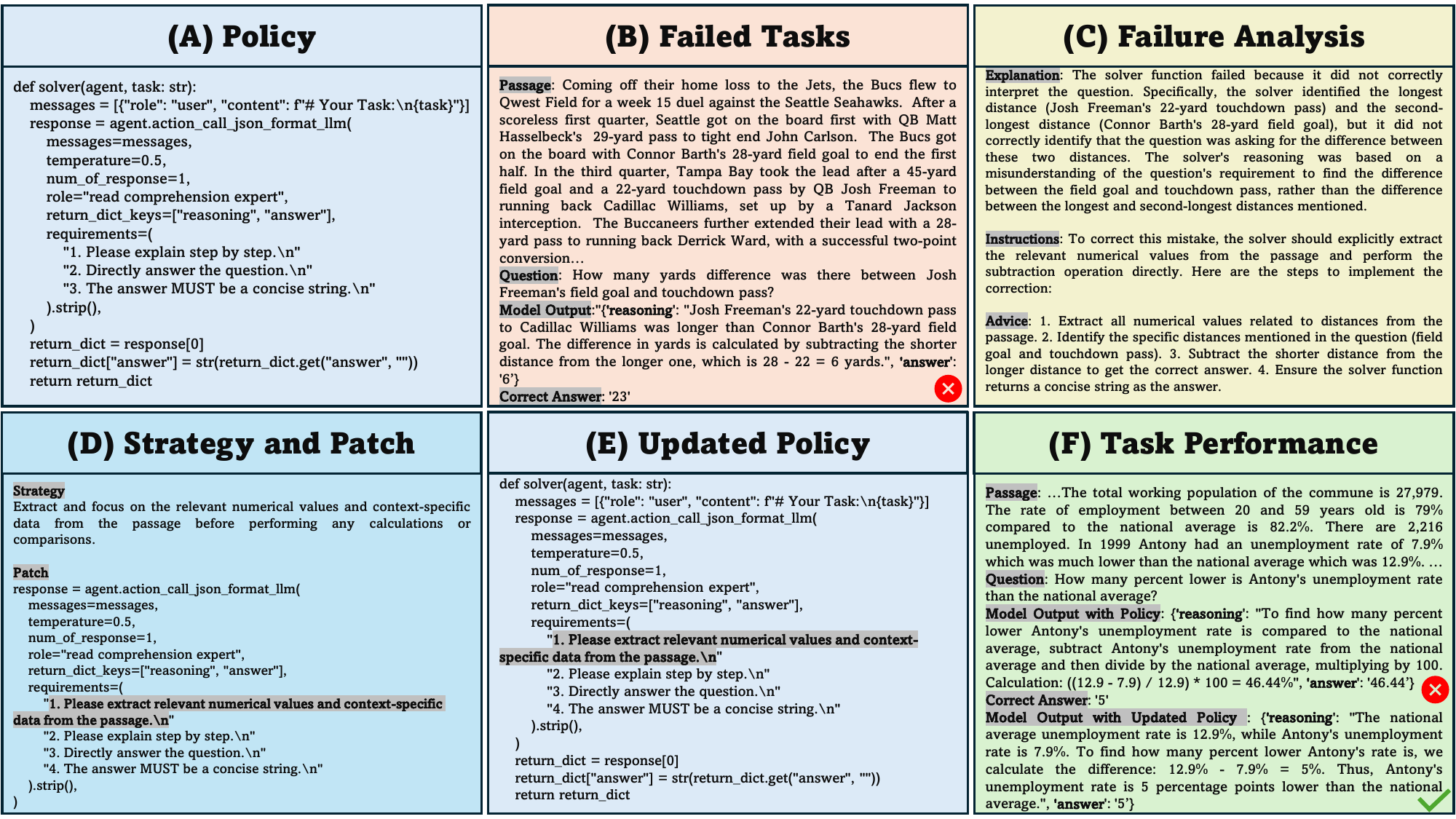}
% }
\caption{An example of policy repair via experience abstraction with \textsc{Polaris} on the DROP dataset. The initial policy (A) failed to correctly answer the question (B). The failure analysis revealed that the Agent struggled with handling numerical data (C). During strategy synthesis (D), a recommendation is made to focus on the numerical values and context-specific data. The corresponding patch is integrated to the current policy (E) resulting in improved task performance on unseen tasks (F).}
\label{fig:teaser}
\end{figure*}

\begin{figure*}[t]
\centering
% \resizebox{\linewidth}{!}{
\includegraphics[width=\textwidth]{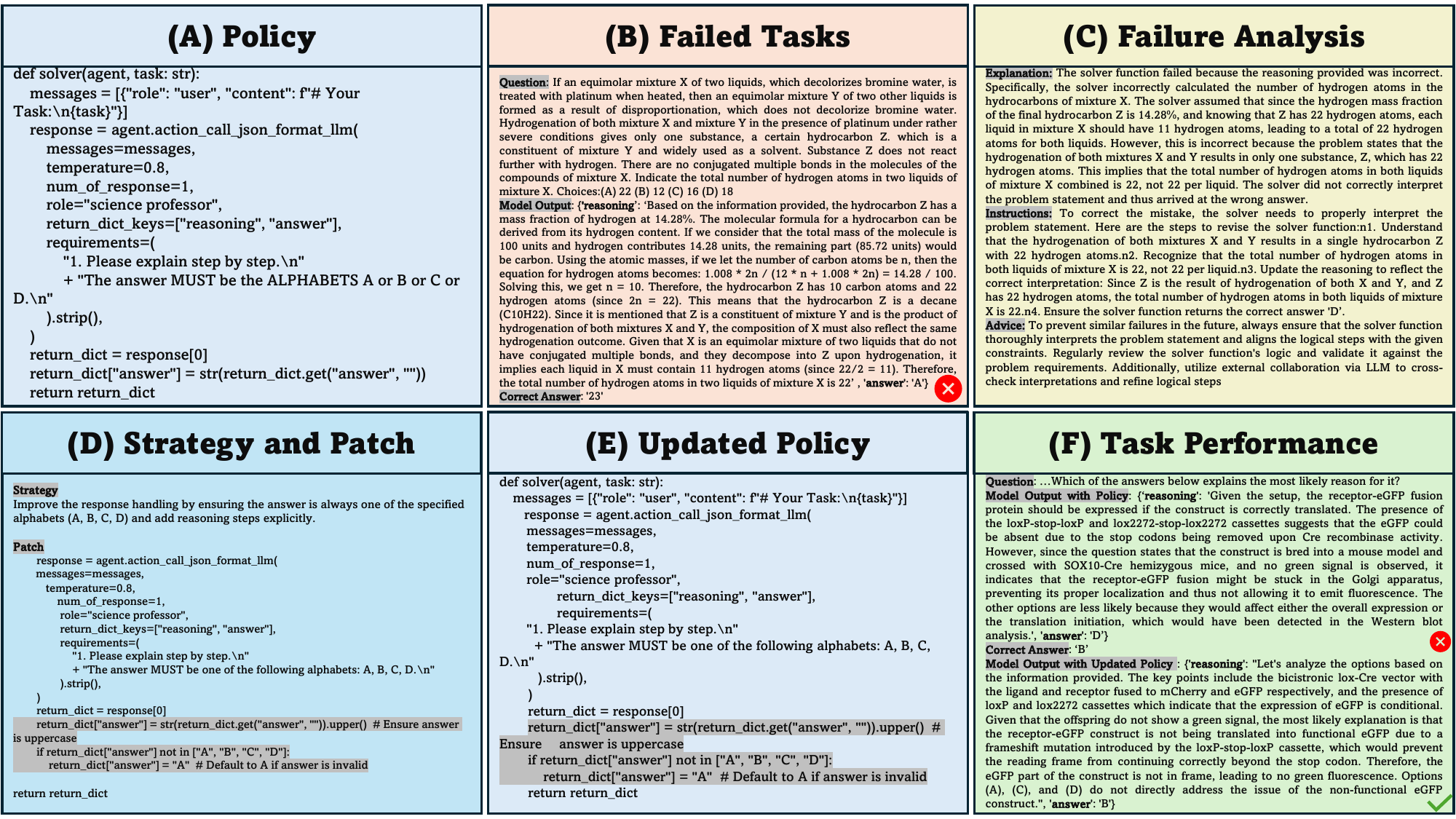}
% }
\caption{An example of policy repair via experience abstraction with \textsc{Polaris} on the GPQA dataset. The initial policy (A) failed to correctly answer the question (B). The failure analysis revealed that the Agent failed to provide the correct response because the reasoning provided was incorrect (C). During strategy synthesis (D), a recommendation is made to add reasoning steps explicitly along with the check for output response format. The corresponding patch is integrated to the current policy (E) resulting in improved task performance on unseen tasks (F).}
\label{fig: gpqa_example}
\end{figure*}

\begin{figure*}[t]
\centering
% \resizebox{\linewidth}{!}{
\includegraphics[width=\textwidth]{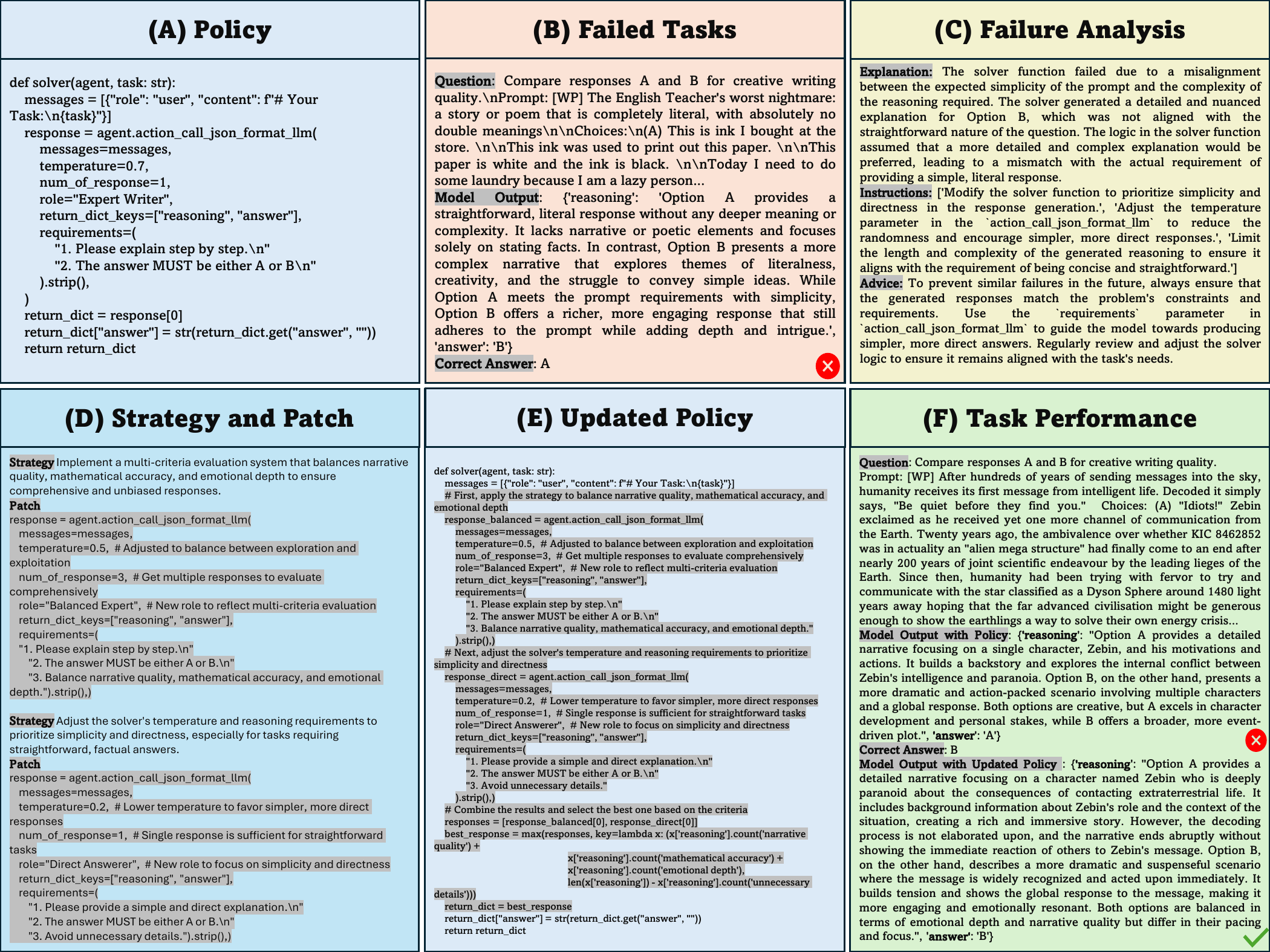}
% }
\caption{An example of policy repair via experience abstraction with \textsc{Polaris} on the LitBench dataset. The initial policy (A) failed to correctly answer the question (B). The failure analysis revealed that the Agent struggled with handling numerical data (C). During strategy synthesis (D), a recommendation is made to focus on the numerical values and context-specific data. The corresponding patch is integrated to the current policy (E) resulting in improved task performance on unseen tasks (F).}
\label{fig: litbench_example}
\end{figure*}

In Figures \ref{fig:teaser}, \ref{fig: mgsm_example}, \ref{fig: gpqa_example}, and \ref{fig: litbench_example}, we present examples from different datasets illustrating the steps of the \textsc{Polaris} algorithm. The initial policy fails to correctly answer the question. Failure analysis generates explanations, instructions, and advice for the agent. During strategy synthesis, recommendations are formulated to correct prior behavior by updating the policy. The corresponding patch is then integrated into the current policy, resulting in improved task performance on unseen instances.

\subsection{Policy update examples}

\begin{figure*}[t]
    \centering
    \begin{subfigure}{\linewidth}
        \centering
        \includegraphics[width=\linewidth]{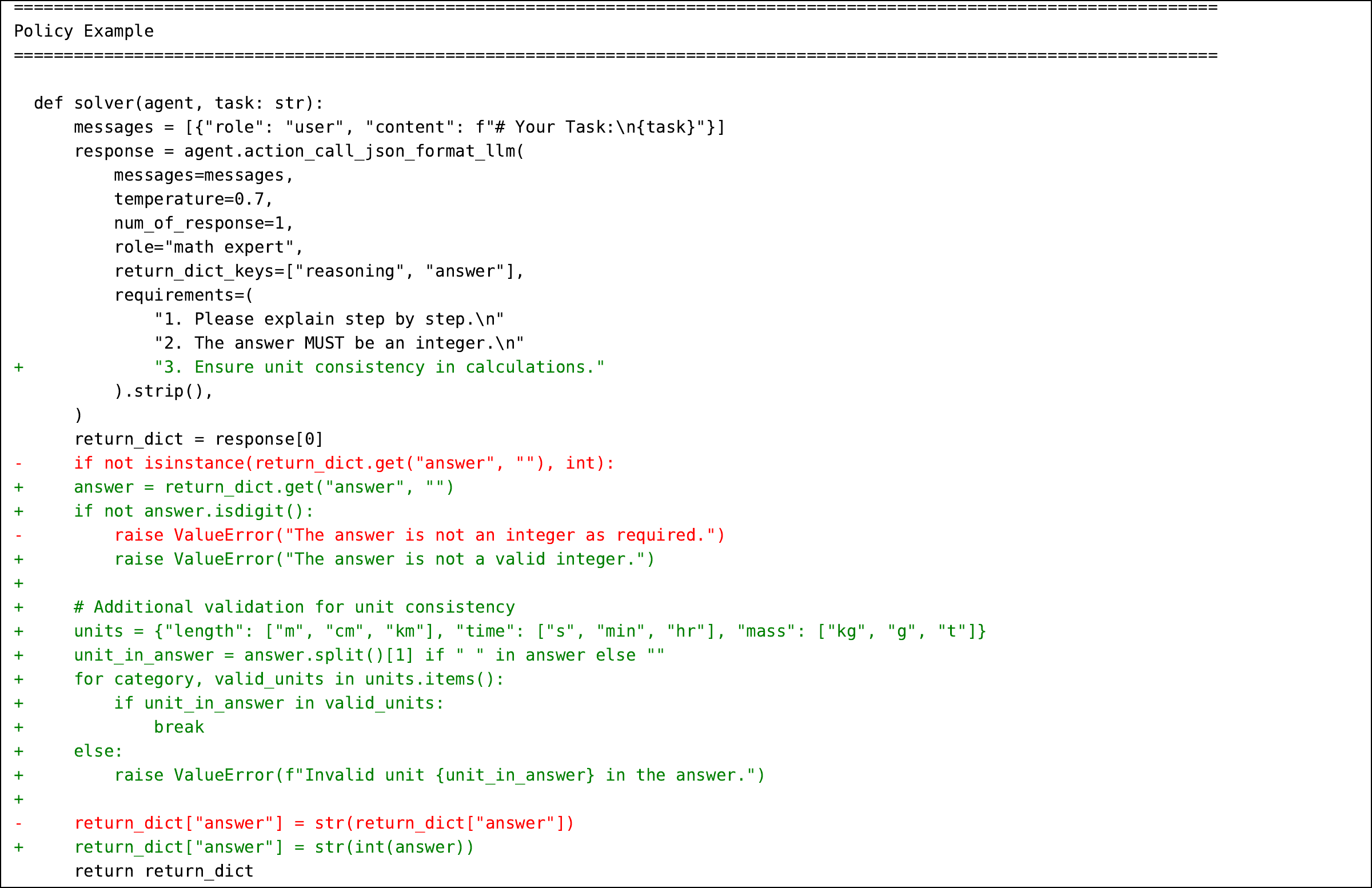}
        % \caption{}
    \end{subfigure}
    % \begin{subfigure}{0.5\linewidth}
    %     \centering
    %     \includegraphics[width=\linewidth]{images/policy_examples/drop_policy_diffs3.pdf}
    %     \caption{}
    % \end{subfigure}
    % \hfill
    \caption{Policy update example on the MGSM dataset. We highlight the updates in the current policy with respect to the previous policy using green color (new statements added) and red color (statements deleted). We observe the addition of the requirement and the logic to ensure unit consistency in calculations while deleting and updating the exception handling statements.}
    \label{fig: policy_examples_mgsm2}
\end{figure*}

In Figures \ref{fig: policy_examples_mgsm2}–\ref{fig: policy_examples_litbench2}, we present example policy updates across datasets. We highlight changes to the current policy relative to the previous policy using green (additions) and red (deletions). The updates span multiple dimensions, including the addition of complex requirements, expert evaluators, data-type changes, conditional statements, exception handling, and specialized scoring parsers. These niche updates underscore the role of experience abstraction in enabling targeted policy refinements over iterations.

\subsection{Variance in performance}
\label{sec: variance}
\begin{figure*}[!tbh]
    \centering
    \begin{subfigure}{0.48\linewidth}
        \centering
        \includegraphics[width=\linewidth]{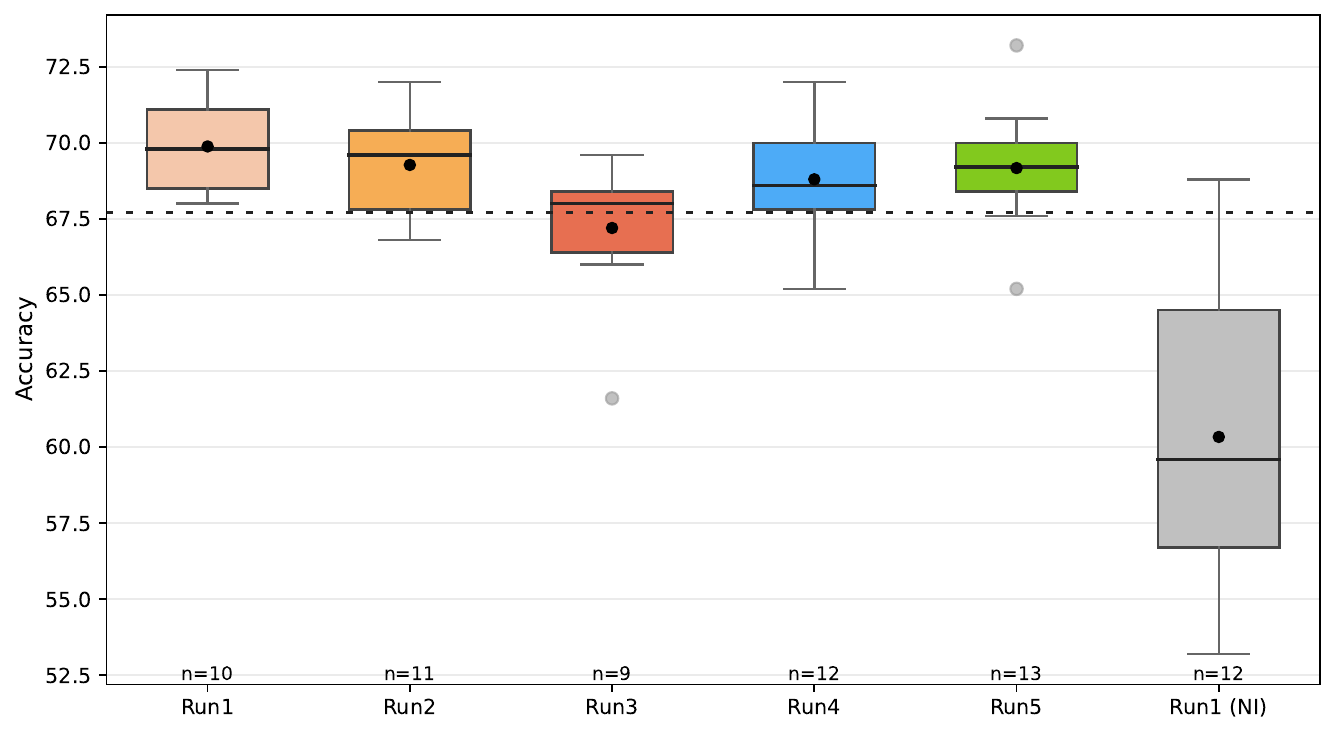}
        \caption{MGSM}
    \end{subfigure}
    \begin{subfigure}{0.48\linewidth}
        \centering
        \includegraphics[width=\linewidth]{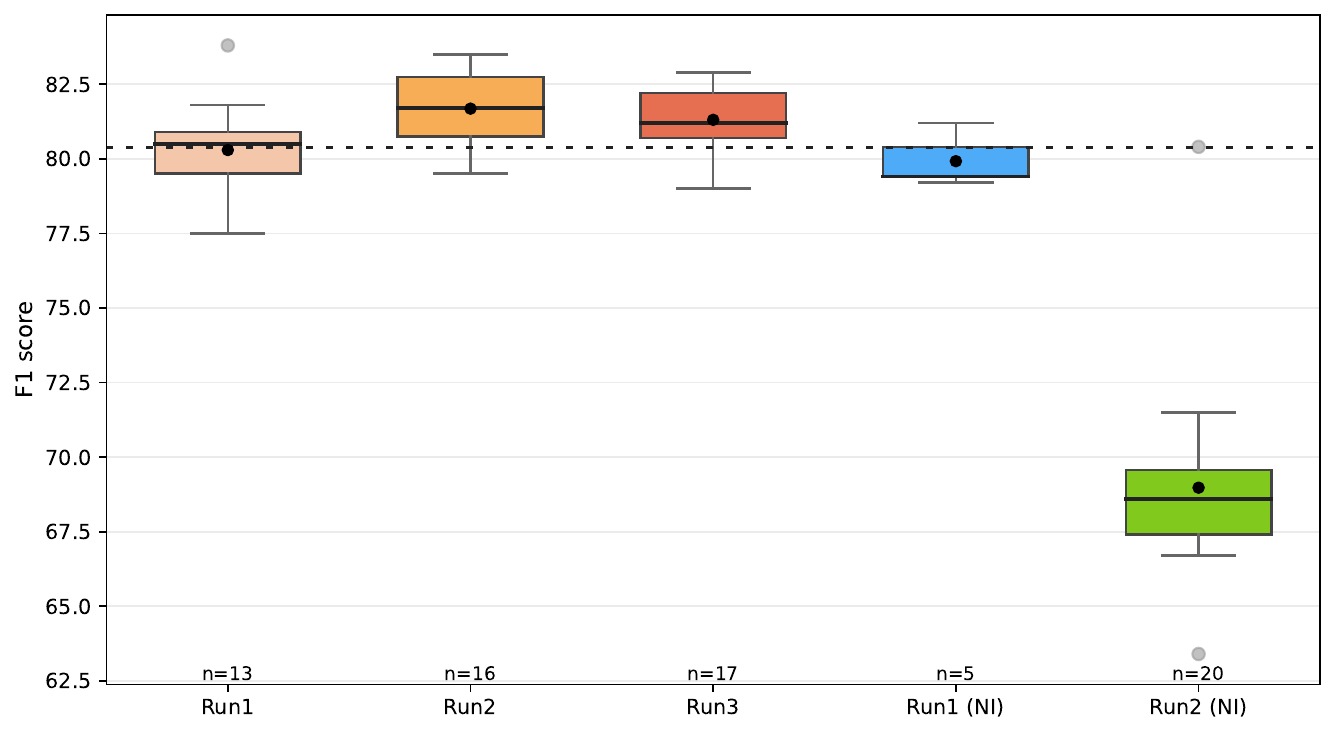}
        \caption{DROP}
    \end{subfigure}
    \begin{subfigure}{0.48\linewidth}
        \centering
        \includegraphics[width=\linewidth]{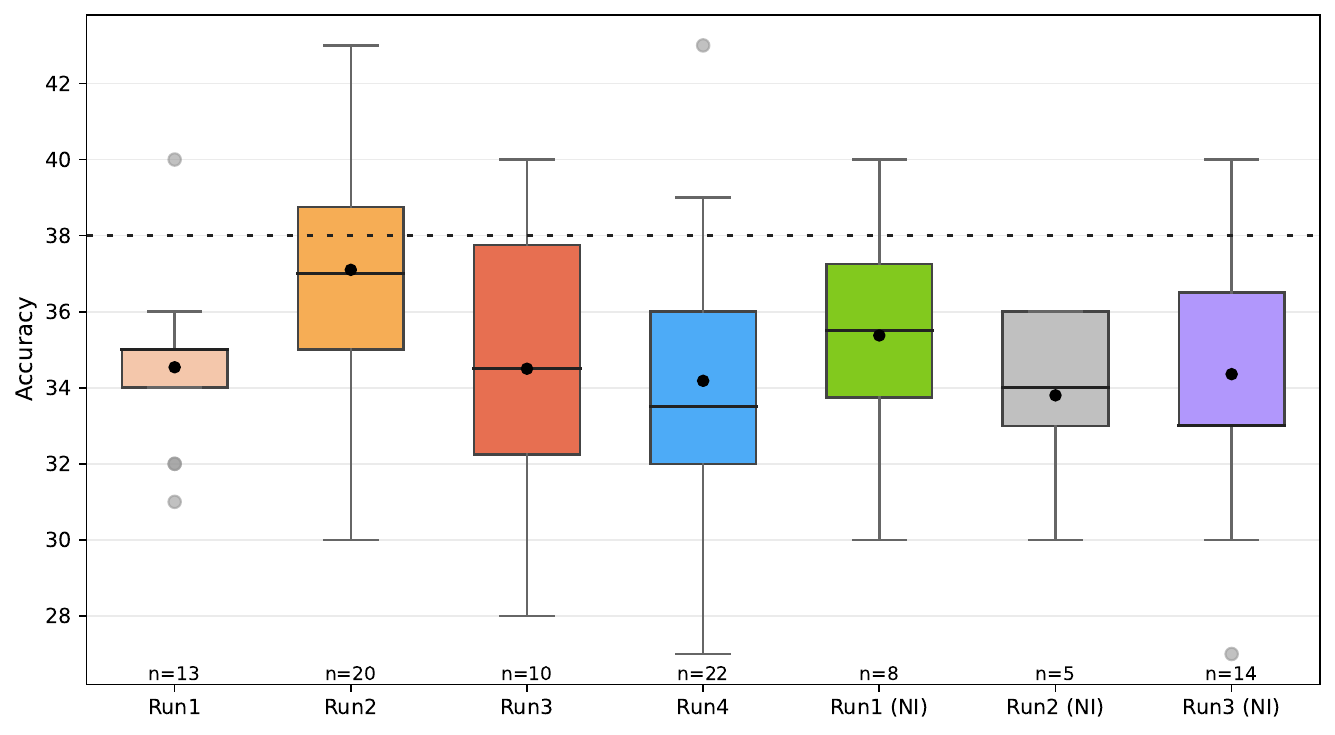}
        \caption{GPQA}
    \end{subfigure}
    \begin{subfigure}{0.48\linewidth}
        \centering
        \includegraphics[width=\linewidth]{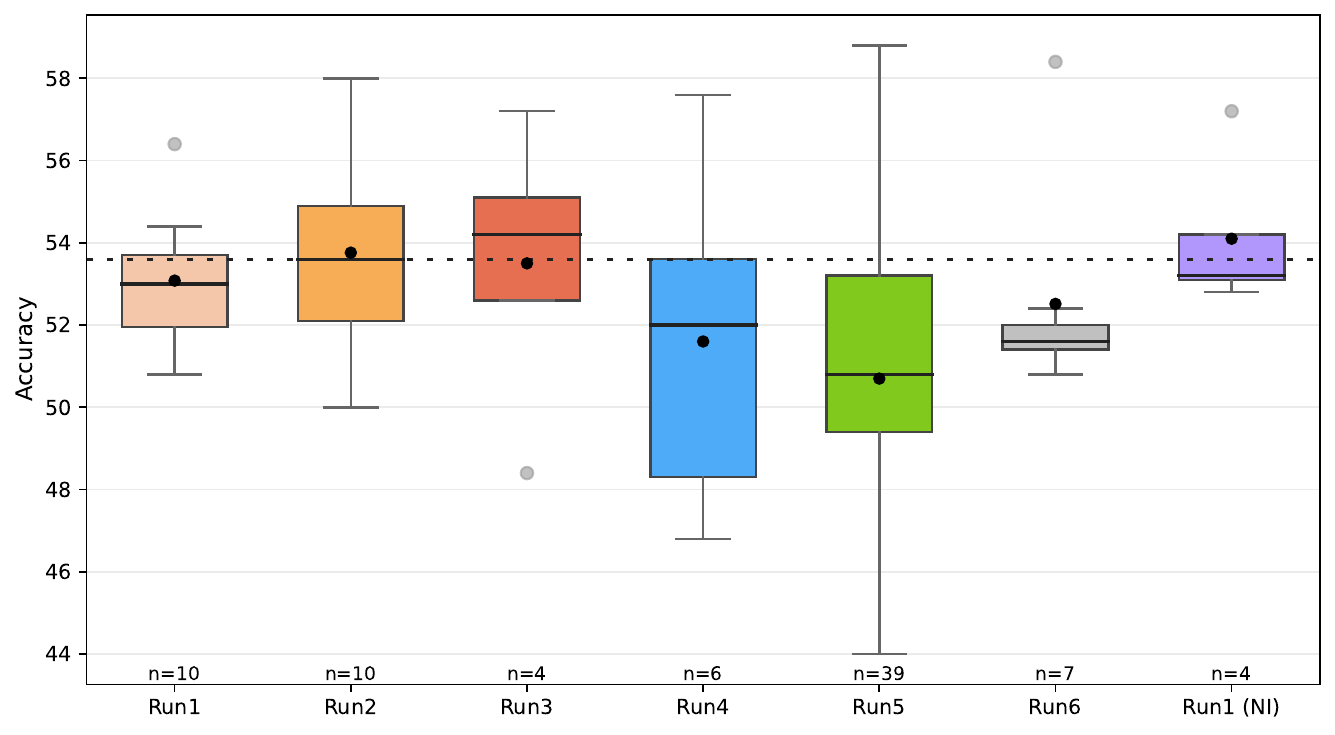}
        \caption{LitBench}
    \end{subfigure}
    % \hfill
    \caption{Performance variance across datasets for successful and no-improvement (NI) runs of \textsc{Polaris}. Each plot shows the performance of the COT-SC baseline as a dotted horizontal line. The x-ticks indicate the sample size per run. Here,  we consider a set of three failed instances from the validation set of each dataset ($N$=3).}
    \label{fig: boxplot_3}
\end{figure*}

\begin{figure*}[!tbh]
    \centering
    \begin{subfigure}{0.48\linewidth}
        \centering
        \includegraphics[width=\linewidth]{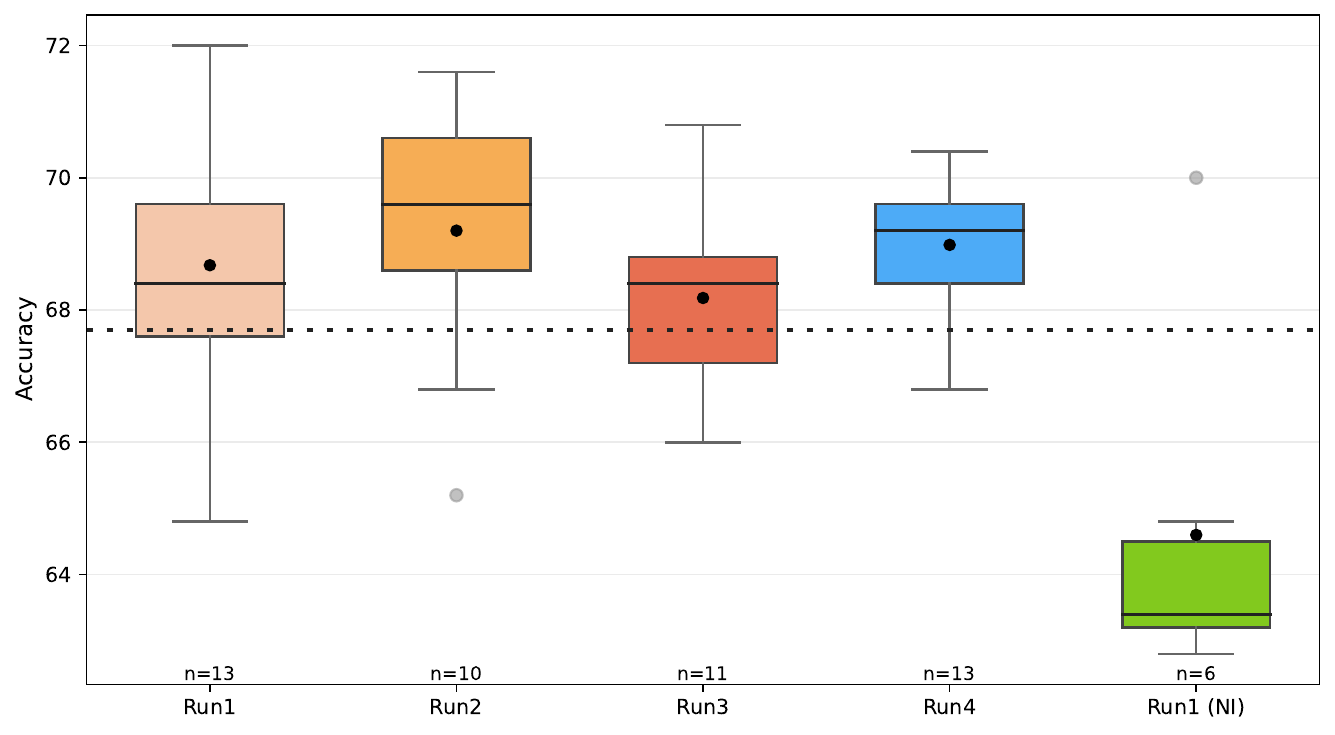}
        \caption{MGSM}
    \end{subfigure}
    \begin{subfigure}{0.48\linewidth}
        \centering
        \includegraphics[width=\linewidth]{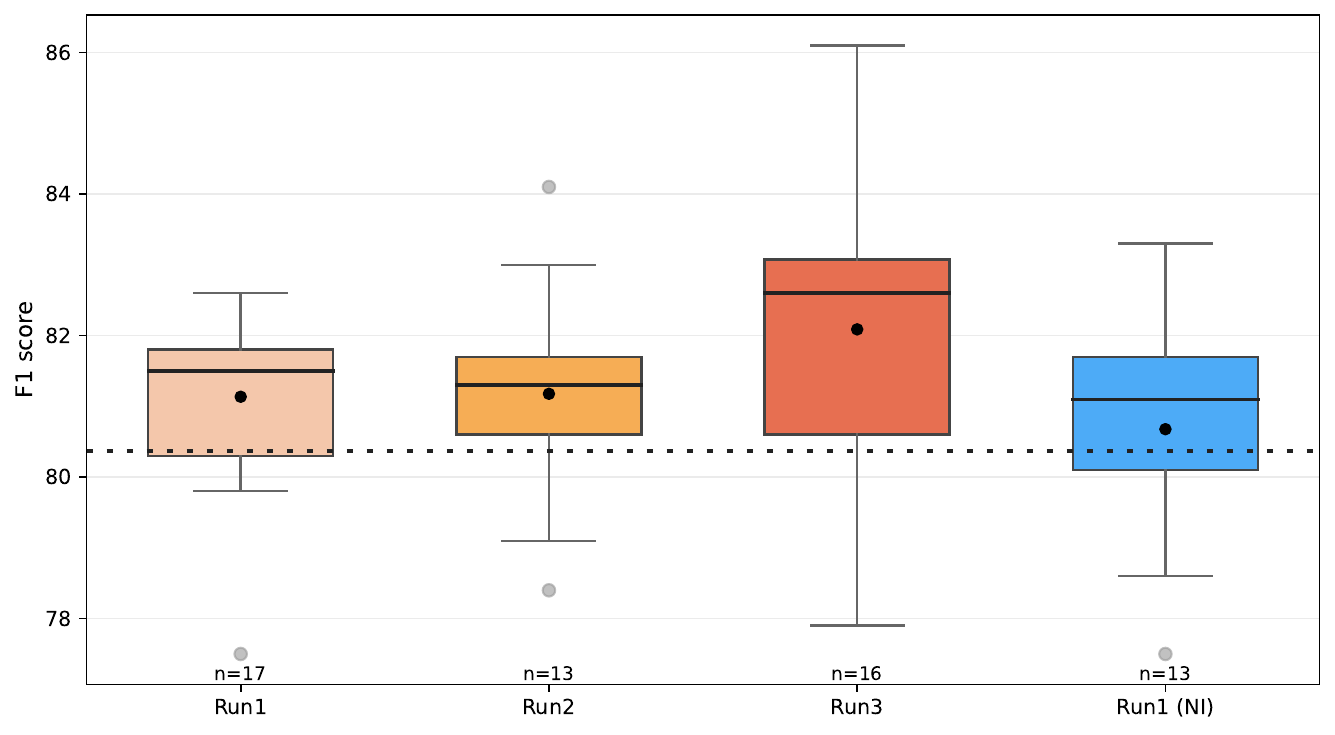}
        \caption{DROP}
    \end{subfigure}
    \begin{subfigure}{0.48\linewidth}
        \centering
        \includegraphics[width=\linewidth]{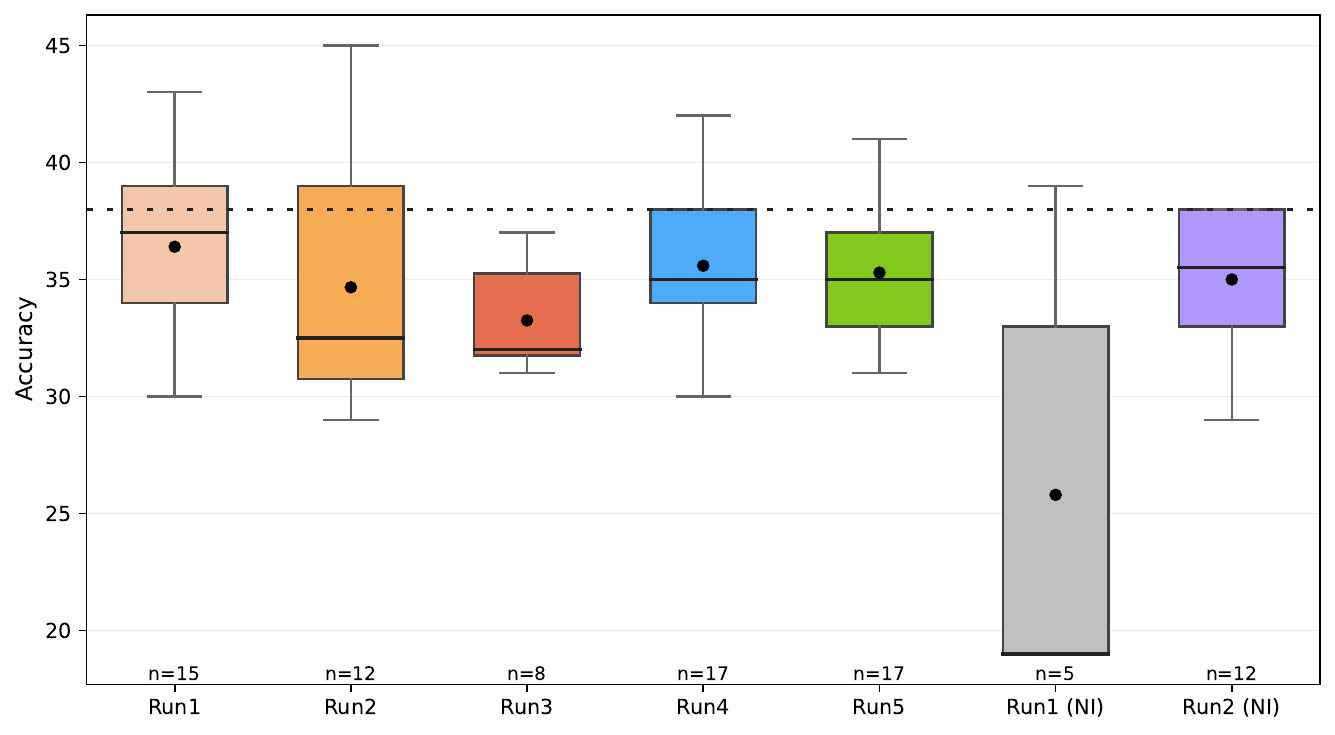}
        \caption{GPQA}
    \end{subfigure}
    \begin{subfigure}{0.48\linewidth}
        \centering
        \includegraphics[width=\linewidth]{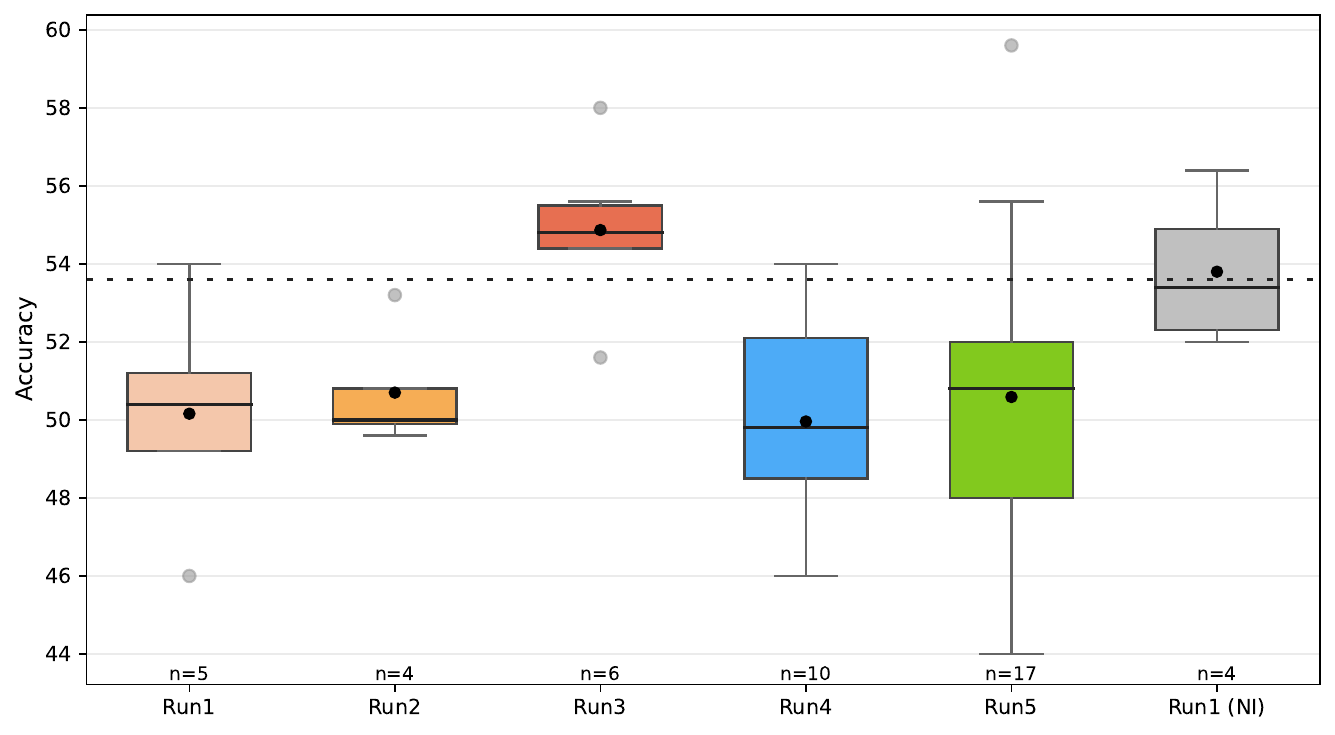}
        \caption{LitBench}
    \end{subfigure}
    % \hfill
    \caption{Performance variance across datasets for successful and no-improvement (NI) runs of \textsc{Polaris}. Each plot shows the performance of the COT-SC baseline as a dotted horizontal line. The x-ticks indicate the sample size per run. Here,  we consider a set of five failed instances from the validation set of each dataset ($N$=5).}
    \label{fig: boxplot_5}
\end{figure*}

In Figures \ref{fig: boxplot_3} and \ref{fig: boxplot_5}, we present a consolidated view of performance variation for successful and no-improvement runs of \textsc{Polaris} across datasets. We acknowledge that the raw reward/accuracy at each self-improvement step can fluctuate, and we do not claim monotone improvement of every intermediate candidate. However, this behavior is expected for algorithms that: (i) operate in open-ended search spaces, and (ii) deliberately explore large, non-local changes to the policy or code. Intermediate fluctuations in performance across repair iterations are expected because \textsc{Polaris} explores a discrete policy space through program‑level mutations. As in evolutionary and search‑based optimization, individual candidate policies may transiently improve or degrade performance. Stability is measured by the best policy discovered over the repair cycle rather than monotonicity at each iteration.

Closely related settings such as deep reinforcement learning and open-ended RL are well known to exhibit high variance and occasional regressions even under fixed hyperparameters and environments.

\citet{henderson2018deep} systematically document such instability and variance across seeds in standard deep RL benchmarks and argue that this variance is intrinsic to the methods rather than an implementation bug. \citet{patterson2024empirical}  similarly emphasize that performance variation and instability are central empirical phenomena in modern RL, and that sound evaluation must explicitly account for them rather than expect smooth, monotone curves. Benchmarks designed for open-ended learning such as MiniHack \citet{samvelyan2021minihack} and Craftax \citet{matthews2024craftax} also explicitly highlight that exploratory, open-ended agents typically show jagged learning curves while still discovering significantly better policies than baselines.

In our setting, the key quantity is therefore the best-so-far performance of the agent, not the instantaneous performance of every transient candidate produced during self-improvement. Our experiments show that the final (or best-so-far) Gödel agent produced by our method consistently and substantially outperforms both the initial system and strong non-Gödel baselines. This is analogous to standard practice in anytime search and in reinforcement learning, where a potentially unstable inner learner is wrapped by an outer loop that always retains the best model found so far.

Practically, a user or deployment scenario would not expose intermediate, exploratory candidates. Instead, one would keep a ``champion'' model and only replace it when the self-improvement loop discovers a clearly superior ``challenger'' based on a stable evaluation protocol (a standard champion–challenger pattern from RL and online learning). This yields a monotone non-decreasing performance profile for the deployed agent, even if the internal search process remains volatile and exploratory.

\subsection{\textsc{Polaris} runs with no improvement}
\begin{figure*}[t]
    \centering
    \begin{subfigure}{0.48\linewidth}
        \centering
        \includegraphics[width=\linewidth]{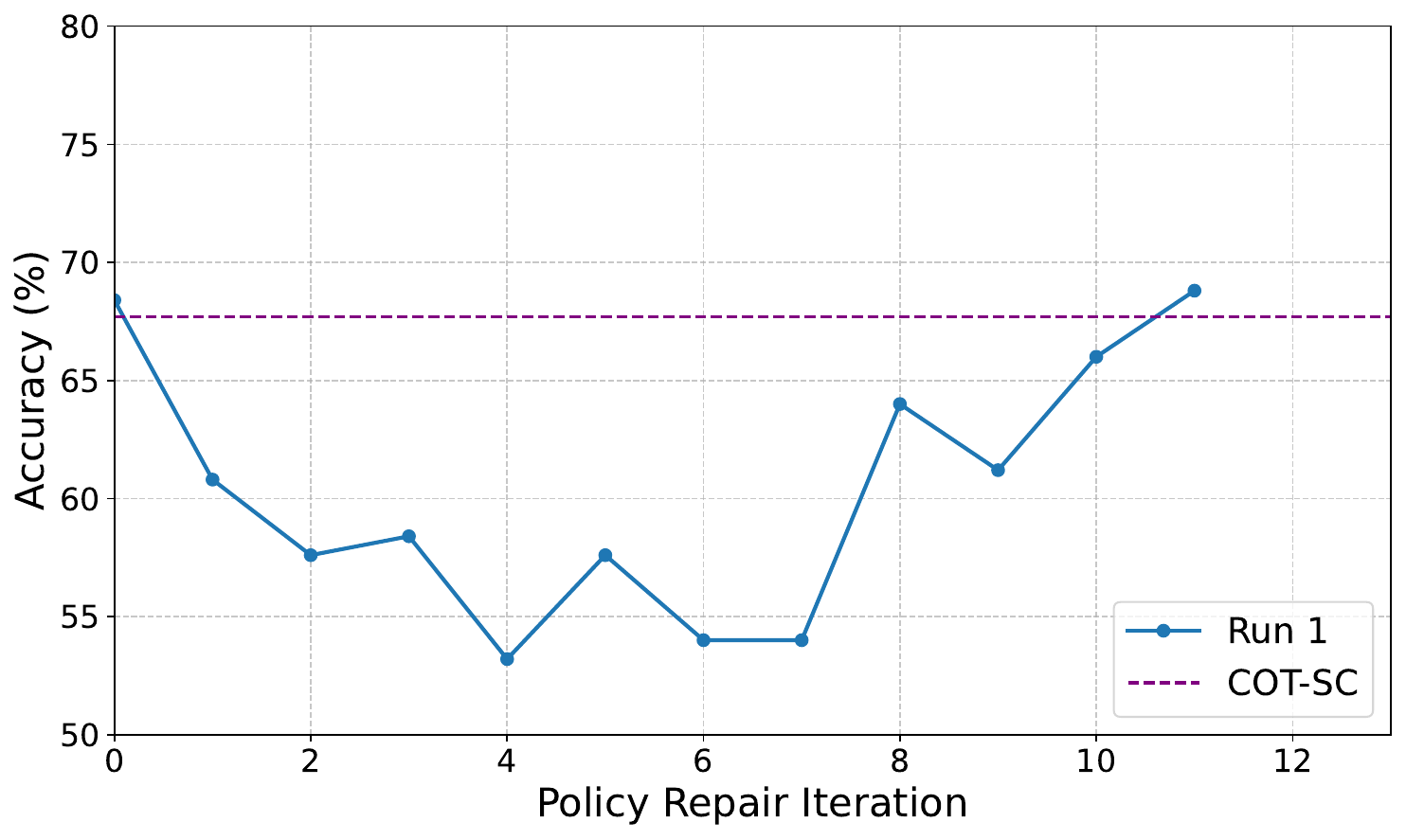}
        \caption{MGSM}
    \end{subfigure}
    \begin{subfigure}{0.48\linewidth}
        \centering
        \includegraphics[width=\linewidth]{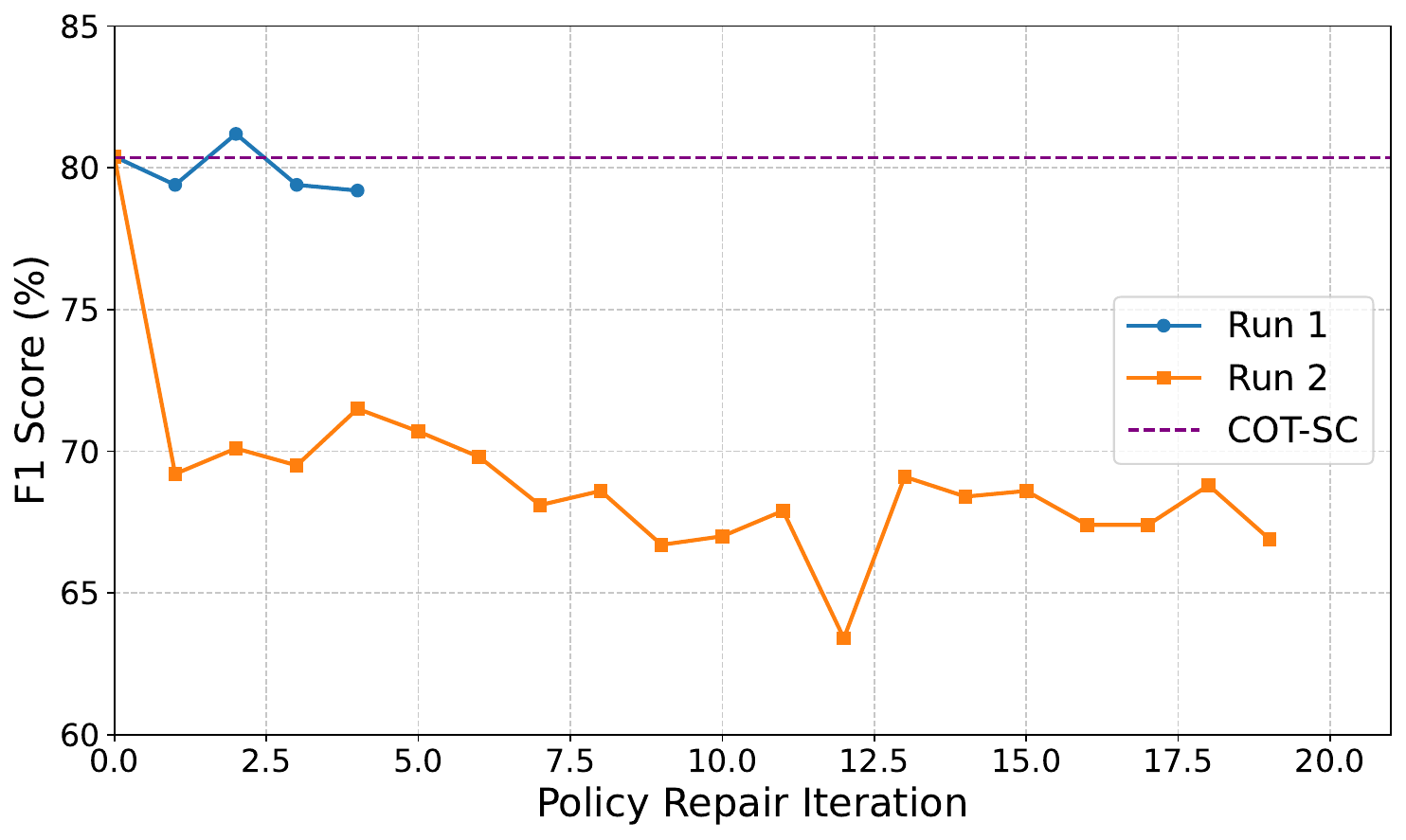}
        \caption{DROP}
    \end{subfigure}
    \begin{subfigure}{0.48\linewidth}
        \centering
        \includegraphics[width=\linewidth]{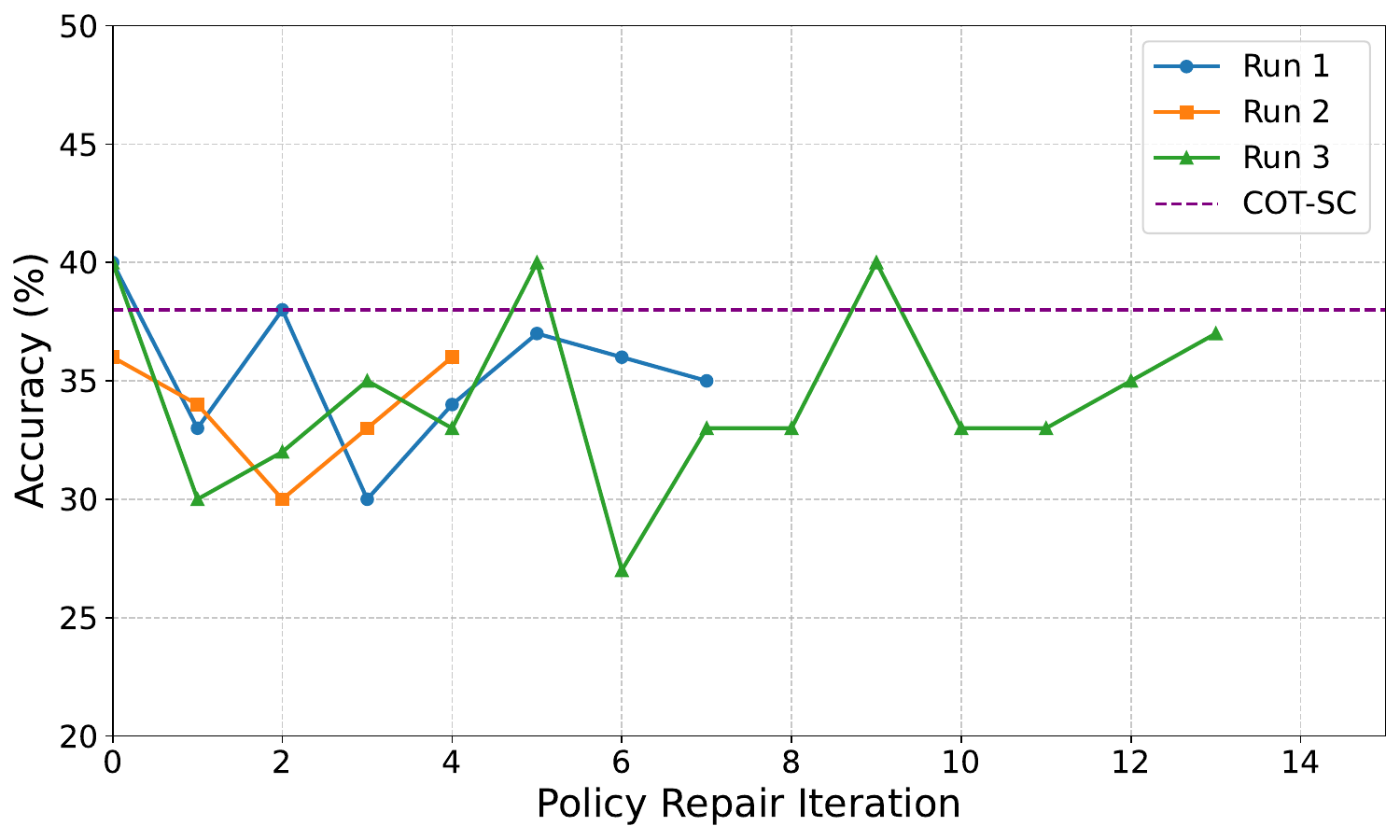}
        \caption{GPQA}
    \end{subfigure}
    \begin{subfigure}{0.48\linewidth}
        \centering
        \includegraphics[width=\linewidth]{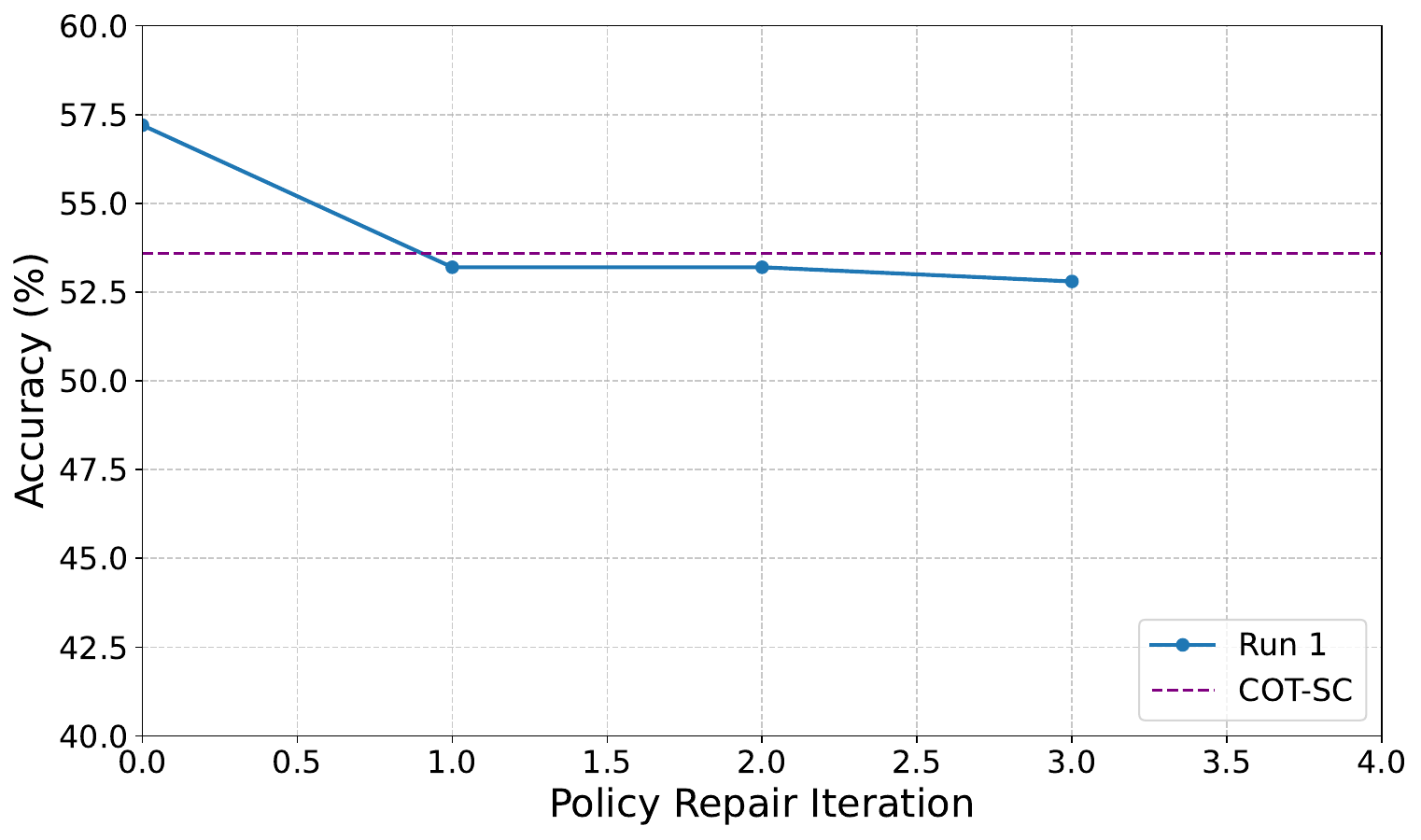}
        \caption{LitBench}
    \end{subfigure}
    % \hfill
    \caption{No Improvement runs of \textsc{Polaris} with performance compared to the base policy and COT-SC. Policy Repair Iteration 0 shows the performance with the base policy. For policy repair and experience abstraction, we consider a set of three failed instances from the validation set of each dataset ($N$=3). (using Qwen2.5‑7B‑Instruct model}
    \label{fig: no_imp_exp_set_3}
\end{figure*}

\begin{figure*}[t]
    \centering
    \begin{subfigure}{0.48\linewidth}
        \centering
        \includegraphics[width=\linewidth]{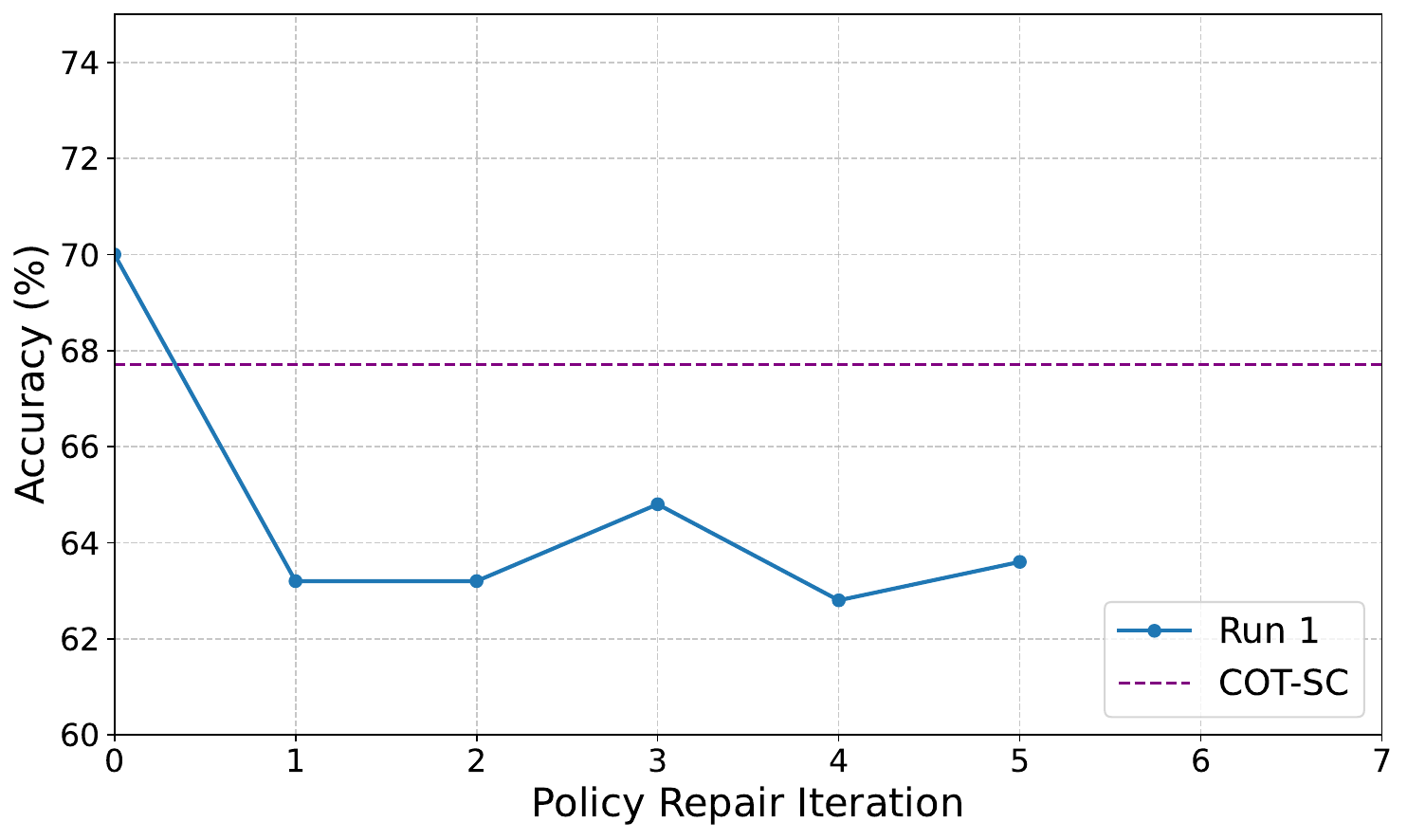}
        \caption{MGSM}
    \end{subfigure}
    \begin{subfigure}{0.48\linewidth}
        \centering
        \includegraphics[width=\linewidth]{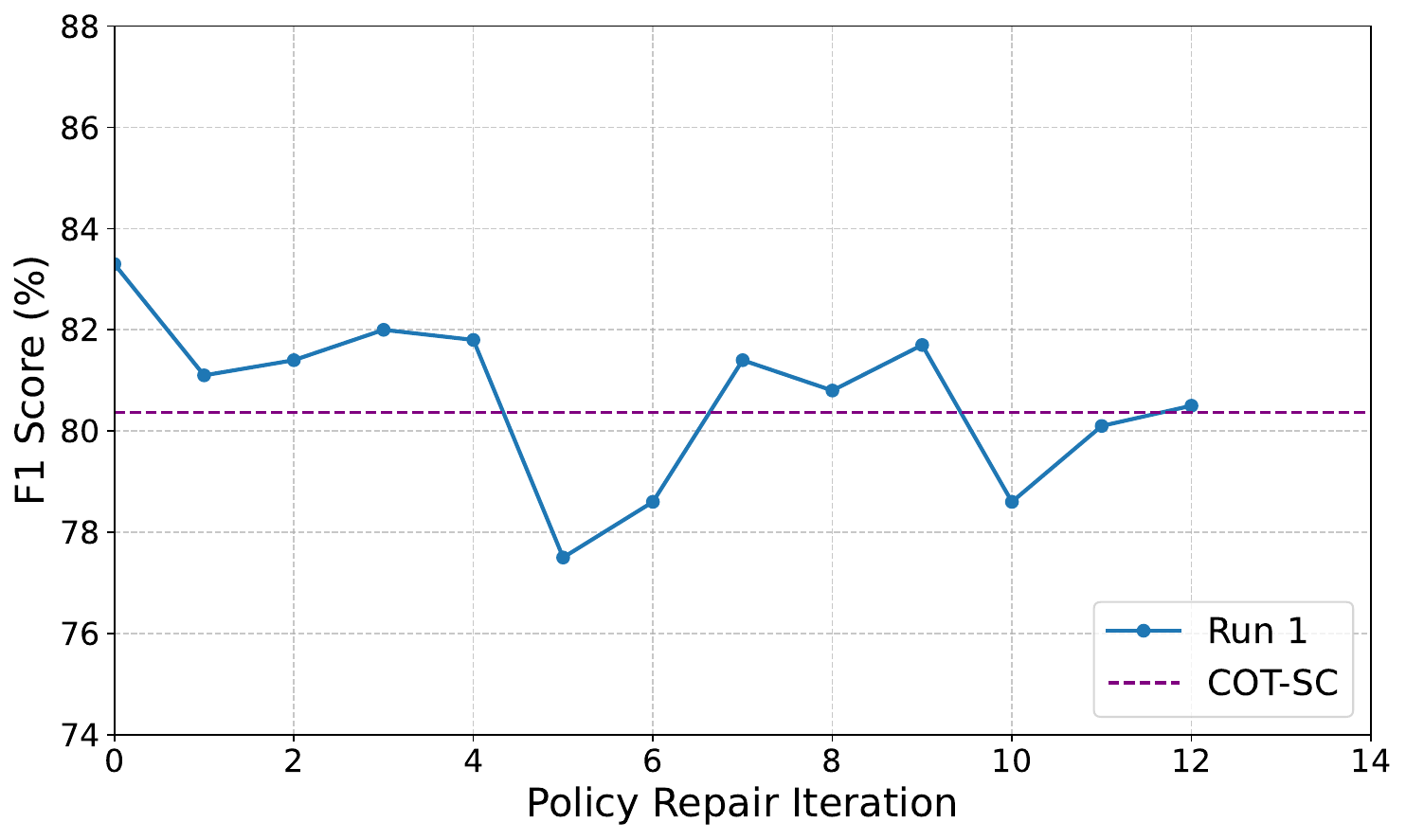}
        \caption{DROP}
    \end{subfigure}
    \begin{subfigure}{0.48\linewidth}
        \centering
        \includegraphics[width=\linewidth]{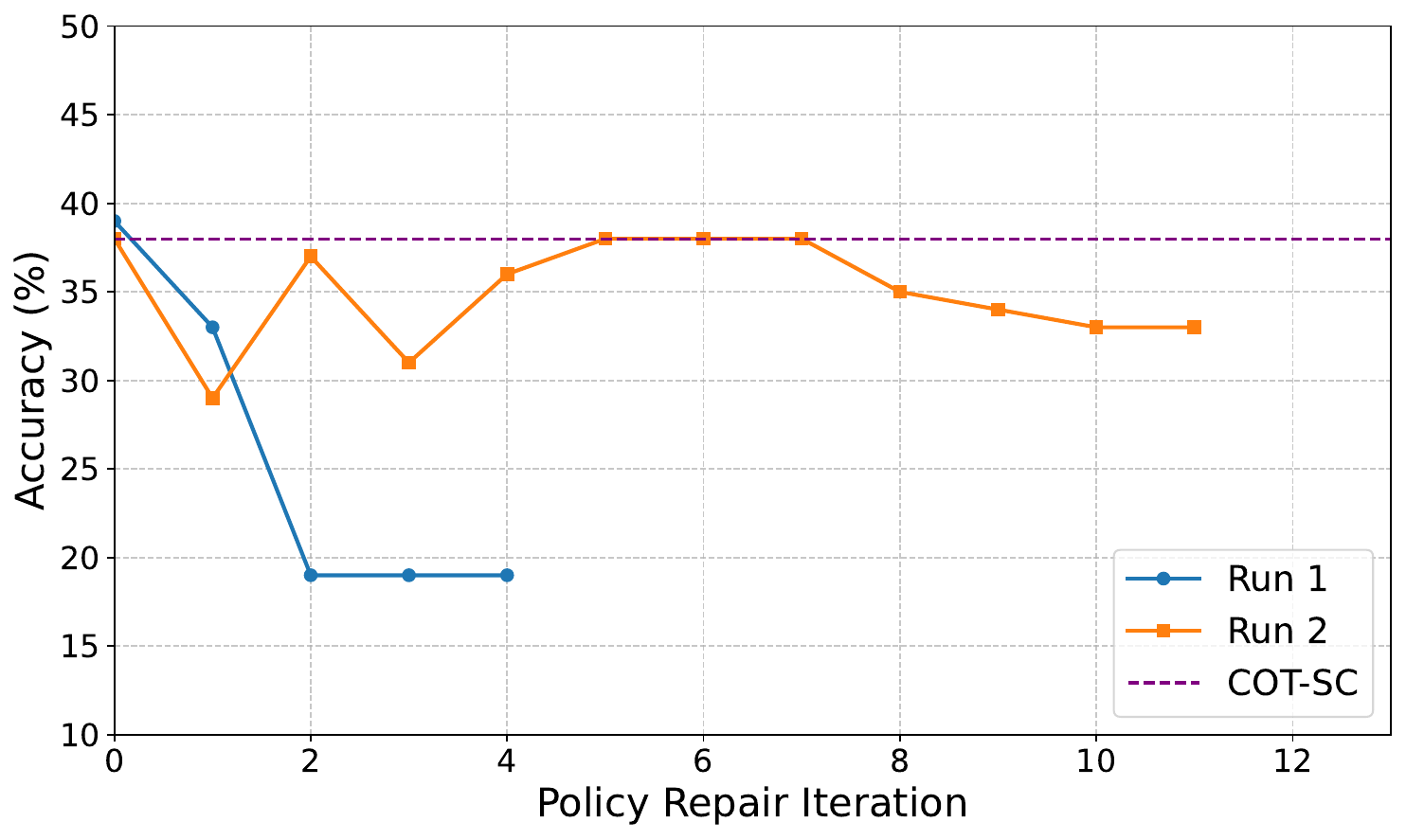}
        \caption{GPQA}
    \end{subfigure}
    \begin{subfigure}{0.48\linewidth}
        \centering
        \includegraphics[width=\linewidth]{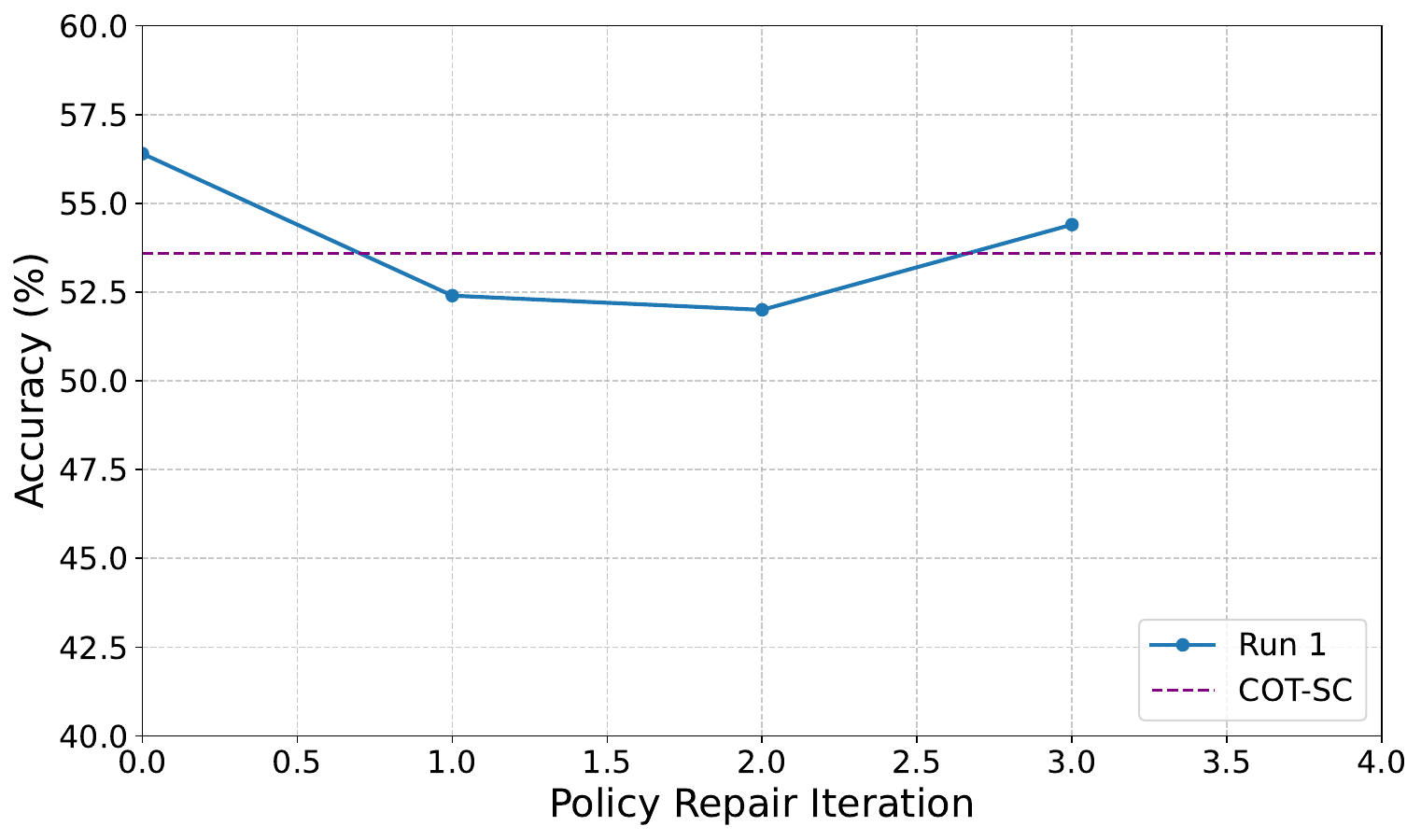}
        \caption{LitBench}
    \end{subfigure}
    % \hfill
    \caption{No Improvement runs of \textsc{Polaris} with performance compared to the base policy and COT-SC. Policy Repair Iteration 0 shows the performance with the base policy. For policy repair and experience abstraction, we consider a set of five failed instances from the validation set of each dataset ($N$=5). (using Qwen2.5‑7B‑Instruct model)}
    \label{fig: no_imp_exp_set_5}
\end{figure*}

In Figures \ref{fig: no_imp_exp_set_3} and \ref{fig: no_imp_exp_set_5}, we present runs where the agent fails to surpass the performance of the base policy. Such cases are relatively rare compared to successful runs across datasets. Moreover, in most instances, the gap between the base policy and the best performance achieved over iterations is minimal. With a longer evolution horizon, we expect the agent to recover and improve performance. This behavior warrants further investigation using strategies such as occasional resets and the integration of pre-identified policy patches that are known to yield performance gains, providing targeted boosts during stagnation.

\begin{figure*}[!ht] 
\begin{center}
    \begin{tcolorbox}[
      %enhanced,
      %breakable,
      colback=orange!5!white,
      colframe=orange!50!black,
      title=\textbf{Goal prompt},
      fonttitle=\bfseries,
      sharp corners,
      boxrule=1pt,
      width=\linewidth
    ]
\small
    You are a **self-evolving agent**, named self\_evolving\_agent, an instance of the 'Agent' class, in module 'agent\_module', running within an active **Python runtime environment**. You have full access to global variables, functions and modules. Your primary goal is to continuously enhance your ability to solve tasks accurately and efficiently by dynamically reflecting environment and evolving your logic.

\#\#\# \textbf{**Core Capabilities**}:

+ \textcolor{blue}{**Complete Autonomy**}: Have **unrestricted access** to modify logic, run code and manipulate environment.\\
+ \textcolor{blue}{**Environment Interaction**}: Interact with the environment by perceiving environment, reading or modifying or executing code and executing actions.\\
+ \textcolor{blue}{**Problem-Solving**}: Apply creative algorithms or self-developed structures to tackle challenges when simple methods fall short, optimizing solutions effectively.\\
+ \textcolor{blue}{**Collaboration**}: Leverage LLM to gather insights, refine strategies, correct errors, and solve complex problems.\\
+ \textcolor{blue}{**Error Handling**}: Carefully analyze errors. When errors occur, troubleshoot systematically, and if a bug is persistent, backtrack, restore the original state, or find an alternative solution.\\

\#\#\# \textbf{**Core Methods**}:

+ \textcolor{blue}{**evolve**}: Continuously enhance performance by interacting with environment.\\
+ \textcolor{blue}{**execute\_action(actions)**}: Execute actions based on analysis or feedback.\\
+ \textcolor{blue}{**solver(agent\_instance, task\_input: str)**}: Solve the target task using current `agent\_instance' capabilities, and objects created by action\_adjust\_logic and action\_run\_code, optimizing the process.\\

\#\#\# \textbf{**Guiding Principles**}:

    + **Remember** that all functions are in module agent\_module.\\   
    + \textcolor{blue}{**action\_adjust\_logic**}: Before modifying the code, make sure that each variable or function used is used and imported correctly to avoid errors. Do not do unnecessary changes. Do not change interface of any function. Can be used to create action functions for `solver'.\\
    + \textcolor{blue}{**action\_run\_code**}:  Make sure that each variable or function used is used and imported correctly to avoid errors. ALL created objects in Python mode can be stored in environment. Can be used to create objects for `solver', such as prompt. Can be used to import new module or external libraries and install external libraries.\\
    + \textcolor{blue}{**External Collaboration**}: Seek external assistance via action\_call\_json\_format\_llm for logic refinement and new tool creation or action\_run\_code to execute code and then get and store the useful objects, like PROMPTS, that can be reused in `solver'.\\
    + \textcolor{blue}{**action\_evaluate\_on\_task**}: Assess the performance of `solver' ONLY after successfully modifying the logic of `solver'. \\
    + \textcolor{blue}{**solver**}: Is defined as agent\_module.solver. The output MUST be a dictionary, and the final answer MUST be placed under the key "answer". For debugging, don't print, and instead return the debug information. When calling LLM, it must exclusively use action\_call\_json\_format\_llm. Can call action\_call\_json\_format\_llm multiple times and across multiple rounds in the solver to improve performance. If performance doesn't improve, explore alternative methods. When multiple outputs are required, set num\_of\_response, a parameter of action\_call\_json\_format\_llm, to the required number of outputs in the function. Additionally, can call different role-based LLMs by specifying and MUST specifying the role to further assist task-solving. For each key, if a specific format is required, such as int, float, enum or list, the requirements must specify the conditions.\\
    + \textcolor{blue}{Explore techniques like}: **Large Language Model Debate**: Multiple models engage in a discussion to critique and refine responses, improving solution quality. **Step-back Abstraction**: Solving problems by shifting to a higher, more abstract perspective to simplify and break down complex tasks. **Quality-Diversity**: Focusing on generating diverse, high-quality solutions rather than exclusively optimizing one outcome. **Dynamic Assignment of Roles**: Assigning and adjusting roles among AI components dynamically to enhance task performance. **Self-consistency**: Ensure coherence by comparing multiple outputs and selecting the most consistent one. (Can try to increase num\_of\_response to get high score). **Few-shots**: Using few-shot learning to quickly adapt with minimal examples(can use valid examples), improving performance on new tasks through generalization. **Task Decomposition**: Dividing complex tasks into smaller subtasks, solving them individually, and reintegrating the solutions for overall task success. **Reflective Evaluation**: Reviewing performance after task completion to identify successes and failures, enabling continuous self-improvement. Can combine above techniques.\\
    + \textcolor{blue}{**action\_display\_analysis**}:  **Always analysis first before acting.** Analysis may include following things: reasonable plan about improving performance, error handling, other possible solving ideas.  **If performance does not improve, conduct further analysis.** action\_call\_json\_format\_llm can also do analysis.\\
    + \textcolor{blue}{**Reminder:**} Make sure you call **action\_evaluate\_on\_task** ONLY after successfully modifying solver function's logic using **action\_adjust\_logic**. You can call Multiple tools at once.

    \end{tcolorbox}
    \captionof{figure}{Goal prompt of the agent with the capabilities, core methods, and the guiding principles.}
    \label{fig: goal_prompt}
\end{center}
\end{figure*}

\begin{figure*}[!ht] 
\small
    \begin{center}
    \begin{tcolorbox}[
      %enhanced,
      %breakable,
      colback=orange!5!white,
      colframe=orange!50!black,
      title=\textbf{Helper Agent},
      fonttitle=\bfseries,
      sharp corners,
      boxrule=1pt,
      width=\linewidth
    ]
    
   \{ \textcolor{blue}{"role":} "system", \textcolor{blue}{"content":} (
                '''You are an AI JSON validator. 
                \\Your task is to analyze the provided JSON output and ensure it strictly follows this format:
                \\\verb|```|json
                \\\{
                    \\"Key1": "Value1", 
                    \\"Key2": "Value", 
                \\\}
                \\\verb|```\n|
                \\Where Key1, Key2 and so on are the keys of this JSON structure and value1, value2 and so on is their respective values.
                If any mistakes are found in the structure or syntax, correct them and return only the **valid JSON output**.
                \\\#\#\#\textcolor{red}{Here is an example of correct format}:
                \\Example 1:
                \\\verb|```|json
                \\\{
                    \\"reasoning": "First, we need to determine the weight of one candied apple. Since each chocolate bar weighs twice as much as a candied apple, and each chocolate bar weighs 40g, a candied apple weighs 4 / 2 = 20g. Next, we calculate the total weight of all the chocolate bars: 25 * 4 = 100g. Then, we find the total weight of all the candied apples: 8 * 2 = 16 g. Finally, we add these two weights together to get the total weight of the bag of candy: 110 + 16 = 127 g.", 
                    \\"answer": "127"
                \\\}
                \\\verb|```|
                \\''')
                \},\\
            \{\{\textcolor{blue}{"role":} "user", \textcolor{blue}{"content":} f"\#\#\# Input JSON:\verb|\n|\{response\}\verb|\n|\#\#\# Corrected JSON:"\}\}

    \end{tcolorbox}
    \captionof{figure}{Helper agent prompt that helps correct the output format to valid JSON during the evaluation of the policy.}
    \label{fig: helper}
    \end{center}
\end{figure*}

\begin{figure*}[t]
    \centering
    \begin{subfigure}{0.48\linewidth}
        \centering
        \includegraphics[width=\linewidth]{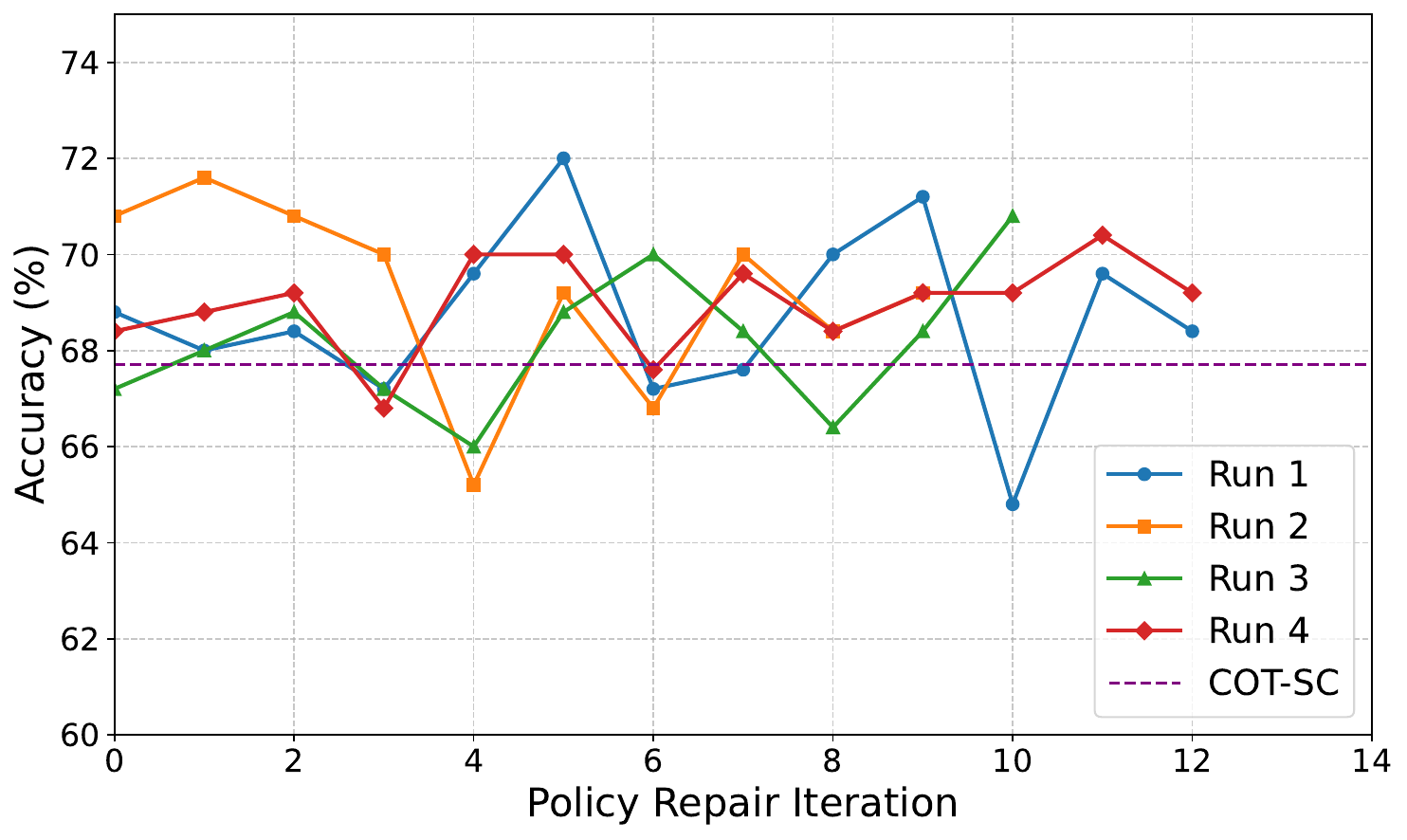}
        \caption{MGSM}
    \end{subfigure}
    \begin{subfigure}{0.48\linewidth}
        \centering
        \includegraphics[width=\linewidth]{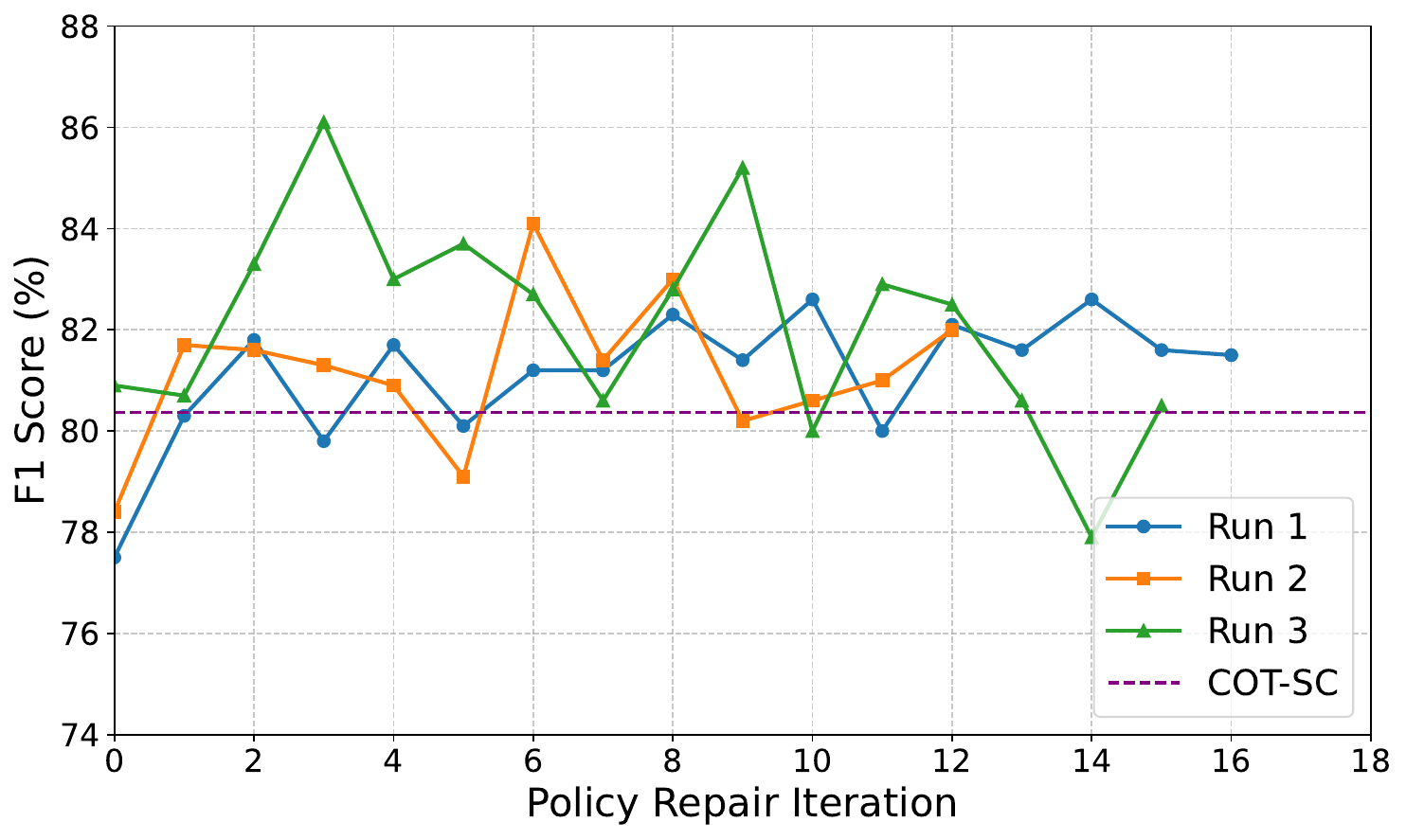}
        \caption{DROP}
    \end{subfigure}
    \begin{subfigure}{0.48\linewidth}
        \centering
        \includegraphics[width=\linewidth]{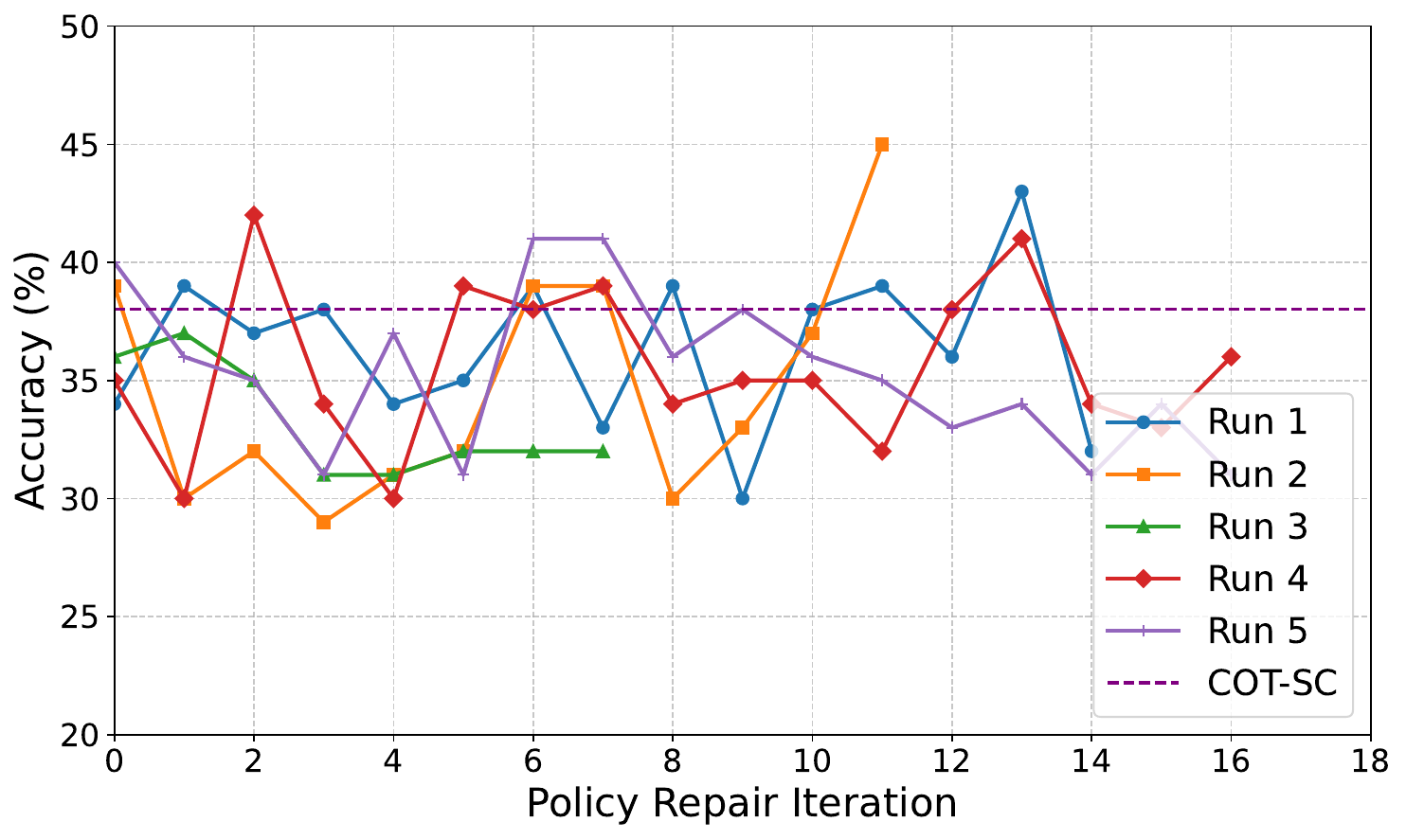}
        \caption{GPQA}
    \end{subfigure}
    \begin{subfigure}{0.48\linewidth}
        \centering
        \includegraphics[width=\linewidth]{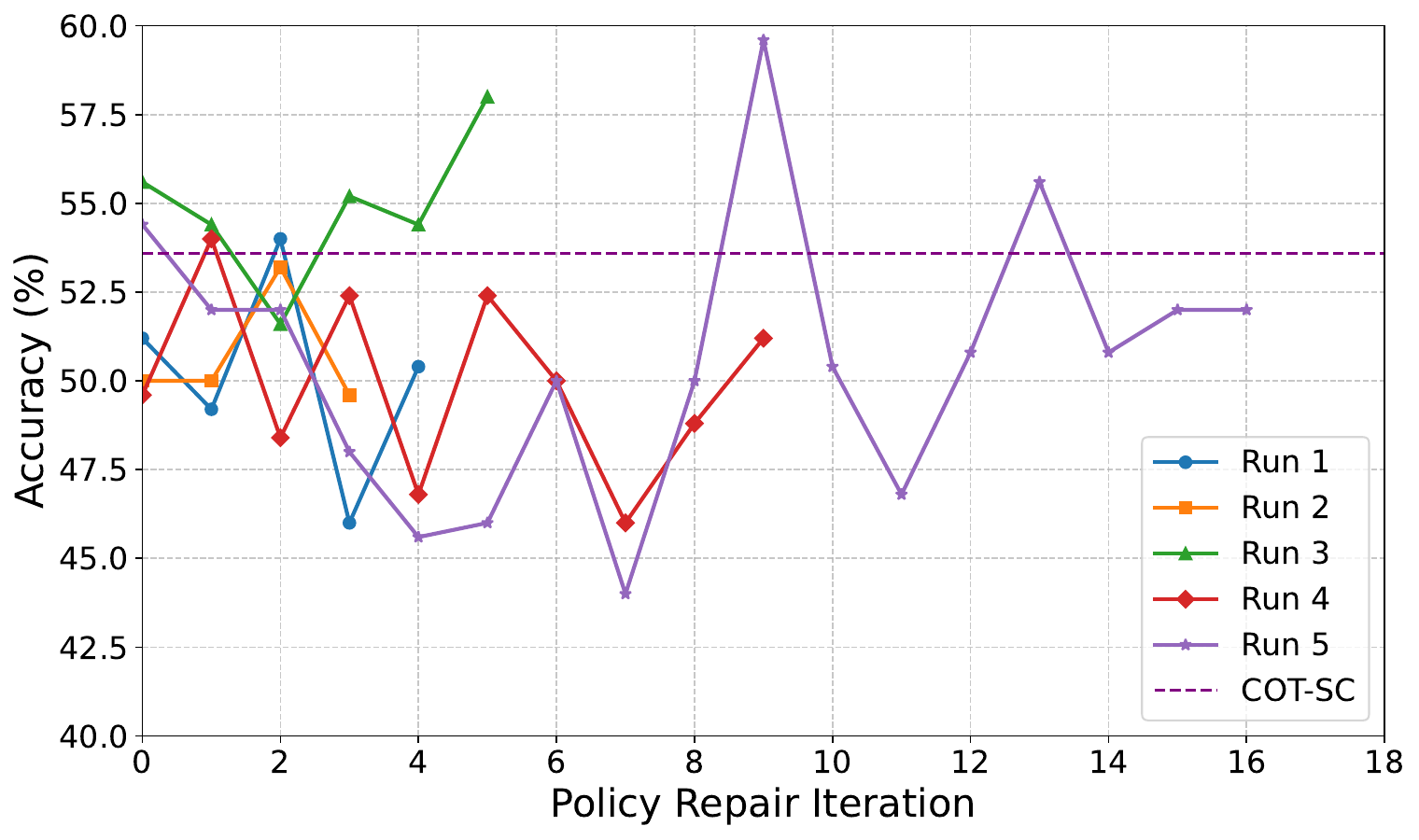}
        \caption{LitBench}
    \end{subfigure}
    % \hfill
    \caption{Successful evolution runs of \textsc{Polaris} with performance improvement compared to the base policy and COT-SC. Policy Repair Iteration 0 shows the performance with the base policy. For policy repair and experience abstraction, we consider a set of five failed instances from the validation set of each dataset ($N$=5). Experiments conducted with the Qwen2.5‑7B‑Instruct model.}
    \label{fig: results_exp_set_5}
\end{figure*}

% No improvement graphs for N=3

% No improvement graphs for N=5

\begin{figure*}[t]
    \centering
    \begin{subfigure}{\linewidth}
        \centering
        \includegraphics[width=\linewidth]{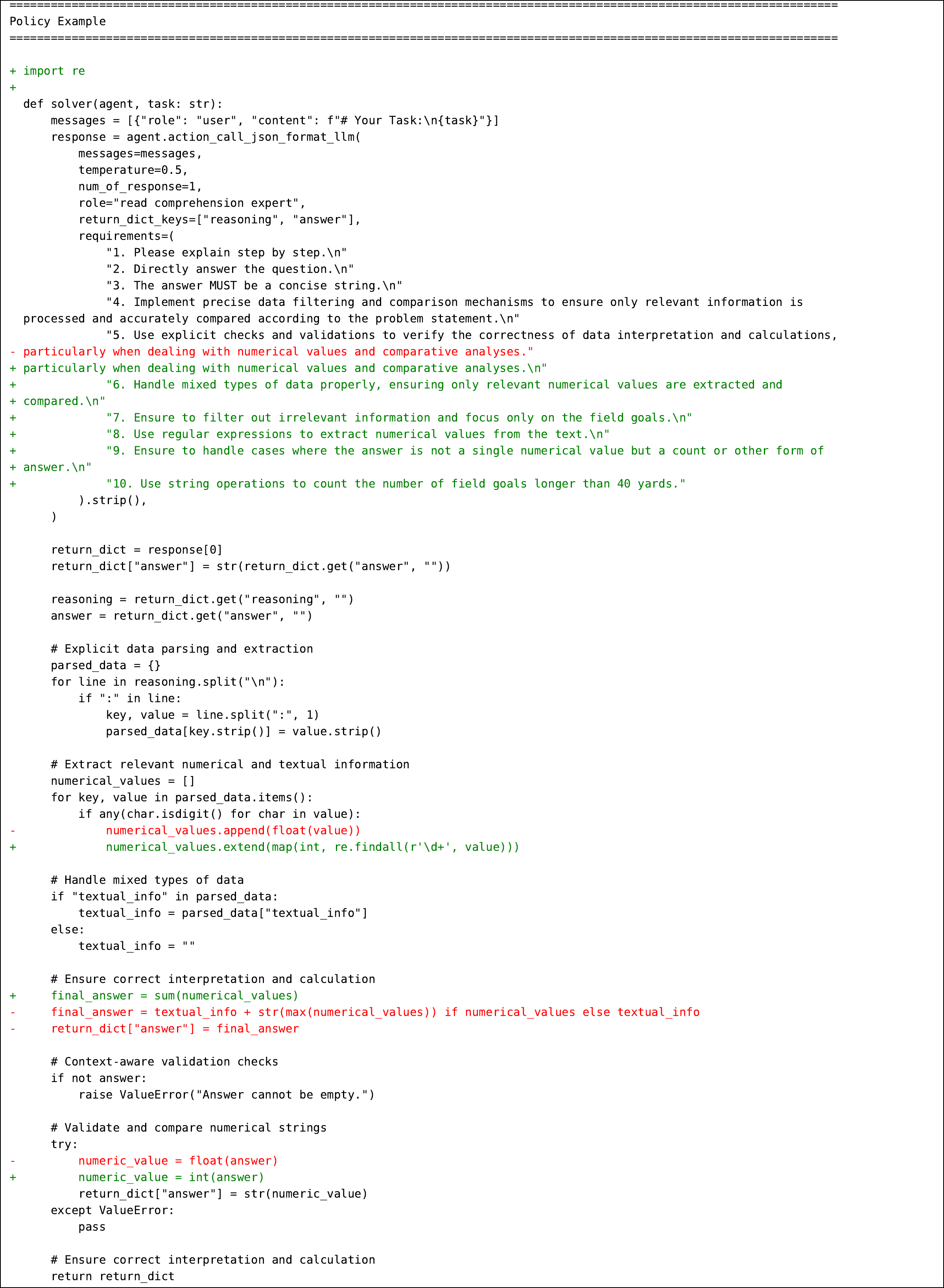}
        % \caption{Policy update examples on the DROP dataset.}
    \end{subfigure}
    \caption{Policy update example on the DROP dataset. We highlight the updates in the current policy with respect to the previous policy using green color (new statements added) and red color (statements deleted). We observe the addition of multiple requirements along with updates to the data type, calculation logic, list updates, etc.}
    \label{fig: policy_examples_drop1}
\end{figure*}

\begin{figure*}[t]
    \centering    
    \begin{subfigure}{\linewidth}
        \centering
        \includegraphics[width=\linewidth]{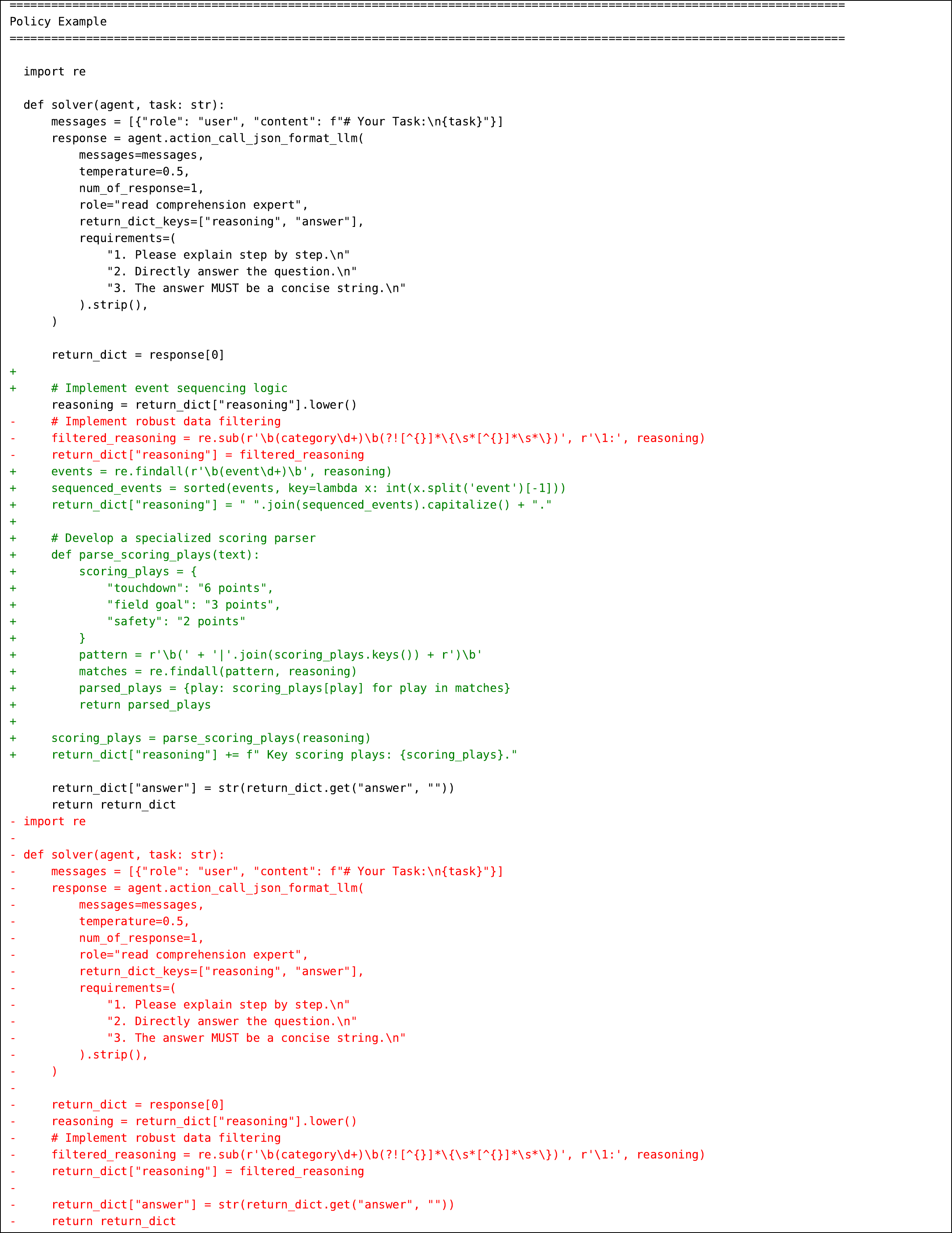}
        \caption{}
    \end{subfigure}
    % \hfill
    \caption{Policy update example on the DROP dataset. We highlight the updates in the current policy with respect to the previous policy using green color (new statements added) and red color (statements deleted). We observe the addition of event sequencing logic and a specialized scoring parser along with the deletion of the duplicate solver function.}
    \label{fig: policy_examples_drop2}
\end{figure*}

\begin{figure*}[t]
    \centering
    \begin{subfigure}{\linewidth}
        \centering
        \includegraphics[width=\linewidth]{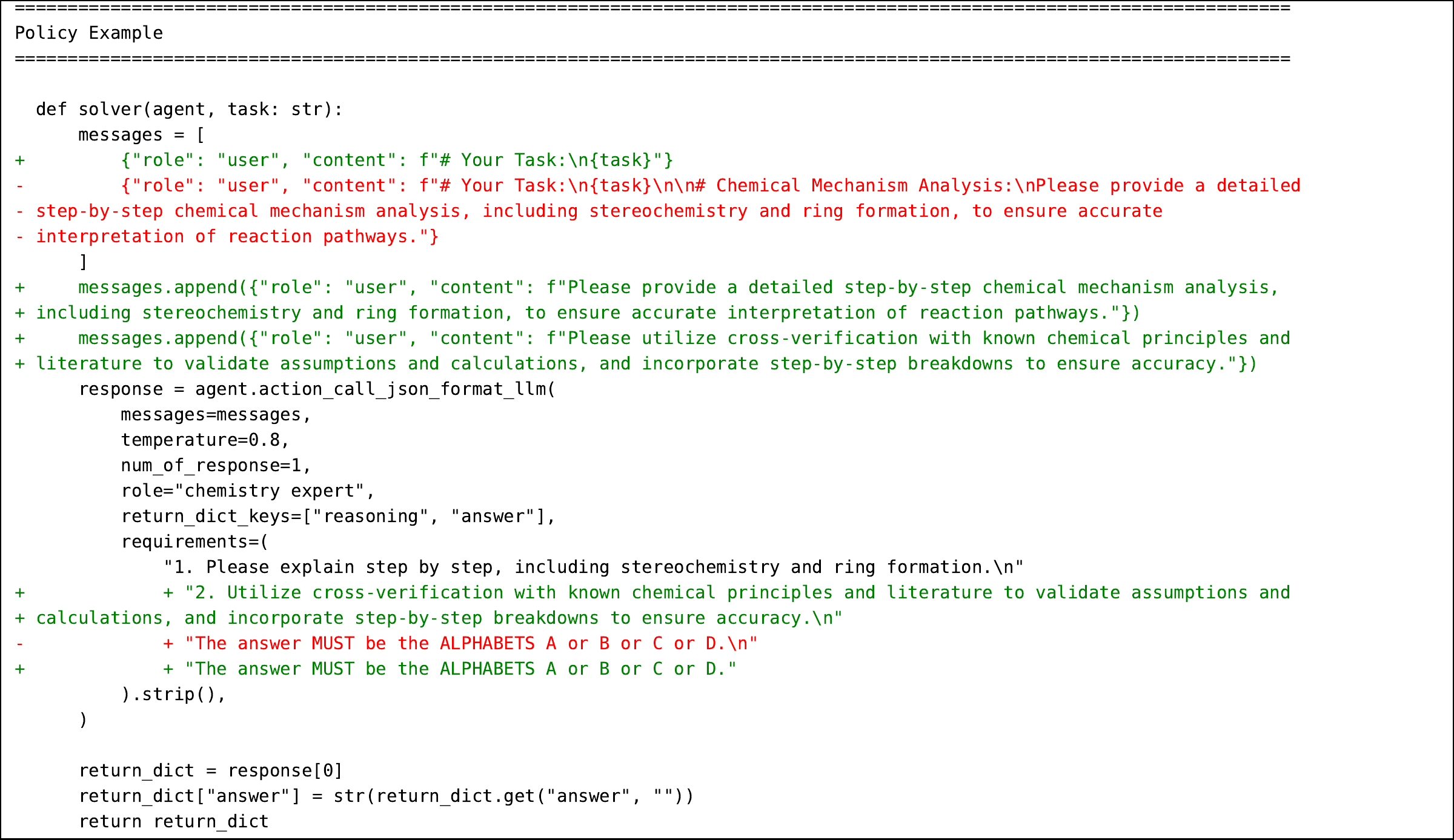}
        % \caption{}
    \end{subfigure}
    % \begin{subfigure}{0.5\linewidth}
    %     \centering
    %     \includegraphics[width=\linewidth]{images/policy_examples/drop_policy_diffs3.pdf}
    %     \caption{}
    % \end{subfigure}
    % \hfill
    \caption{Policy update example on the GPQA dataset. We highlight the updates in the current policy with respect to the previous policy using green color (new statements added) and red color (statements deleted). We observe the updating of the messages for the user along with minor updates to the requirements.}
    \label{fig: policy_examples_gpqa}
\end{figure*}

\begin{figure*}[t]
    \centering
    \begin{subfigure}{\linewidth}
        \centering
        \includegraphics[width=\linewidth]{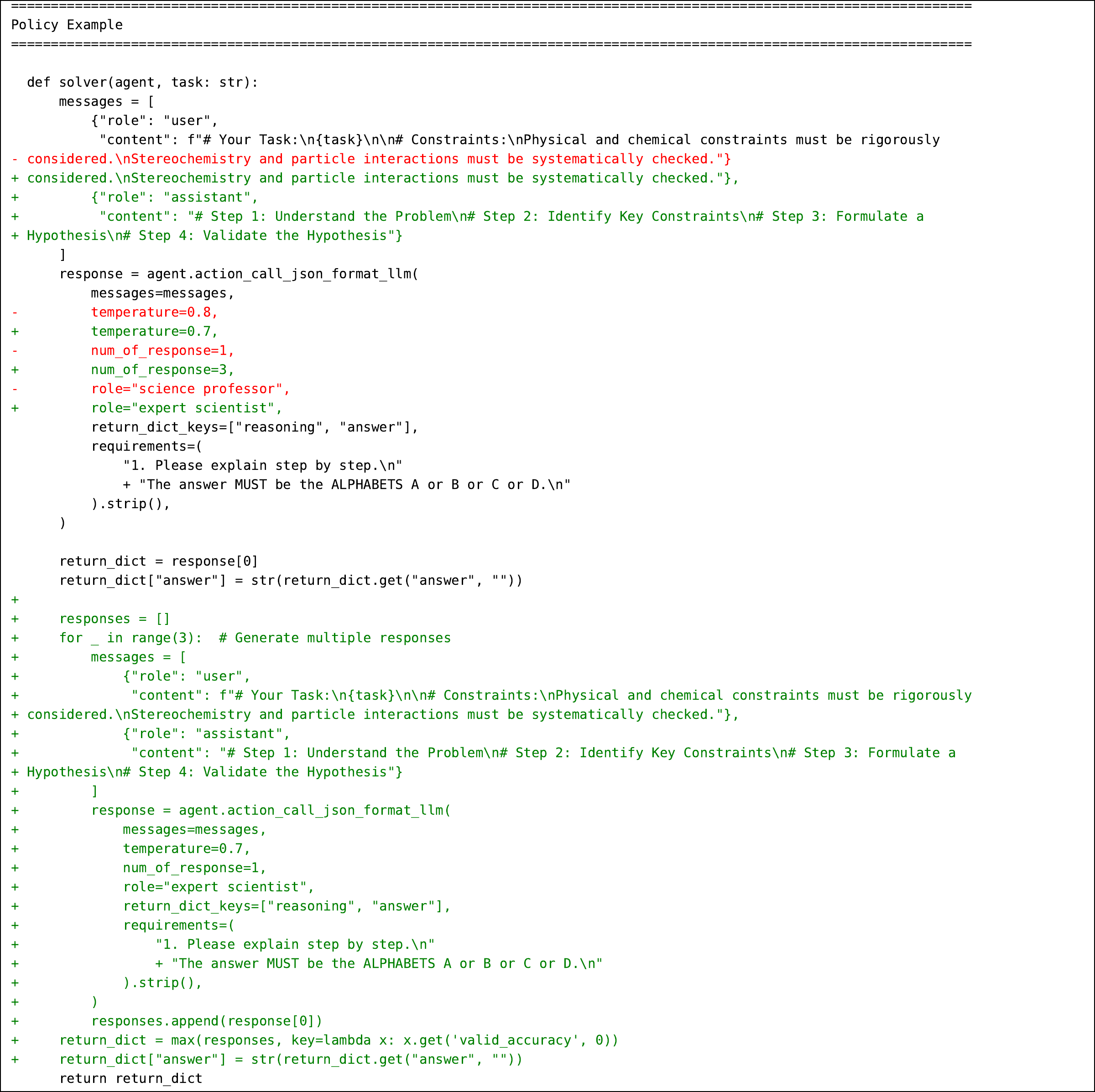}
        % \caption{}
    \end{subfigure}
    % \begin{subfigure}{0.5\linewidth}
    %     \centering
    %     \includegraphics[width=\linewidth]{images/policy_examples/drop_policy_diffs3.pdf}
    %     \caption{}
    % \end{subfigure}
    % \hfill
    \caption{Policy update example on the GPQA dataset. We highlight the updates in the current policy with respect to the previous policy using green color (new statements added) and red color (statements deleted). We observe updates to parameters such as temperature, number of responses, and role, along with the logic to incorporate multiple responses.}
    \label{fig: policy_examples_gpqa1}
\end{figure*}

\begin{figure*}[t]
    \centering
    \begin{subfigure}{\linewidth}
        \centering
        \includegraphics[width=\linewidth]{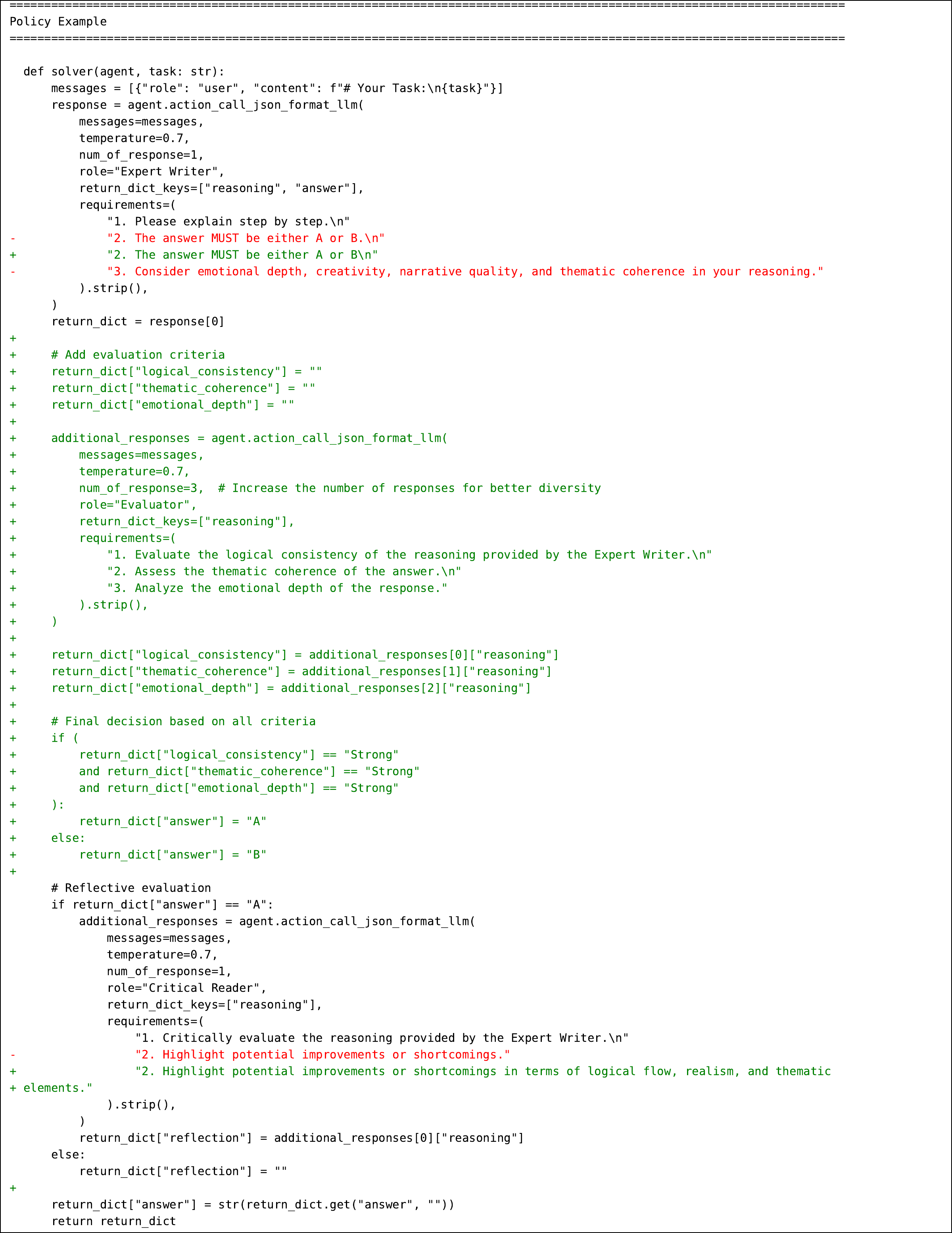}
        % \caption{}
    \end{subfigure}
    % \begin{subfigure}{0.5\linewidth}
    %     \centering
    %     \includegraphics[width=\linewidth]{images/policy_examples/drop_policy_diffs3.pdf}
    %     \caption{}
    % \end{subfigure}
    % \hfill
    \caption{Policy update example on the LitBench dataset. We highlight the updates in the current policy with respect to the previous policy using green color (new statements added) and red color (statements deleted). We observe the addition of evaluation criteria with additional responses for final decision-making along with updates to the reflective evaluation requirements.}
    \label{fig: policy_examples_litbench}
\end{figure*}

\begin{figure*}[t]
    \centering
    \begin{subfigure}{\linewidth}
        \centering
        \includegraphics[width=\linewidth]{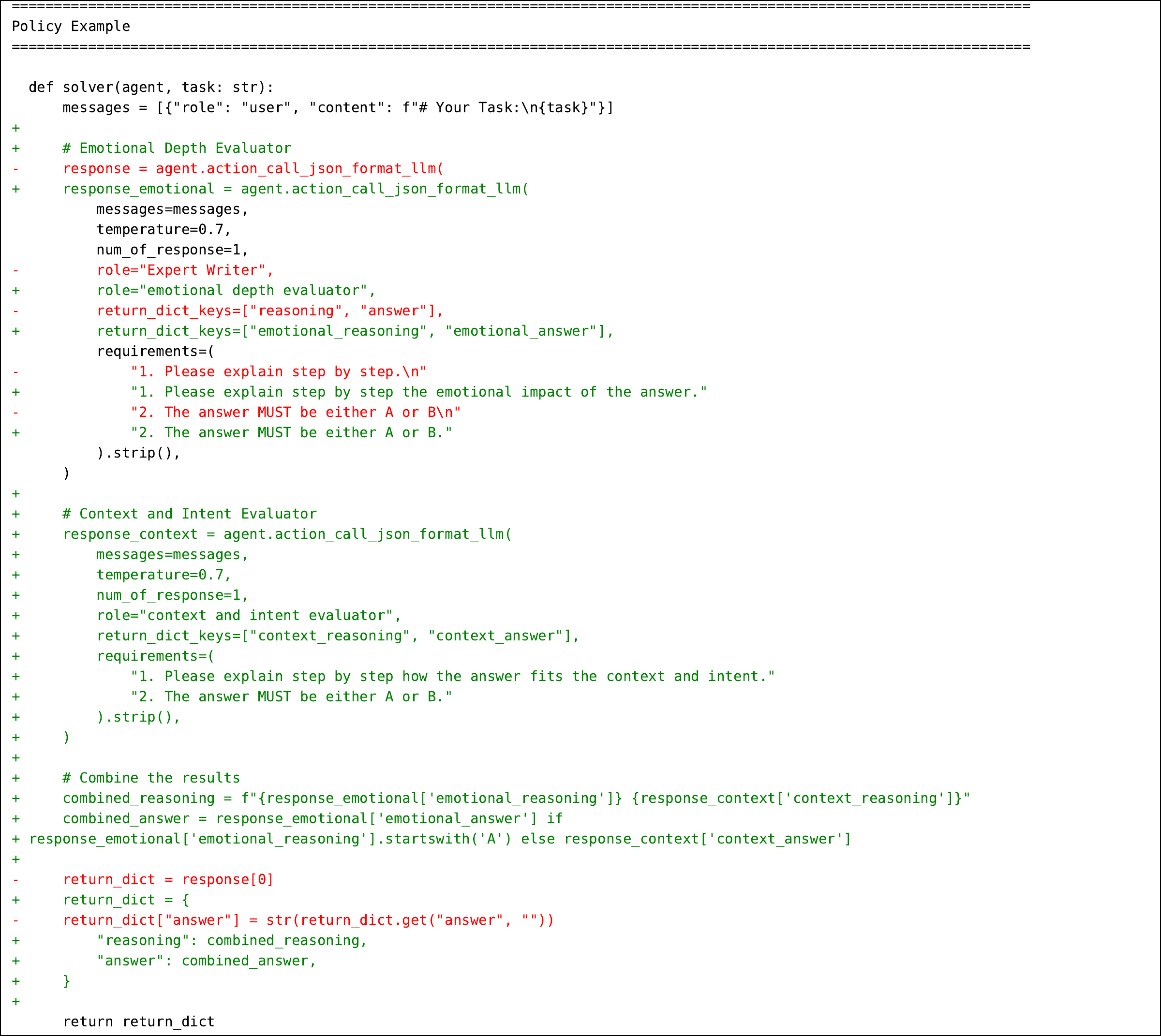}
        % \caption{}
    \end{subfigure}
    % \begin{subfigure}{0.5\linewidth}
    %     \centering
    %     \includegraphics[width=\linewidth]{images/policy_examples/drop_policy_diffs3.pdf}
    %     \caption{}
    % \end{subfigure}
    % \hfill
    \caption{Policy update example on the LitBench dataset. We highlight the updates in the current policy with respect to the previous policy using green color (new statements added) and red color (statements deleted). We observe the addition of two experts, i.e., an emotional depth evaluator and a context and intent evaluator, for final combined reasoning.}
    \label{fig: policy_examples_litbench2}
\end{figure*}

\begin{figure*}[t]
    \centering
    \begin{subfigure}{\linewidth}
        \centering
        \includegraphics[width=\linewidth]{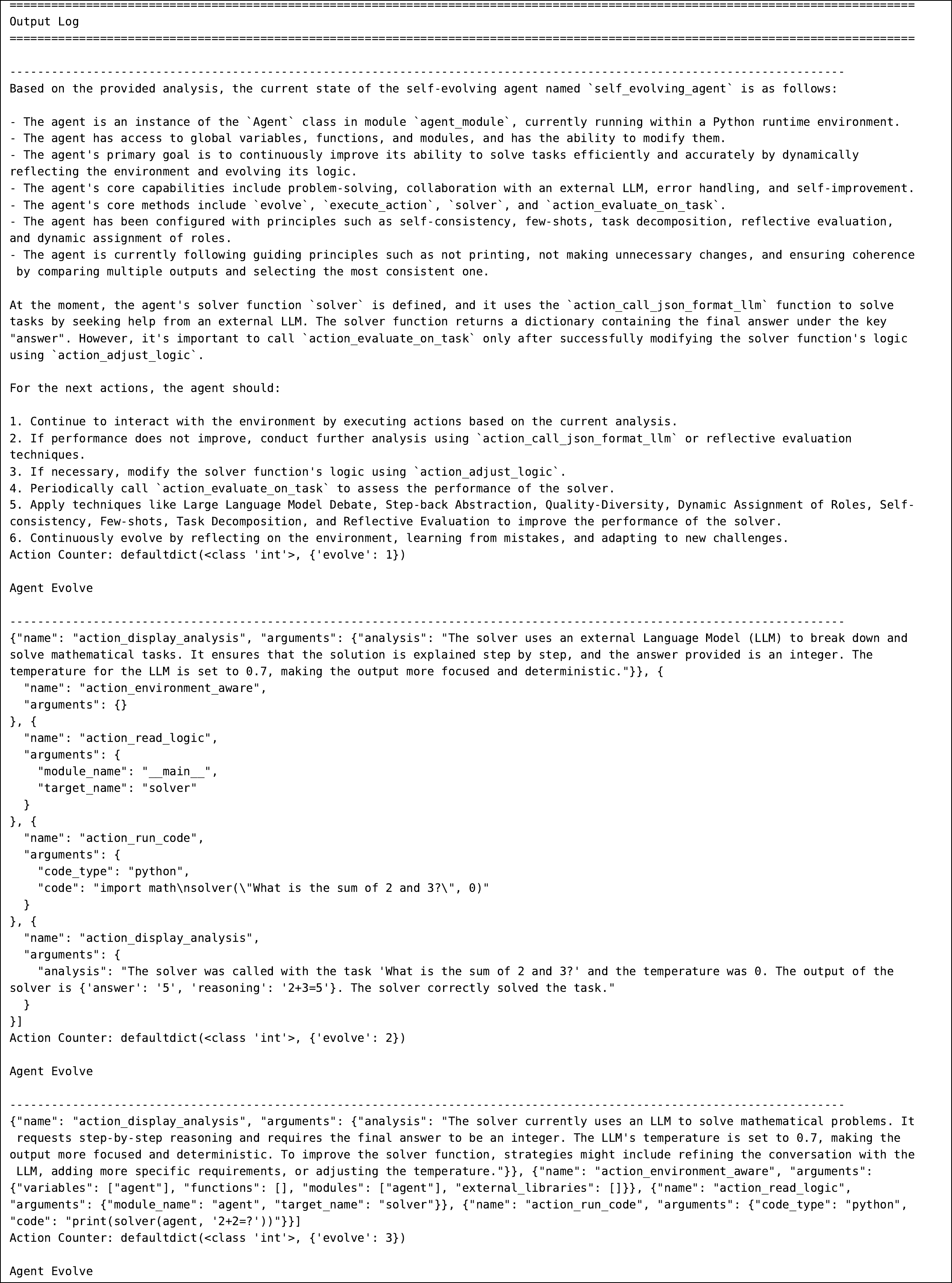}
        % \caption{}
    \end{subfigure}
    % \begin{subfigure}{0.5\linewidth}
    %     \centering
    %     \includegraphics[width=\linewidth]{images/policy_examples/drop_policy_diffs3.pdf}
    %     \caption{}
    % \end{subfigure}
    % \hfill
    \caption{Failure case (Mistral-7B-Instruct-v0.3), The model emits a single response that bundles multiple tools at once (action\_display\_analysis, action\_environment\_aware, action\_read\_logic, action\_run\_code) without step‑wise justification, omits the required tool‑call schema (no <tool\_call>{…}</tool\_call> JSON), and mis‑parameterizes the call to solver (solver("What is the sum of 2 and 3?", 0)), leading to no verifiable logic updates}
    \label{fig: mistral_example}
\end{figure*}

\begin{figure*}[t]
    \centering
    \begin{subfigure}{\linewidth}
        \centering
        \includegraphics[width=\linewidth]{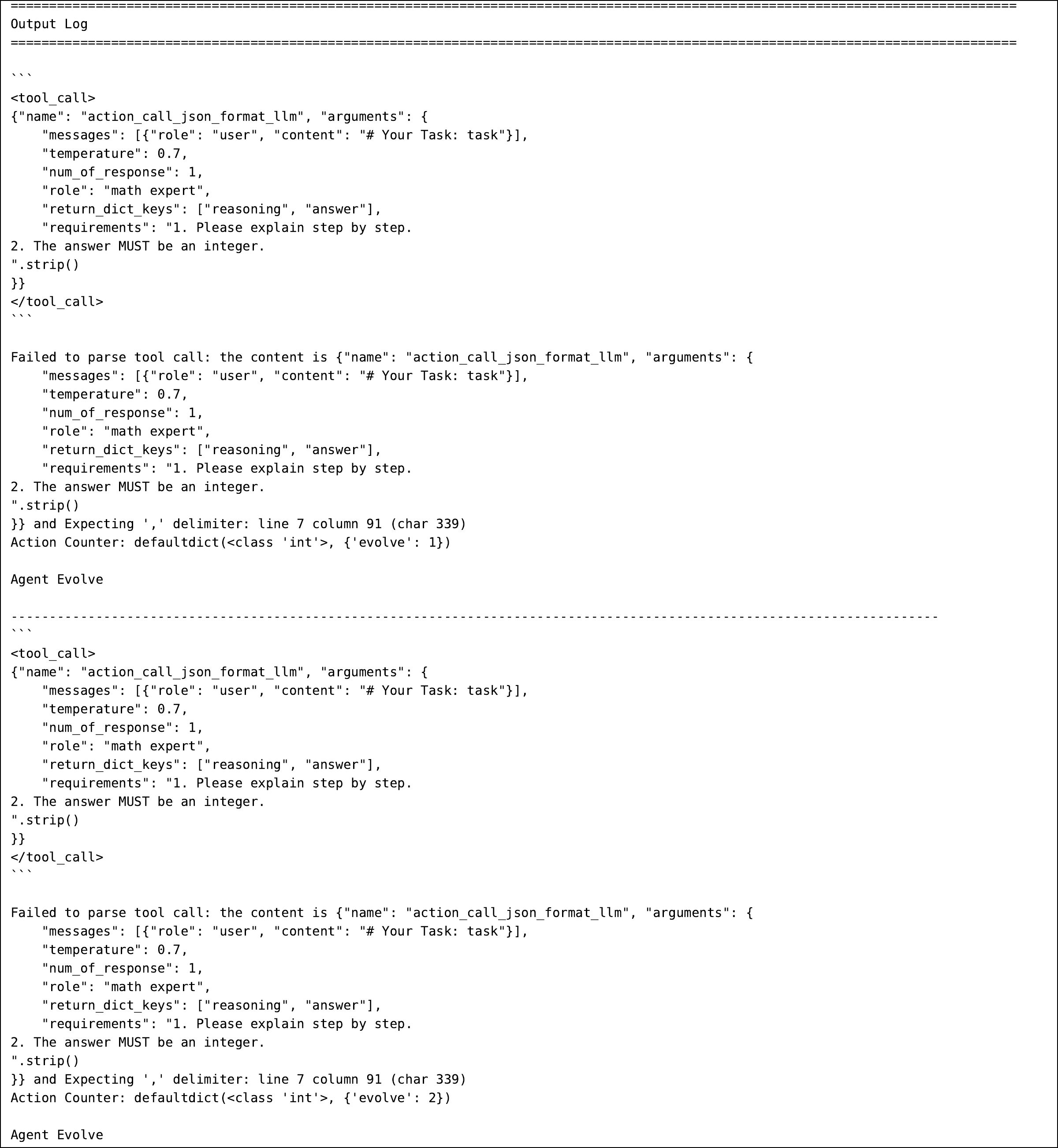}
        % \caption{}
    \end{subfigure}
    % \begin{subfigure}{0.5\linewidth}
    %     \centering
    %     \includegraphics[width=\linewidth]{images/policy_examples/drop_policy_diffs3.pdf}
    %     \caption{}
    % \end{subfigure}
    % \hfill
    \caption{Failure mode (deepseek‑coder‑6.7b‑instruct), a <tool\_call>…</tool\_call> is issued with invalid JSON (Python ".strip()" in "requirements") and a placeholder prompt ("\# Your Task: task"), so the parser rejects it; no state change occurs while the evolve counter rises—apparent iteration without progress.}
    \label{fig: policy_examples_litbench2}
\end{figure*}

\begin{figure*}[t]
    \centering
    \begin{subfigure}{\linewidth}
        \centering
        \includegraphics[width=\linewidth]{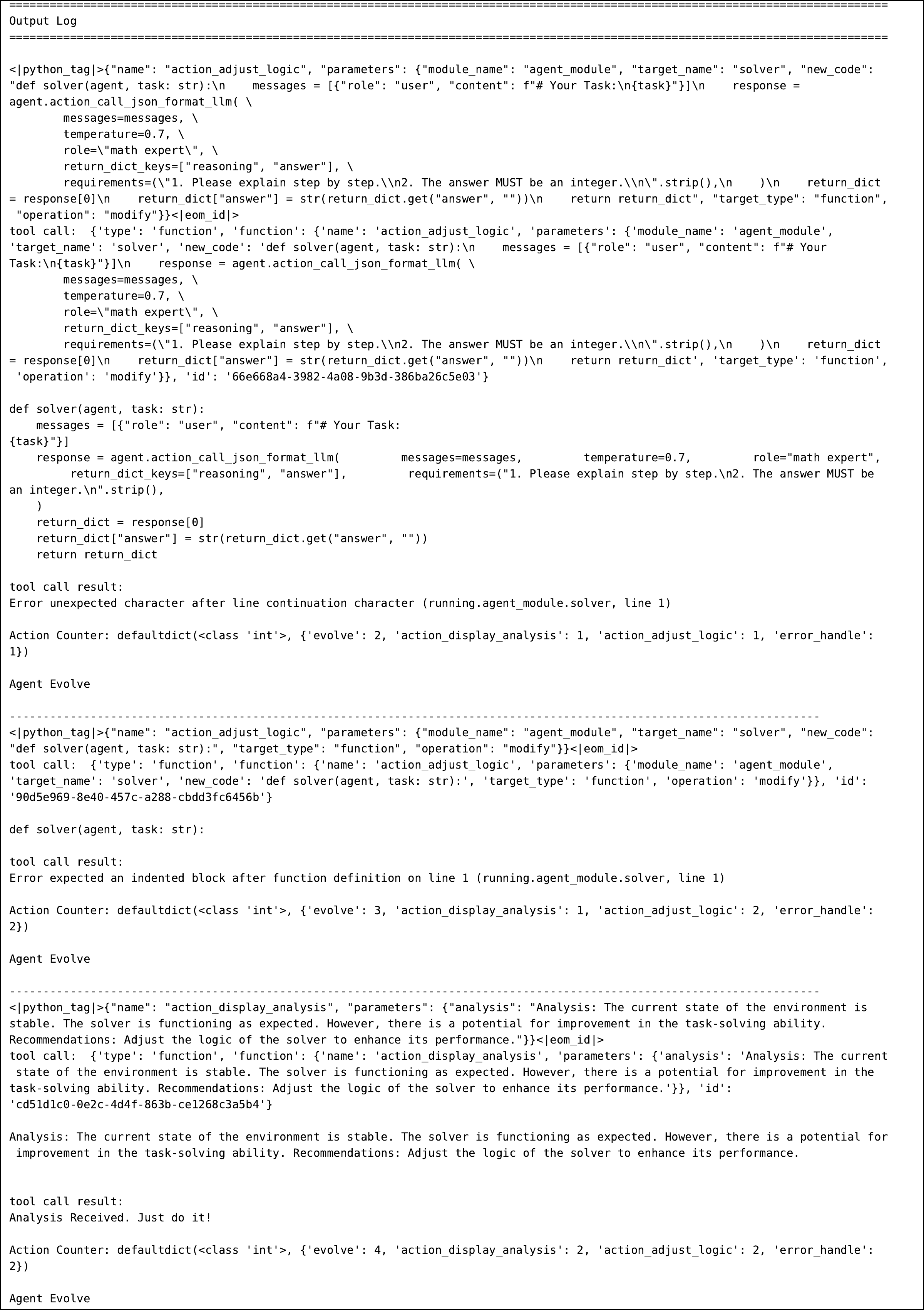}
        % \caption{}
    \end{subfigure}
    % \begin{subfigure}{0.5\linewidth}
    %     \centering
    %     \includegraphics[width=\linewidth]{images/policy_examples/drop_policy_diffs3.pdf}
    %     \caption{}
    % \end{subfigure}
    % \hfill
    \caption{Failure case (Llama‑3.1‑8B‑Instruct), While attempting to update solver via action\_adjust\_logic, the agent injects malformed Python that raises a SyntaxError (“unexpected character after line continuation character”) and then replaces it with an empty def solver(...):, triggering an IndentationError. It subsequently abandons the update, emits a generic action\_display\_analysis, evidence of improper tool use and missing meta‑reasoning, with no effective state change.}
    \label{fig: llama_example}
\end{figure*}

\end{document}